\documentclass[10pt,twocolumn,letterpaper]{article}
	
	\usepackage{cvpr}
	\usepackage{times}
	\usepackage{epsfig}
	\usepackage{graphicx}
	\usepackage{amsmath}
	\usepackage{amssymb}
\usepackage{cleveref}
	\usepackage[export]{adjustbox}
	
	\usepackage[pagebackref=true,breaklinks=true,letterpaper=true,colorlinks,bookmarks=false]{hyperref}
	
	 \cvprfinalcopy 
	
	\def\cvprPaperID{1579} 

	\ifcvprfinal\fi

\crefformat{footnote}{#2\footnotemark[#1]#3}

\begin{document}

\title{The MegaFace Benchmark: 1 Million Faces for  Recognition at Scale}
\author{Ira Kemelmacher-Shlizerman~~~~~ Steve Seitz~~~~~Daniel Miller~~~~~~Evan Brossard\\
Dept. of Computer Science and Engineering\\
University of Washington}
	\teaser{
	\begin{tabular}{ccc}
			\includegraphics[width=.19\linewidth, valign=t]{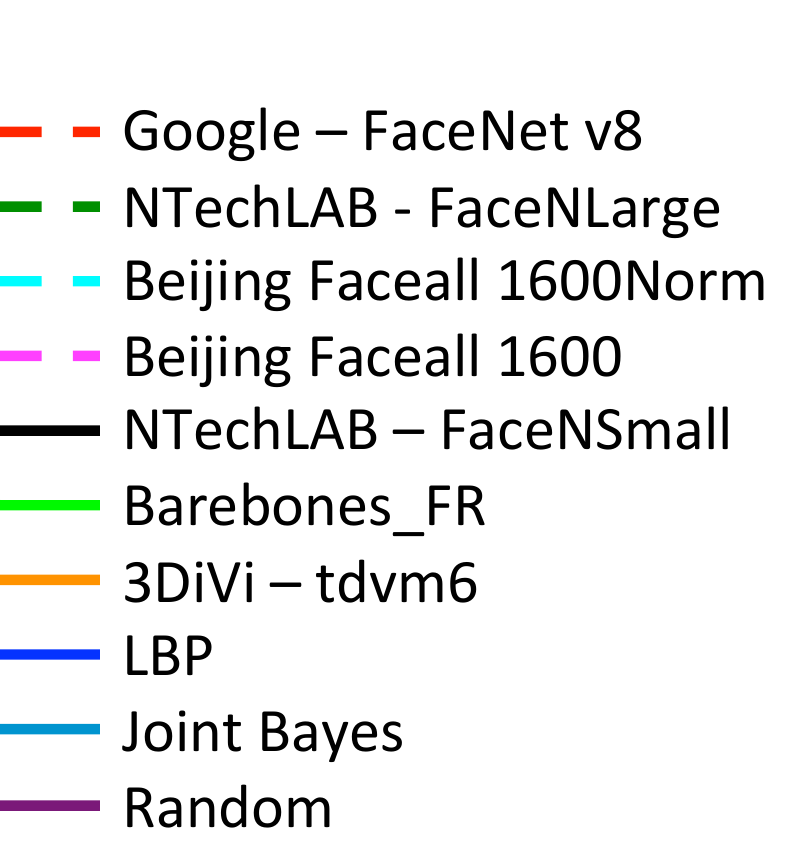} & 
	\includegraphics[width=.31\linewidth, valign=t]{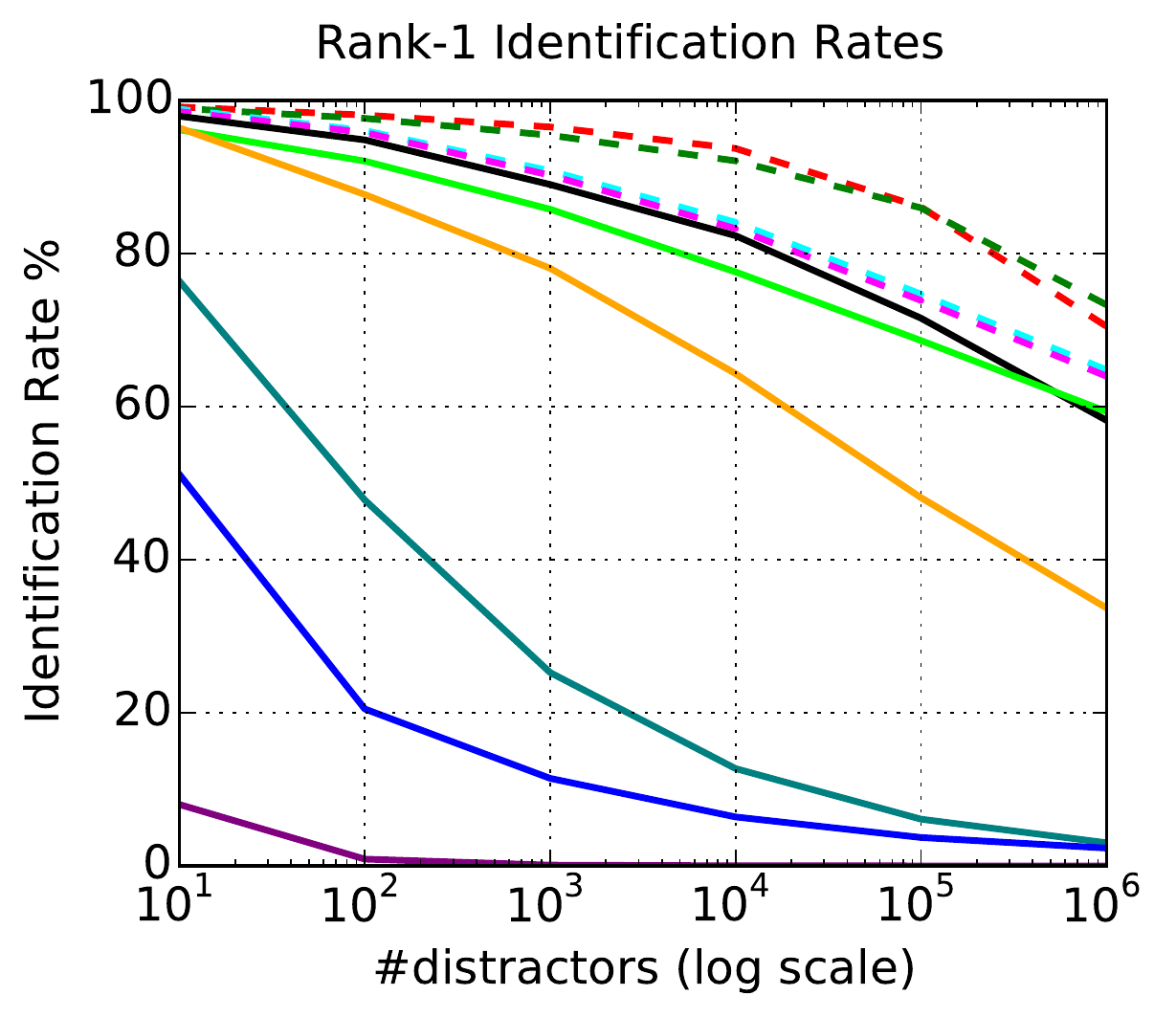} &
	\includegraphics[width=.31\linewidth, valign=t]{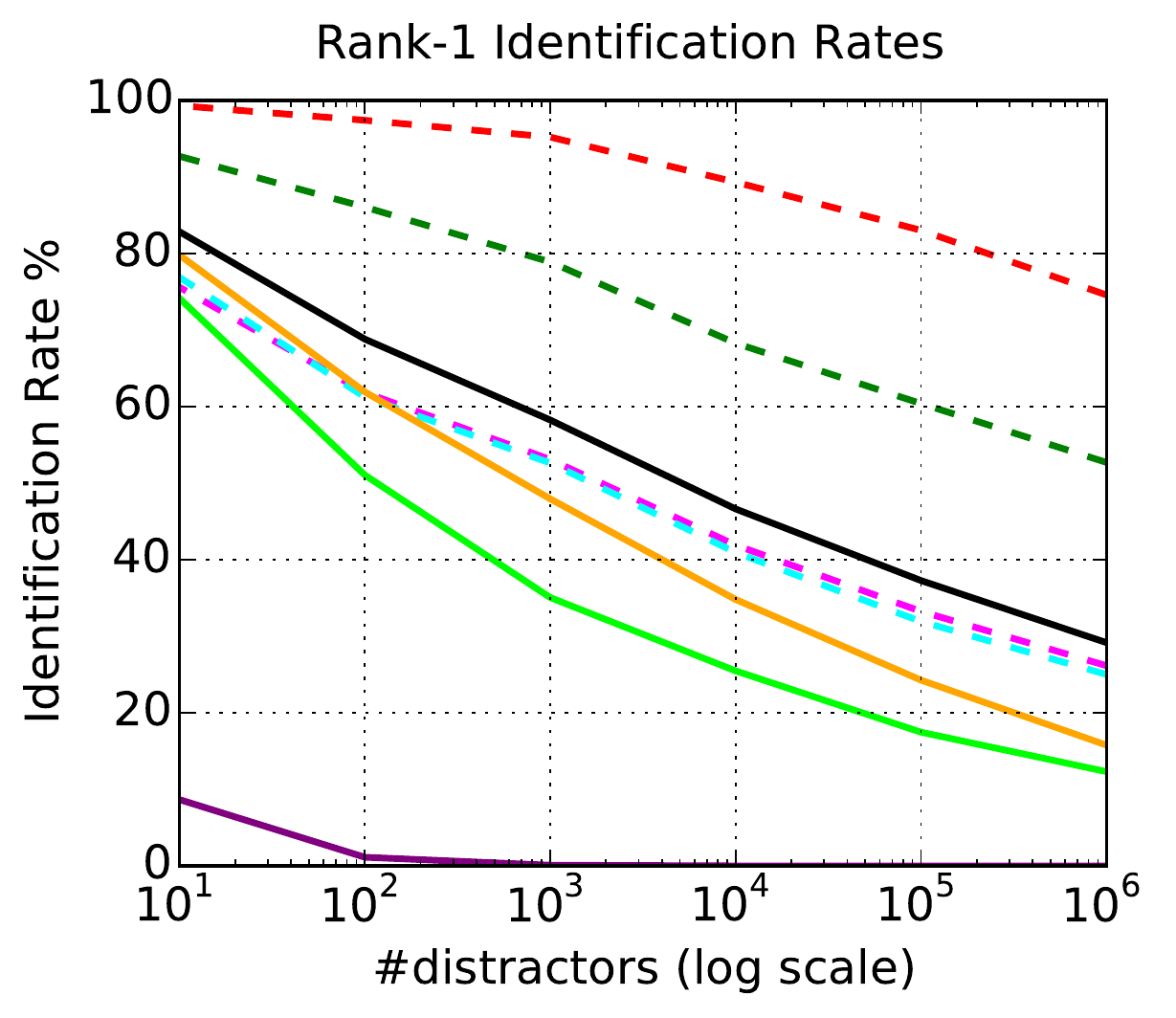} \\
	& {\footnotesize (a) FaceScrub + MegaFace} & {\footnotesize (b)  FGNET + MegaFace}\\
	\vspace{.05in}
	\end{tabular}
		\caption{The MegaFace benchmark evaluates  identification and verification as a function of  increasing  size of gallery  (going from 10 to 1 Million distractors).  We use two probe sets (a) FaceScrub--photos of celebrities, (b) FGNET--photos with a large variation in \textbf{age}. We present rank-1 identification of participating  state of the art algorithms: on the left side of  each plot is current major benchmark LFW scale (i.e., 10 distractors, see how all the top algorithms are clustered above 95\%),  on the right is mega-scale (with a million distractors).  Rates drop with increasing numbers of distractors, even though the probe set is fixed,   algorithms trained on larger sets (dashed lines) generally perform better, and   testing at scale reveals that age invariant recognition is still challenging for most  (b). }
		\label{fig:teaser}
	}
	
\maketitle

\begin{abstract}
	Recent face recognition experiments on a major benchmark (LFW \cite{huang2007labeled}) show stunning performance--a number of algorithms achieve near to perfect score, surpassing human recognition rates. In this paper, we advocate evaluations at the \textbf{million scale}  (LFW includes only 13K photos of 5K people). To this end, we have assembled the MegaFace dataset and created the first MegaFace challenge. Our dataset includes One Million photos that capture more than 690K different individuals. The challenge evaluates performance of algorithms with increasing numbers of ``distractors'' (going from 10 to 1M) in the gallery set. We present both identification and verification performance, evaluate performance with respect to pose and a person’s age, and compare as a function of training data size (\#photos and  \#people).  We report results of state of the art  and baseline algorithms. The MegaFace dataset, baseline code, and evaluation scripts, are all publicly released for further experimentations\footnote{\label{megafaceurl}MegaFace data, code, and challenge can be found at: {\scriptsize \url{http://megaface.cs.washington.edu}}}. 
\end{abstract}



\section{Introduction}


Face recognition has seen major breakthroughs in the last couple of years, with new results by multiple groups   \cite{schroff2015facenet,taigman2014deepface,sun2015deepid3} surpassing human performance on the leading 
Labeled Faces in the Wild (LFW) benchmark \cite{huang2007labeled} and achieving near perfect results.  

Is face recognition solved?  
Many applications require accurate identification at {\em planetary scale}, i.e., finding the best matching face in a database of billions of people.  This is truly like finding a needle in a haystack.  Face recognition algorithms did not deliver when the police were searching for the suspect of the Boston marathon bombing~\cite{klontz2013case}.  Similarly, do you believe that current cell-phone face unlocking programs will protect you against anyone on the planet who might find your lost phone?  These and other face recognition applications require finding the true positive match(es) with negligible false positives.  They also require training and testing on datasets that contain vast numbers of different people.

In this paper, we introduce the {\em MegaFace} dataset and benchmark to evaluate and encourage development of face recognition algorithms at scale.   The goal of MegaFace is to evaluate the performance of current face recognition algorithms with up to a million {\em distractors}, i.e., up to a million people who are not in the test set.  Our key objectives for assembling the  dataset are that  1) it should contain a million photos \textbf{``in the wild''}, i.e., with unconstrained pose, expression, lighting, and exposure, 2) be broad rather than deep, i.e., \textbf{contain many different people} rather than many photos of a small number of people, and most importantly 3) it will be \textbf{publicly available}, to enable benchmarking and distribution within the research community.


\begin{figure*}
	\begin{center}
		\includegraphics[width=1\linewidth]{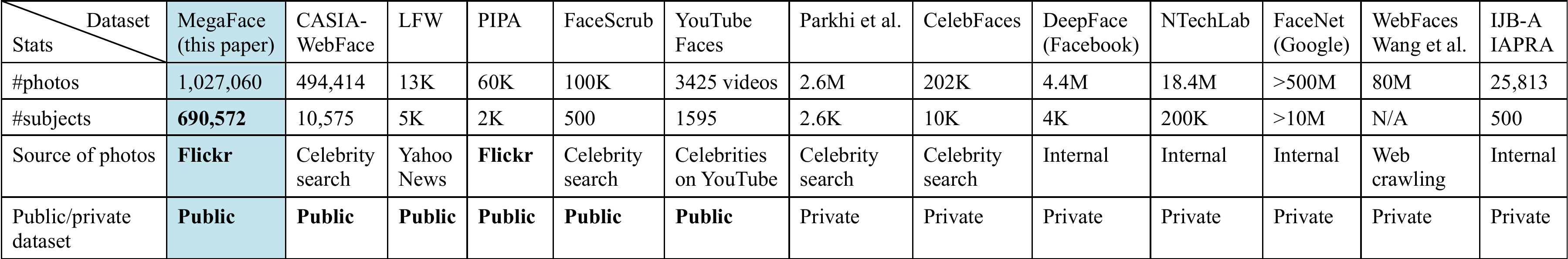}
	\end{center}
	\caption{Representative sample of recent face recognition datasets (in addition to LFW). Current public datasets include up to 10K unique people, and a total of 500K photos.  Several companies have access to  orders of magnitude more  photos and  subjects, these however are subject to privacy constraints and are not public. MegaFace (this paper) includes 1M photos of more than 690K unique subjects, collected from Flickr (from creative commons photos), and is available publicly.}
	\label{fig:datasets}
\end{figure*}

While recent face datasets have leveraged celebrity photos crawled from the web, such datasets have been limited to a few thousand unique individuals;  it is challenging to find a million or more unique celebrities.
Instead, we leverage 
Yahoo's  recently released database of Flickr  photos  \cite{thomee2015new}. The Yahoo dataset includes 100M \textbf{creative commons} photographs and hence can be released for research. 
While these photos are unconstrained and do not target face recognition research per se, they capture a large number of faces.    Our algorithm  samples the Flickr set searching for faces while  optimizing for large number of unique people via analysis of Flickr user IDs and group photos.  MegaFace  includes 1 Million photos of more than 690,000 unique subjects.  

The MegaFace challenge evaluates how face recognition algorithms perform with a very large number of ``distractors,'' i.e., individuals that are not in the probe set.
MegaFace is used as the gallery; the two probe sets we use are FaceScrub \cite{ng265data} and FG-NET \cite{cootes2008fg,kemelmacher2014illumination}.  	
We address fundamental questions and introduce the following key findings (Fig.~\ref{fig:teaser}):
\begin{itemize}

\item{\bf How well do current face recognition algorithms scale?}  
Algorithms that achieve above 95\% performance on LFW (equivalent of 10 distractors in our plots), achieve 35-75\% identification rates with 1M distractors. Baselines (Joint Bayes and LBP) while achieving reasonable results on LFW drop to less than 10\%.

\item \vspace{-0.1in} {\bf Is the size of training data important?}  
We observe that  algorithms that were trained on larger sets (top two are FaceNet that was trained on more than 500M photos of 10M people, and FaceN that was trained on 18M of 200K people) tend to perform better at scale.  Interestingly, however, FaceN  (trained on 18M) compares favorably to FaceNet (trained on 500M) on the FaceScrub set.

\item \vspace{-0.1in} {\bf How does age affect recognition performance?}  We found that the performance with 10 distractors for FGNET as a probe set is lower than for FaceScrub, and the drop off spread is much bigger (Fig.~\ref{fig:teaser} (b)) . A deeper analysis also reveals that children (below age 20) are more challenging to recognize than adults, possibly due to training data availability, and that larger gaps in age (between gallery and probe) are similarly more challenging to recognize. These observations become evident by analyzing at large scale.

\item \vspace{-0.1in} {\bf How does pose affect recognition performance?} Recognition drops for larger variation in pose between matching probe and gallery, and the effect is much more significant at scale.


\end{itemize}

In the following sections we describe how the MegaFace database was created, explain the challenge, and describe the outcomes.

\section{Related Work}

\subsection{Benchmarks}

Early work in face recognition focused on controlled datasets where subsets of lighting, pose, or facial expression were kept fixed, e.g.,  \cite{georghiades1997yale, gross2010multi}. With the advance of algorithms, the focus moved to unconstrained scenarios with a number of important benchmarks appearing,  e.g., FRGC, Caltech Faces, and many more (see \cite{huang2007labeled}, Fig. 3, for a list of all the datasets), and thorough evaluations \cite{grother2010report,zhao2003face}.  A big challenge, however,  was to collect photos of large number of individuals. 

Large scale evaluations were previously performed  on \textit{controlled} datasets (visa photographs, mugshots, lab captured photos) by NIST \cite{grother2010report}, and report recognition results of 90\% on 1.6 million people.  However, these results are not representative of photos in the wild.

In 2007, Huang et al. \cite{huang2007labeled} created the benchmark Labeled Faces in the Wild (LFW). The LFW database includes 13K photos of 5K different people. It was collected by running Viola-Jones face detection \cite{viola2004robust} on  Yahoo News photos. LFW captures  celebrities photographed under  unconstrained conditions (arbitrary lighting,  pose, and expression) and it has been an amazing resource for the face analysis community (more than 1K citations). Since 2007, a number of databases appeared that include larger numbers of photos per person (LFW has 1620 people with more than 2 photos), video information, and even 3D information, e.g.,  \cite{kumar2009attribute, beveridge2013challenge, yi2014learning, wolf2011face, chen2012bayesian,  ng265data}.   However, LFW remains the leading benchmark on which all state of the art recognition methods are evaluated and compared.  Indeed, just in the last year a number of methods (11 methods at the time of writing this paper), e.g.,  \cite{schroff2015facenet,sun2014deeply,sun2015deepid3,taigman2014deepface,taigman2014web} reported recognition rates above 99\%+ \cite{hu2015face} (better than human recognition rates estimated on the same dataset by \cite{kumar2011describable}). The perfect recognition rate on LFW is 99.9\% (it is not 100\% since there are 5 pairs of photos that are mislabeled), and current top performer reports 99.77\%.


\subsection{Datasets}

While, some companies have access to massive photo collections, e.g., Google in \cite{schroff2015facenet}
trained on 200 Million photos of 8 Million people (and more recently on 500M of 10M), these datasets are not available to the public and were used only for training and not testing. 

The largest public data set is
CASIA-WebFace  \cite{yi2014learning} that includes 500K photos of 10K celebrities, crawled from the web.  While CASIA is a great resource, it contains only 10K individuals, and does not have an associated benchmark (i.e., it's used for training not testing). 

Ortiz et al. \cite{ortiz2014face} experimented with large scale identification from Facebook photos assuming there is more than one gallery photo per person. Similarly Stone et al. \cite{stone2008autotagging} show that social network's context improves large scale face recognition. Parkhi et al. \cite{parkhi2015deep} assembled a  dataset of 2.6 Million  of  2600 people, and used it for training (testing was done on the smaller scale LFW and YouTube Faces~\cite{wolf2011face}).  Wang et al. \cite{wang2015face} propose a hierarchical  approach on top of commercial recognizer to enable fast  search in a dataset of 80 million faces.   Unfortunately, however, none of these efforts have produced publicly available datasets or public benchmarks.
Note also that \cite{parkhi2015deep} and \cite{wang2015face} are contemporaneous, as their arxiv papers appeared after ours \cite{miller2015megaface}.


\subsection{Related Studies}

Age-invariant recognition is an important problem that has been studied in the literature, e.g., \cite{chen14cross,li2011discriminative}.  
FG-NET \cite{cootes2008fg} includes 975 photos of 82 people, each with several photos spanning many ages.  More recently, 
Chen et al. \cite{chen14cross} created a dataset of 160k photos of 2k celebrities across many ages.  However, most modern face recognition algorithms have not been evaluated for age-invariance.  We attempt to rectify this by including an FG-NET test (augmented with a million distractors) in our benchmark.

Other recent studies have considered both identification as well as verification results on LFW \cite{best2014unconstrained, taigman2014web, sun2014deeply, sun2015deepid3}. 
Finally, 
Best-Rowden et al. \cite{best2014unconstrained}  performed an interesting Mechanical Turk study to evaluate human recognition rates on LFW and YouTube Faces datasets. They  report  that humans are better than computers when recognizing from videos due to additional cues, e.g., temporal information, familiarity with the subject (celebrity), workers' country of origin (USA vs. others), and also discovered errors in labeling of YouTube Faces via crowdsourcing. In the future, we will use this study's useful conclusions  to help annotate  MegaFace and create a training set in addition to the currently provided distractor set. 



\section{Assembling MegaFace}

\begin{figure*}
	\begin{center}
		\includegraphics[width=1\linewidth]{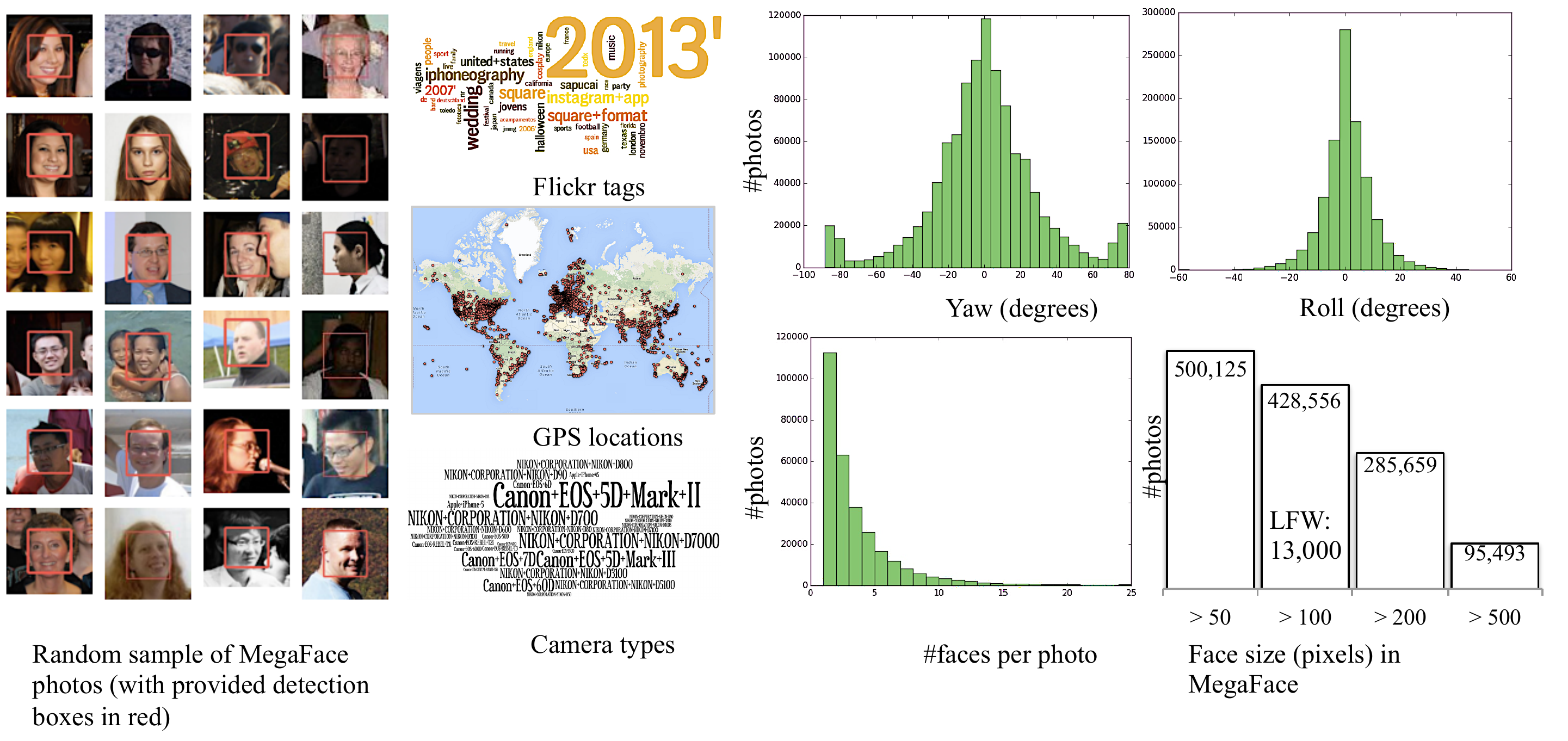}
	\end{center}
	\caption{MegaFace statistics. We present  randomly selected photographs (with provided detections in red), along with  distributions of  Flickr tags,  GPS locations, and camera types. We also show the pose distribution (yaw and roll),  number of faces per photograph, and number of faces for different resolutions (compared to LFW in which faces are approximately 100x100).}
	\label{fig:data_stats} 
\end{figure*}

In this section, we provide an overview of the MegaFace dataset, how it was assembled, and its statistics.
We created MegaFace to evaluate and drive the development of face recognition algorithms that work at scale.
As motivated in Section 1, we sought to create a public dataset, free of licensing restrictions, that captures photos taken with unconstrained imaging conditions, and with close to a million unique identities.
After exploring a number of avenues for data collection, we decided to leverage Yahoo's 100M Flickr set \cite{thomee2015new}. Yahoo's set was not created with face analysis in mind, however, it includes a very large number of faces and satisfies our requirements.



\textbf{Optimizing for large number of unique identities.} 
Our strategy for maximizing the number of unique identities is based on two techniques:  1) drawing photos from many different Flickr users---there are 500K unique user IDs---and 2) assuming that two or more faces appear in the same photo, they are likely different identities.  Note that these assumptions do not need to be infallible, as our goal is to produce a very diverse distractor set--it is not a problem if we have a small number of photos of the same person.
Our algorithm for detecting and downloading faces is as follows. We generated a list of images and user IDs in a round-robin fashion,
by going through each of the 500K users and selecting the first photo with a face larger than $50\times 50$ and adding it to the dataset.  If the photo contains multiple faces above that resolution, we add them all, given that they are different people with high probability.  We then repeated this process (choosing the second, then the third, etc. photo from each user), until a sufficient number of faces were assembled.
Based on our experiments face detection can have up to 20\% false positive rate.  Therefore, to  ensure that our final set includes a million faces, the  process was terminated once  $1,296,079$ faces were downloaded.  Once face detection was done, we ran additional stricter detection, and removed blurry faces.  We assembled a total of $690,572$ faces in this manner that have a high probability of being unique individuals.
While not guaranteed, the remaining $310$K in our dataset likely also contain additional unique identities. Figure~\ref{fig:data_stats} presents a histogram of number of faces per photo.


\textbf{Face processing.}  We downloaded the highest resolution available per photo. The faces are detected using the HeadHunter\footnote{{\scriptsize \url{http://markusmathias.bitbucket.org/2014_eccv_face_detection/}}} algorithm by Mathias et al. \cite{Mathias2014Eccv}, which reported state of the art results in face detection, and is especially robust to a wide range of head poses including profiles.   We crop detected faces such that the face spans 50\% of the photo height, thus including the full head (Fig.~\ref{fig:data_stats}). We further estimate  49 fiducial points and yaw and pitch angles, as computed by the IntraFace\footnote{{\scriptsize \url{http://www.humansensing.cs.cmu.edu/intraface/}}} landmark model \cite{xiong2013supervised}.

\textbf{Dataset statistics.}
Figure~\ref{fig:data_stats} presents MegaFace's statistics: 
\begin{itemize}
	\item \vspace{-.1in} Representative photographs and bounding boxes. Observe that the photographs contain people from different countries, gender, variety of poses, glasses/no glasses, and many more variations. 
	\item \vspace{-.1in}  Distribution of Flickr tags that accompanied the downloaded photos.   Tags range from 'instagram' to 'wedding,' suggesting a range of photos from selfies to high quality portraits (prominence of '2013' likely due to timing of when the Flickr dataset was released).
	\item \vspace{-.1in}  GPS locations demonstrate photos taken all over the world.
	\item \vspace{-.1in}  Camera types dominated by DSLRs (over mobile phones), perhaps correlated with creative commons publishers, as well as our preference for higher resolution faces.
	\item \vspace{-.1in}  3D pose information:  more than 197K of the faces have yaw angles \textit{larger} than $\pm 40$ degrees. Typically unconstrained face datasets include yaw angles of \textit{less} than $\pm 30$ degrees.
	\item \vspace{-.1in}  Number of faces per photo, to indicate the number of group photos. 
	\item \vspace{-.1in}  Face resolution: more than 50\% (514K) of the  photos in MegaFace have resolution more than 40 pixels  inter-ocular distance (40 IOD corresponds to 100x100 face size, the resolution in LFW).
\end{itemize}
\vspace{-.1in}   We believe that this dataset is extremely useful for a variety of research areas in recognition and face modeling, and we plan to maintain and expand it in the future. In the next section, we describe the MegaFace challenge.

\section{The MegaFace Challenge}

In this section, we describe the challenge and evaluation protocols. 
Our goal is to test performance of face recognition algorithms with up to a million distractors, i.e., faces of unknown people.
In each test, a {\em probe} image is compared against a {\em gallery} of up to a million faces drawn from the Megaface dataset.

\textbf{Recognition scenarios}  The first scenario is identification:  given a probe photo, and a gallery containing at least one photo of the same person, the algorithm rank-orders all photos in the gallery based on similarity to the probe. 
Specifically, the probe set includes $N$ people; for each person we have $M$ photos. We then  test each of the $M$ photos (denote by $i$) per person by adding it the gallery of distractors and use each of the other $M-1$ photos as a probe. Results are presented with Cumulative Match Characteristics (CMC) curves-- the probability that a correct gallery image will be chosen for a random probe by rank $=K$.

The second scenario is verification, i.e., a pair of photos is given and the algorithm should output whether the person in the two photos is the same or not.  To evaluate verification we computed all pairs between the probe dataset and the Megaface distractor dataset.
Our verification experiment has in total 4 billion negative pairs. We report verification results with ROC curves; this explores the trade off between falsely accepting non-match pairs and falsely rejecting match pairs. 

Until now, verification received most of  the focus in  face recognition research since it was tested by the LFW benchmark \cite{huang2007labeled}. Recently, a number of groups, e.g., \cite{best2014unconstrained, taigman2014web, sun2014deeply, sun2015deepid3} also performed identification experiments on LFW. The relation between the identification and verification protocols was studied  by Grother and Phillips \cite{grother2004models} and DeCann and Ross \cite{decann2012can}.  In our challenge, we evaluate both scenarios with an emphasis on very large number of  distractors. For comparison, testing identification on LFW is equivalent  to 10 distractors in our challenge.

\textbf{Probe set.}  MegaFace is used to create a gallery with a large number of distractors.  For the probe set (testing known identities), we use two sets: 
\begin{enumerate}
	\item  The FaceScrub dataset \cite{ng265data}, which includes 100K photos of 530 celebrities, is available online.
	FaceScrub has a similar number of male and female photos (55,742 photos of 265 males and 52,076 photos of 265 females) and a large variation across photos of the same individual which reduces possible bias, e.g.,  due to backgrounds and hair style \cite{kumar2011describable}, that may occur in LFW.  For efficiency, the evaluation was done on a subset of FaceScrub which includes 80 identities (40 females and 40 males) by randomly selecting from a set of people that had more than 50 images each (from which 50 random photos per person were used). 
	\item The FG-NET aging dataset \cite{cootes2008fg,kemelmacher2014illumination}: it includes 975 photos of 82 people. For some of the people the age range in photos is more than 40 years. 
\end{enumerate}


\textbf{Evaluation and Baselines.} Challenge participants were asked to calculate their features on MegaFace, full FaceScrub, and FGNET.  We  provided code that runs identification and verification on the FaceScrub set. After the results were submitted by all groups we re-ran the experiments with FaceScrub and 3 different random distractor sets per gallery size. We further ran the FGNET experiments on all methods\footnote{Google's FaceNet was ran by the authors since their features could not be uploaded due to licensing conditions} and each of the three random MegaFace subsets per gallery size. The metric for comparison is  $L_2$ distance. Participants were asked not to train on FaceScrub or FGNET.  As a baseline,  we implemented two simple recognition algorithms: 1) comparison by LBP \cite{ahonen2006face} features--it achieves 70\% recognition rates on LFW, and uses no training, 2) a Joint Bayesian (JB) approach represents each face as the sum of two Gaussian variables
$x = \mu + \epsilon$ where $\mu$ is identity and $\epsilon$ is inter-personal variation.  To determine whether two faces, $x_1$ and $x_2$ belong to the same identity, we calculate $P(x_1, x_2 | H_1)$ and $P(x_1, x_2 | H_2)$ where $H_1$ is the hypothesis that the two faces are the same and $H_2$ is the hypothesis that the two faces are different. These distributions can also be written as normal distributions, which allows for efficient inference via a log-likelihood test. JB  algorithm was trained on the CASIA-WebFace dataset \cite{yi2014learning}.

	\begin{figure*}
		\centering
		\includegraphics[width=.8\linewidth]{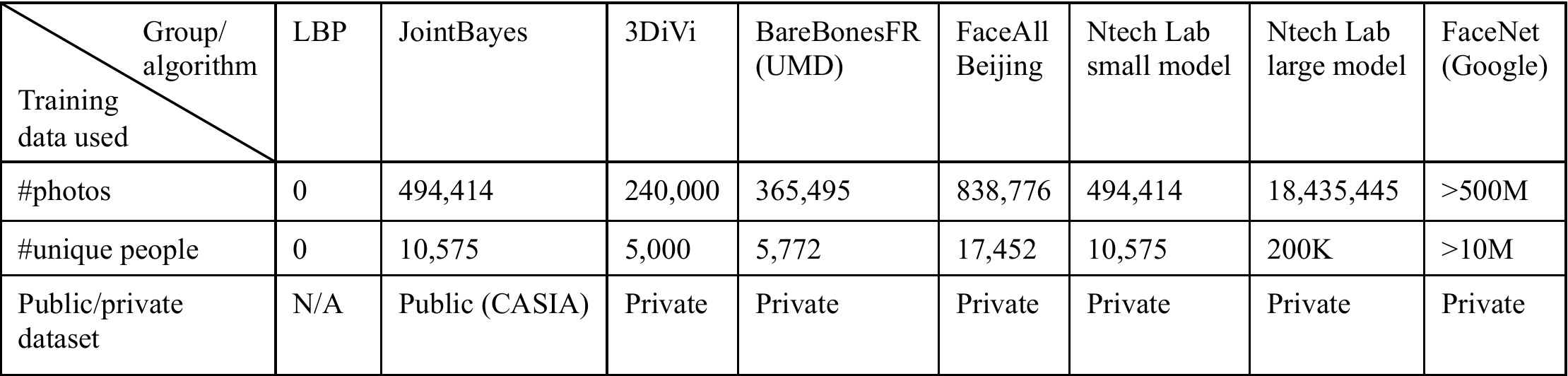}
		\caption{Number of training photos and unique people used by each participating method.  }\label{fig:trainingsize}
	\end{figure*}

	\section{Results}

	\begin{figure*}
		\centering
		\begin{tabular}{ccc}
			\includegraphics[width=.15\linewidth, valign=t]{plots16/ll.pdf} & 
			\includegraphics[width=0.27\linewidth, valign=t]{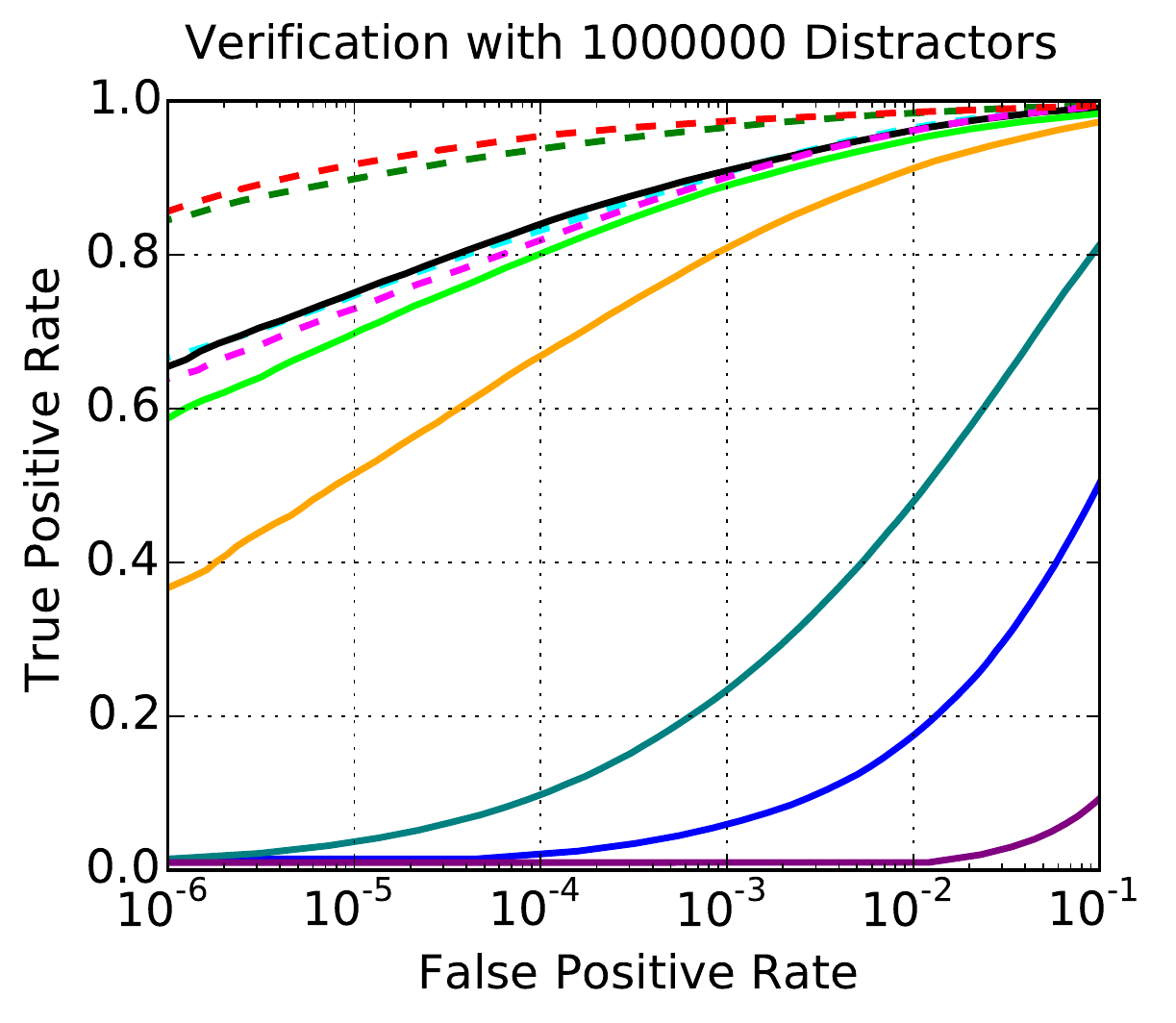} & 
			\includegraphics[width=0.27\linewidth, valign=t]{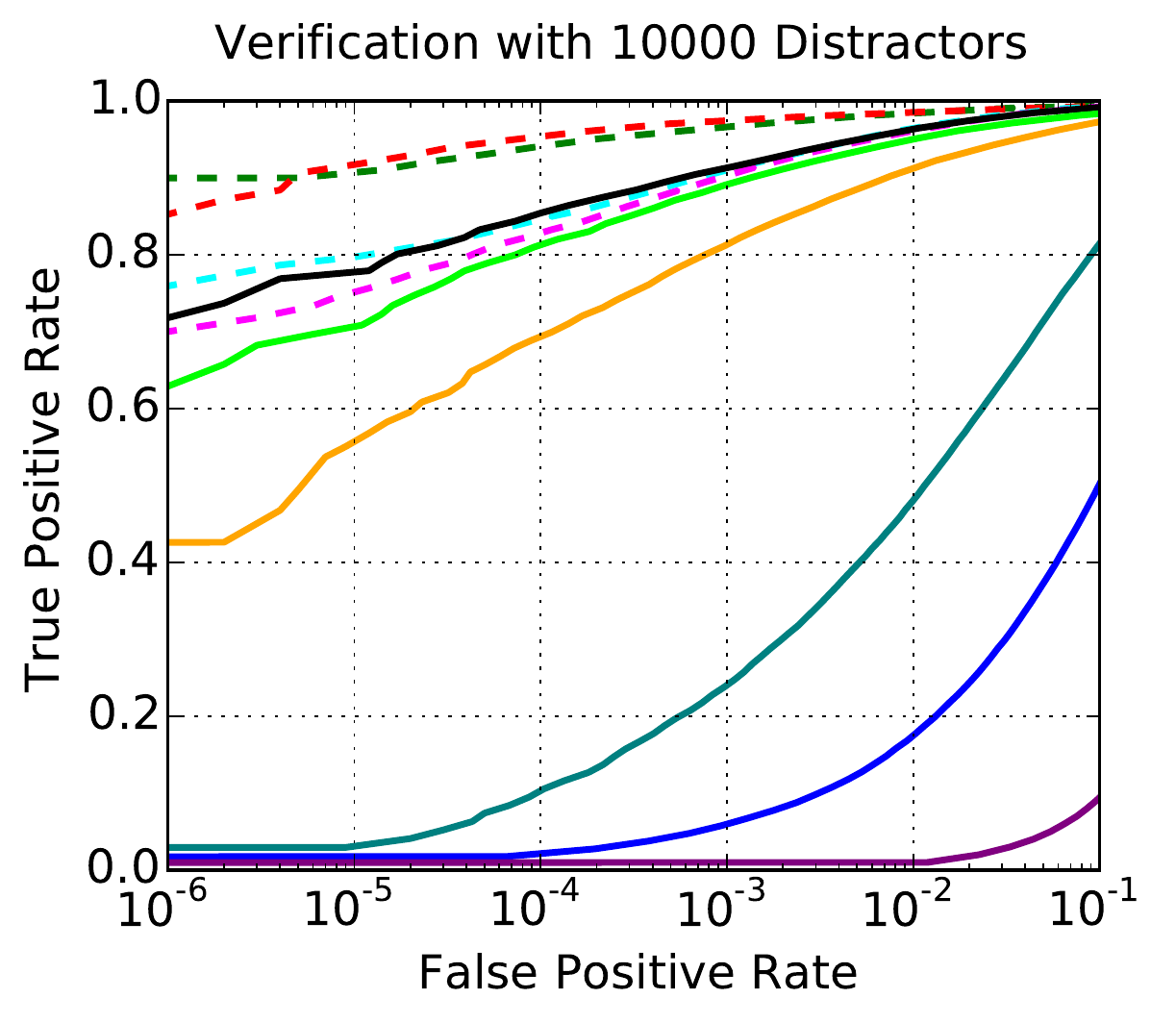} \\
			&{\footnotesize  (a) FaceScrub + 1M} & {\footnotesize (b) FaceScrub + 10K}  \\
			\includegraphics[width=.15\linewidth, valign=t]{plots16/ll.pdf} & 
			\includegraphics[width=0.27\linewidth, valign=t]{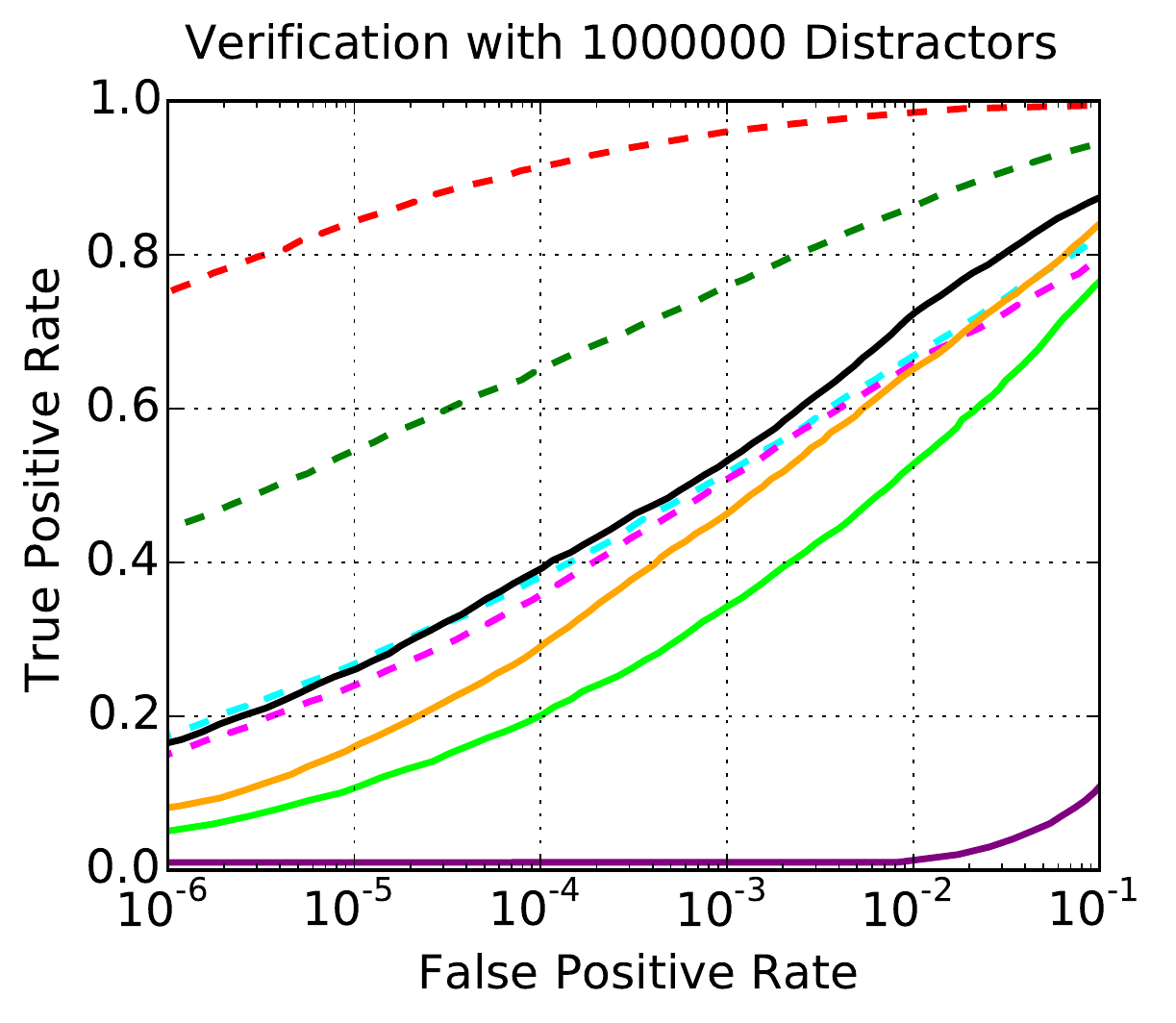} & 
			\includegraphics[width=0.27\linewidth, valign=t]{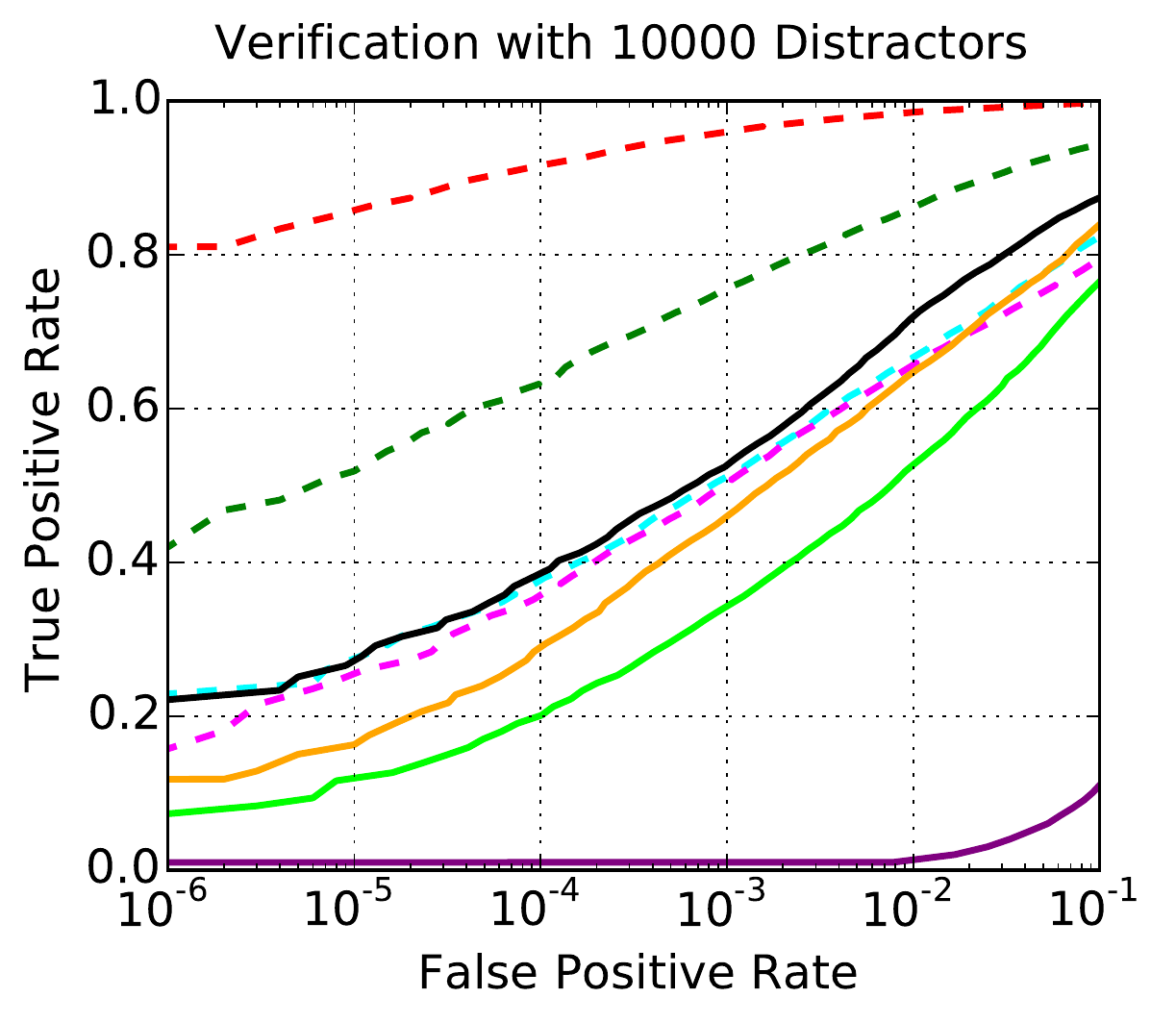} \\
			&{\footnotesize  (c) FGNET + 1M} & {\footnotesize (d) FGNET + 10K}  \\
		\end{tabular}
		\caption{\textbf{Verification} performance with (a,c) 1 Million and (b,d) 10K distractors on both probe sets. Note the performance at low false accept rates (left side of each plot). }\label{fig:verf}
		\label{fig:dataset_size_roc} 
	\end{figure*}
		
This section describes the results and analysis  of the challenge. Our challenge was released on Sep 30, 2015.  Groups were given three weeks to finish their evaluations.  More than 100 groups registered to participate. We present results from 5 groups  that  uploaded all their features by the deadline. 	We keep maintaining the challenge and data--currently 20 more groups  are working on their submissions. 
	
	\textbf{Participating algorithms} In addition to baseline algorithms LBP, and Joint Bayes, we  present results of the following methods (some provided more than 1 model):
	\begin{enumerate}
		\item \vspace{-.1in}Google's FaceNet:  achieves 99.6\% on LFW, was trained on more than 500M photos of 10M people (newer version of \cite{schroff2015facenet}).	
		\item \vspace{-.1in}FaceAll (Beijing University of Post and Telecommunication), was trained on 838K photos of 17K people, and provided two types of features. 
		\item\vspace{-.1in}  NTechLAB.com (FaceN algorithm): provided two models (small and large)--small was trained on 494K photos of 10K people, large on more than 18M of 200K. 
		\item \vspace{-.1in} BareBonesFR (University group): was trained on 365K photos of 5K people. 
		\item \vspace{-.1in} 3DiVi.com: was trained on 240K photos of 5K people.   
	\end{enumerate}
	\vspace{-.1in}Figure~\ref{fig:trainingsize} summarizes the models, training sizes (240K-500M photos, 5K-10M people) and availability of the training data. Below we describe all the experiments and key conclusions.

%

	\begin{figure}
		\includegraphics[width=1\linewidth]{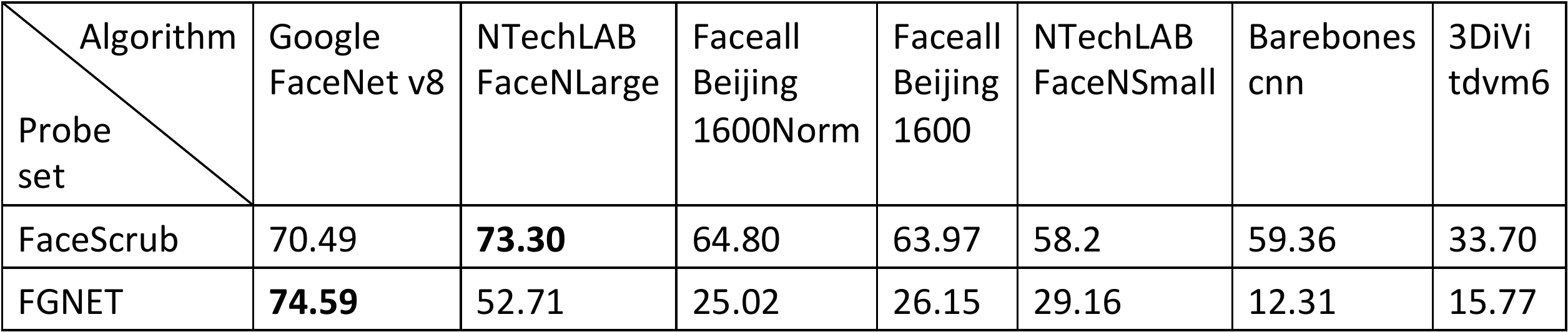}
		\caption{Rank-1 identification results (in\%) with 1M distractors on the two probe sets.}
		\label{fig:rank1numbers}
	\end{figure}

					\begin{figure*}
						\centering
						\begin{tabular}{cccc}
										\includegraphics[width=.15\linewidth, valign=t]{plots16/ll.pdf} & 
	\includegraphics[width=0.25\linewidth, valign=t]{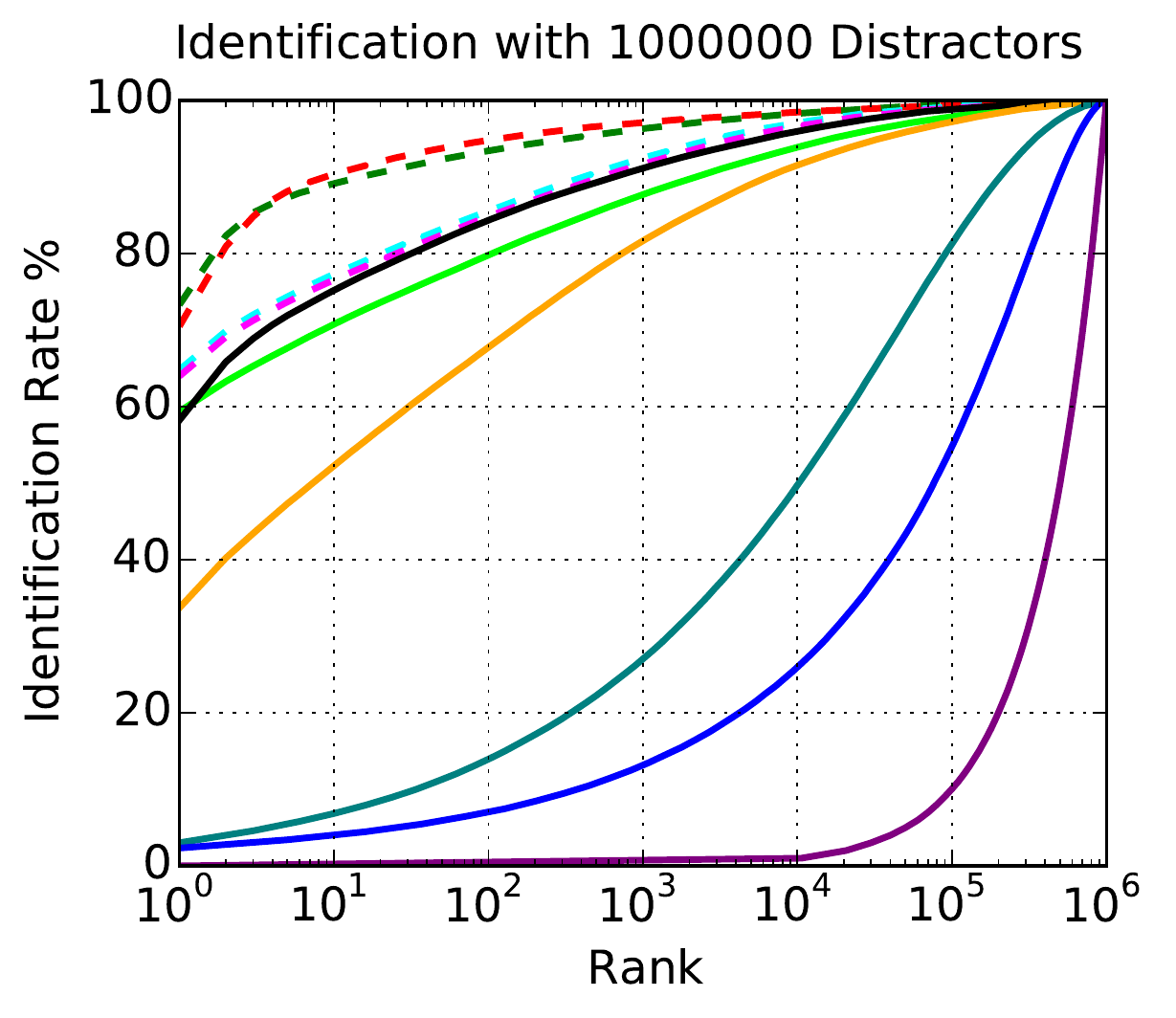} & 
	\includegraphics[width=0.25\linewidth, valign=t]{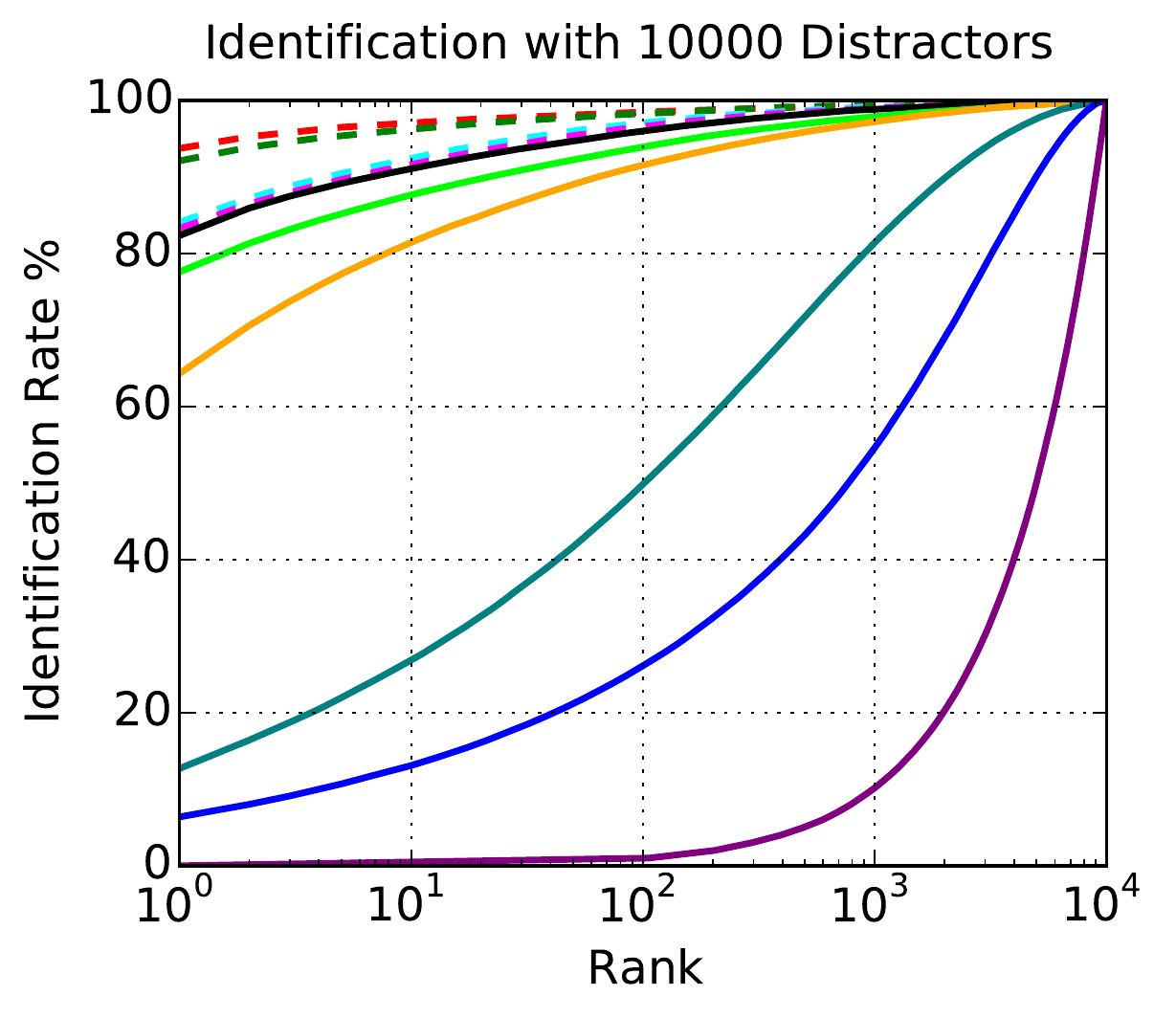} & 
		\includegraphics[width=0.25\linewidth, valign=t]{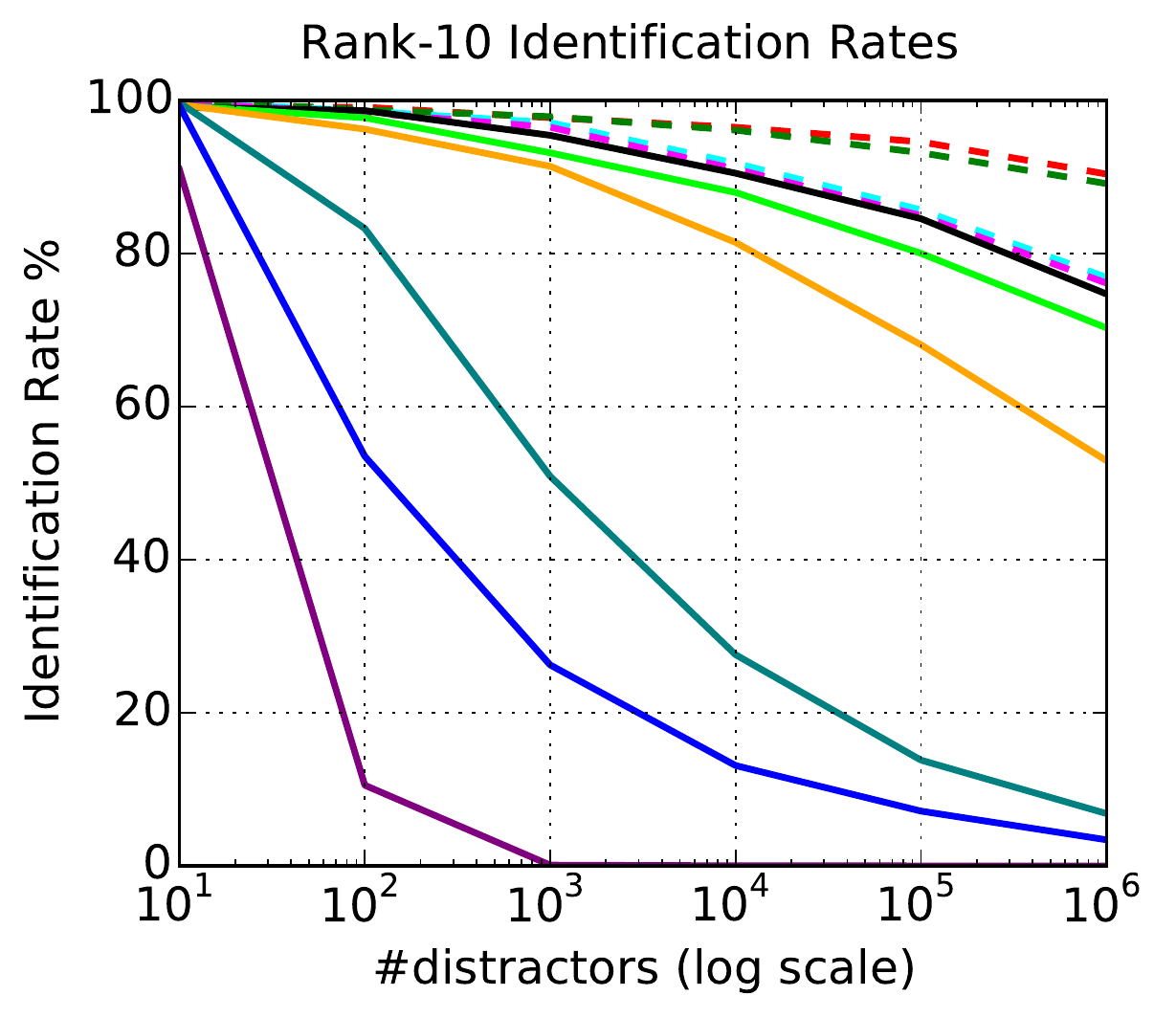}\\  
		&{\footnotesize  (a) FaceScrub + 1M} & {\footnotesize (b) FaceScrub + 10K} &    {\footnotesize (c) FaceScrub + rank-10} \\
												\includegraphics[width=.15\linewidth, valign=t]{plots16/ll.pdf} & 
	\includegraphics[width=0.25\linewidth, valign=t]{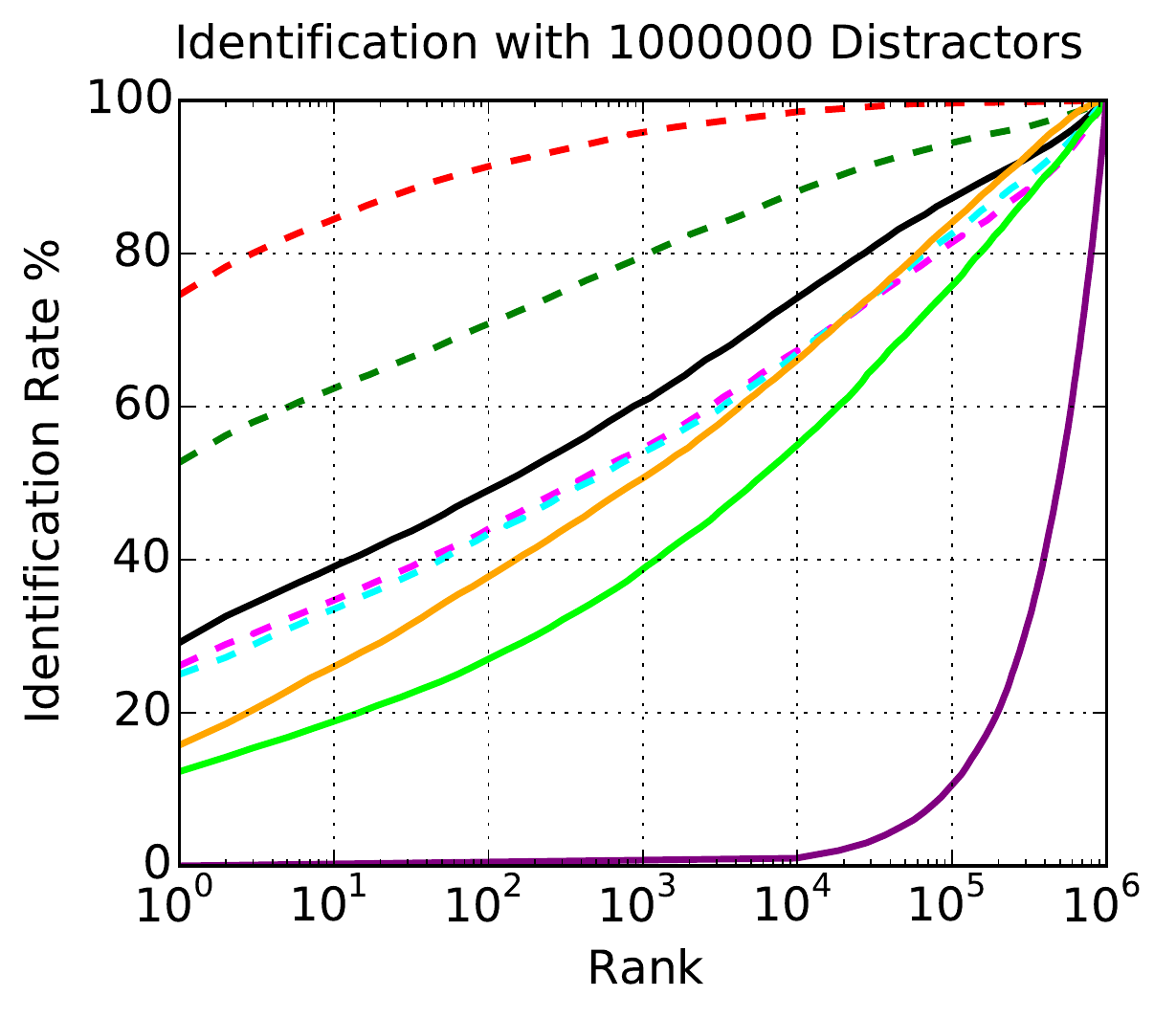} & 
	\includegraphics[width=0.25\linewidth, valign=t]{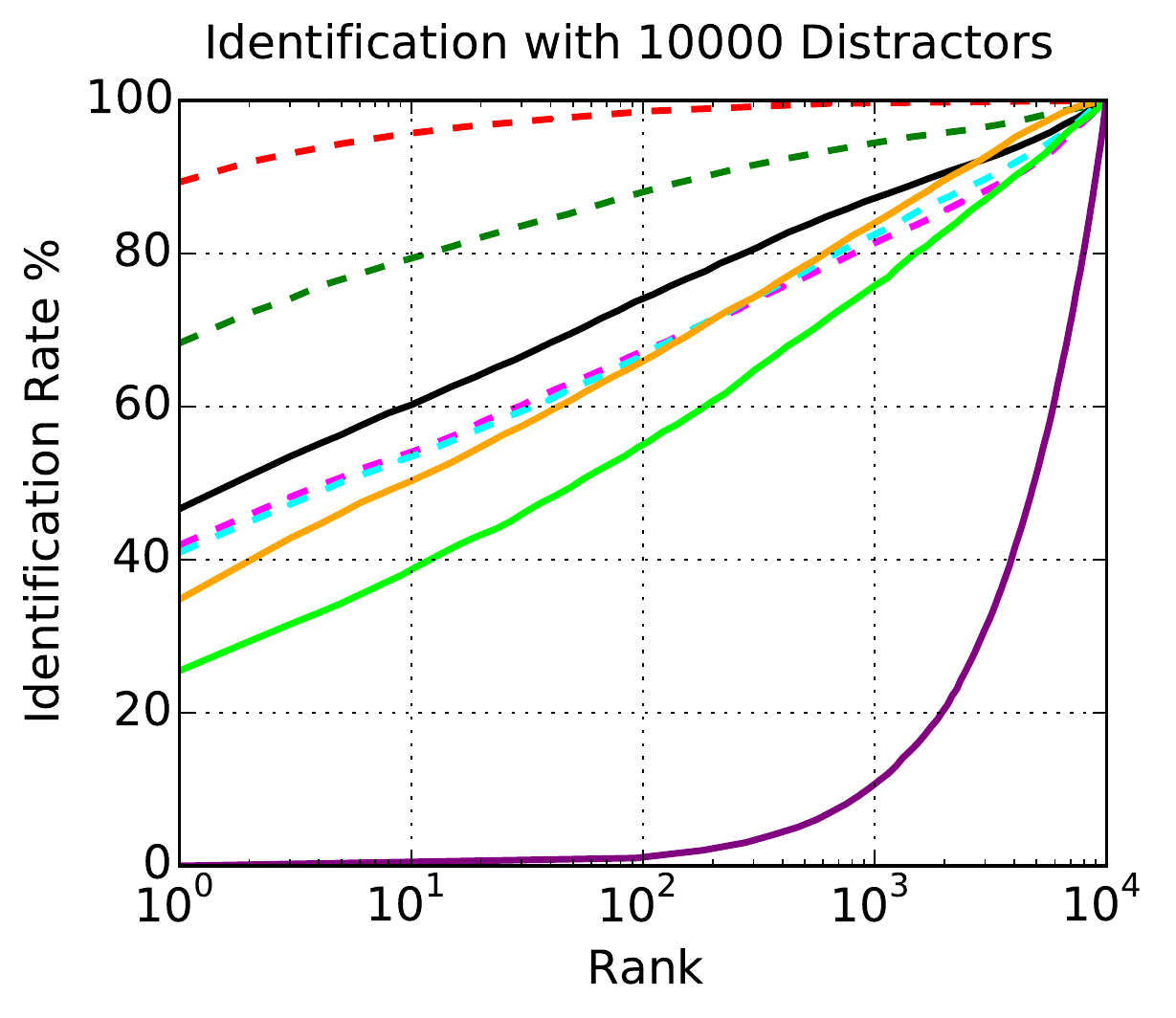} &
									\includegraphics[width=0.25\linewidth, valign=t]{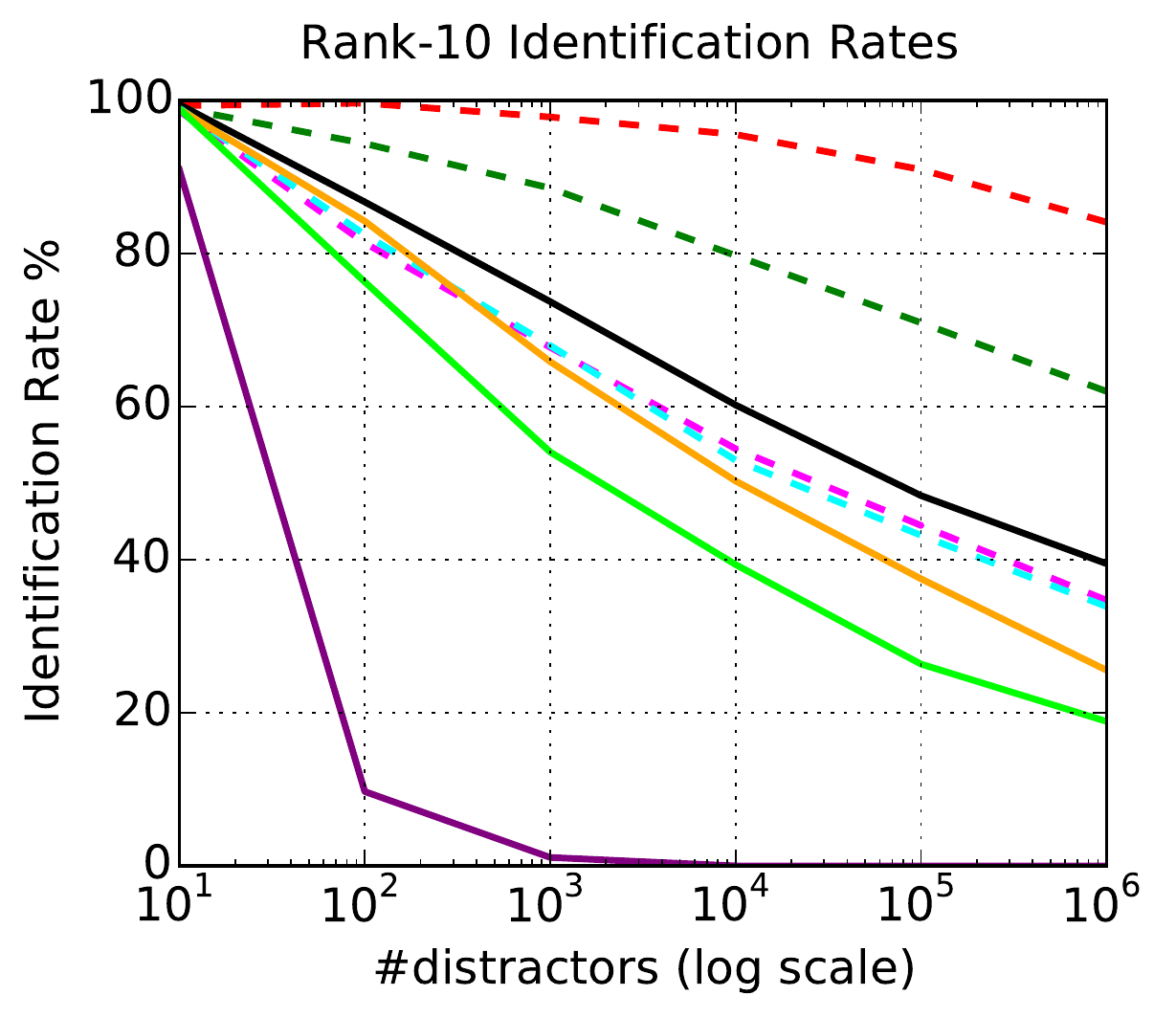}\\
							&		{\footnotesize (d) FGNET + 1M} & {\footnotesize (e) FGNET + 10K}   & {\footnotesize (f) FGNET + rank-10} 
						\end{tabular}
						
						\caption{\textbf{Identification} performance for all methods with (a,d) 1M distractors and (b,e) 10K distractors, and (c,f) rank-10 for both probe sets.  Fig.~\ref{fig:teaser} also shows  rank-1 performance as a function of number of distractors on both probe sets. }
						\label{fig:dataset_size_cmc}
					\end{figure*}


				\begin{figure*}
					\centering
					\begin{tabular}{ccccc}
						\includegraphics[width=0.18\linewidth]{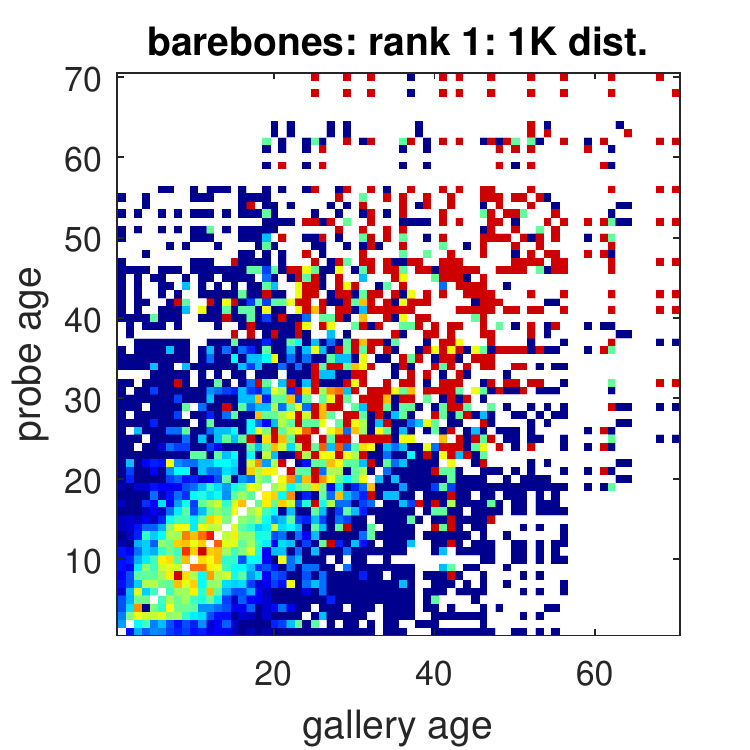}	 &				\includegraphics[width=0.18\linewidth]{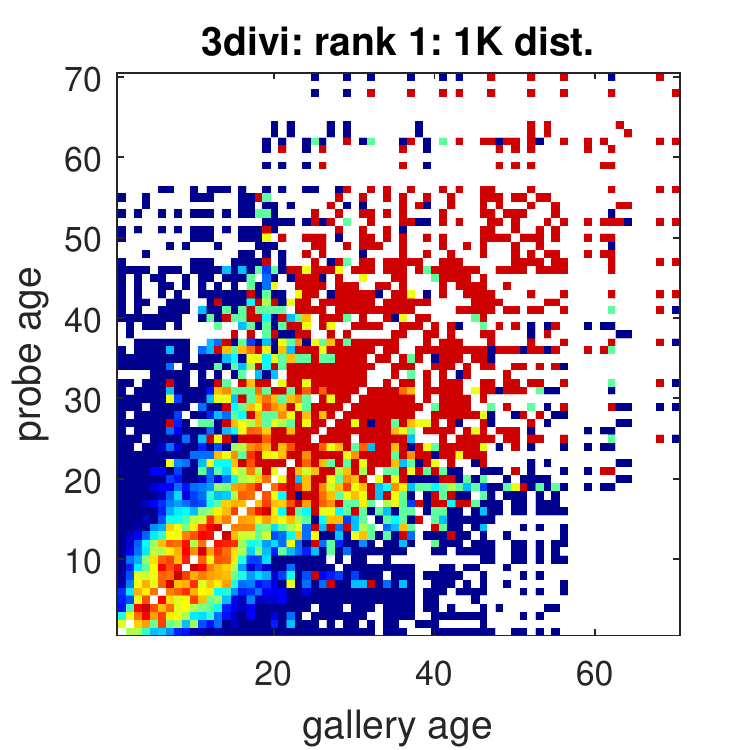} &
						\includegraphics[width=0.18\linewidth]{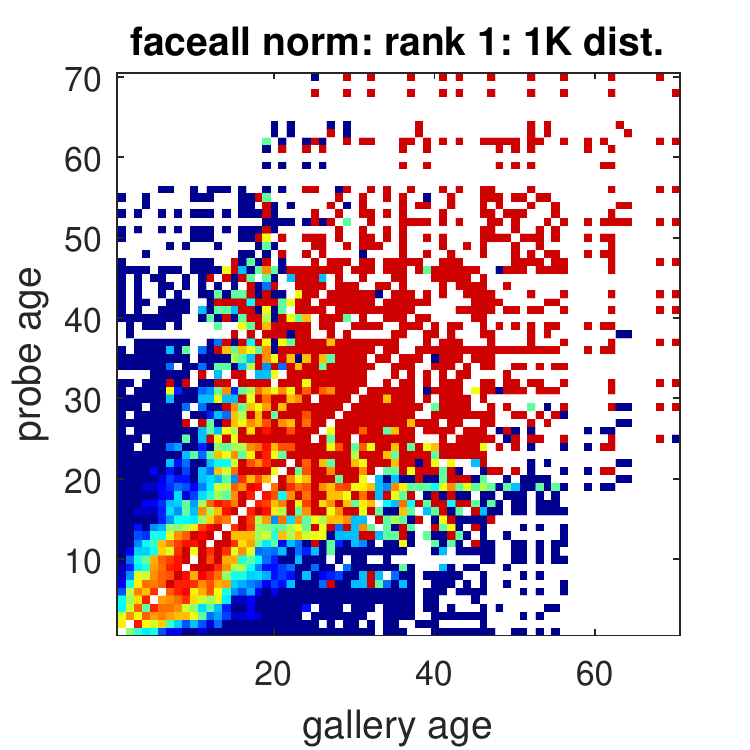} &
						\includegraphics[width=0.18\linewidth]{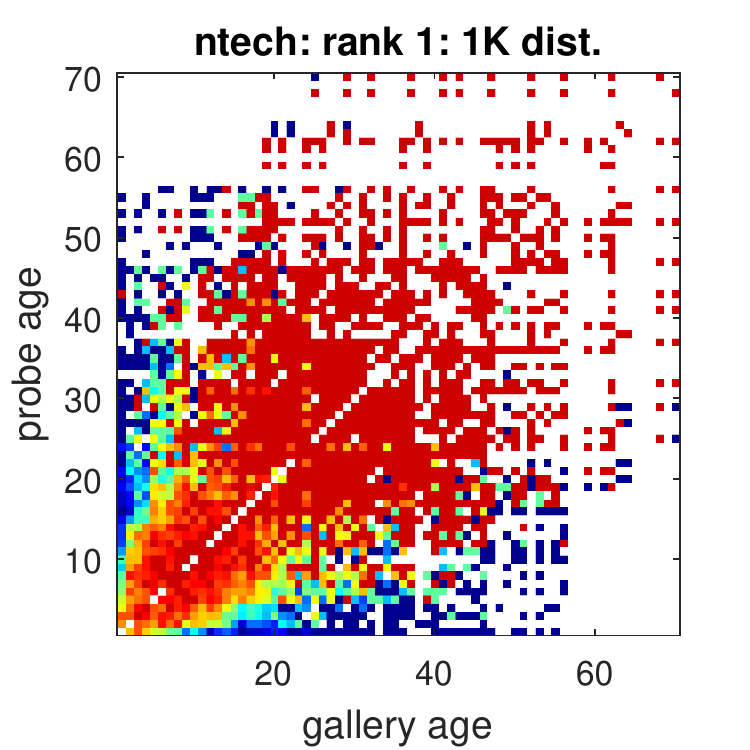} & 
							\includegraphics[width=0.18\linewidth]{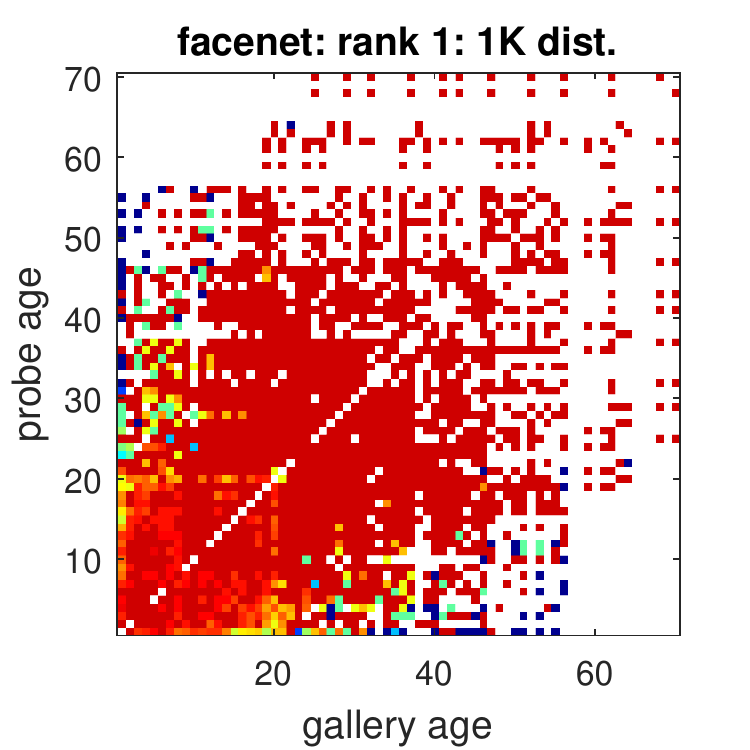}
						\\
						\includegraphics[width=0.18\linewidth]{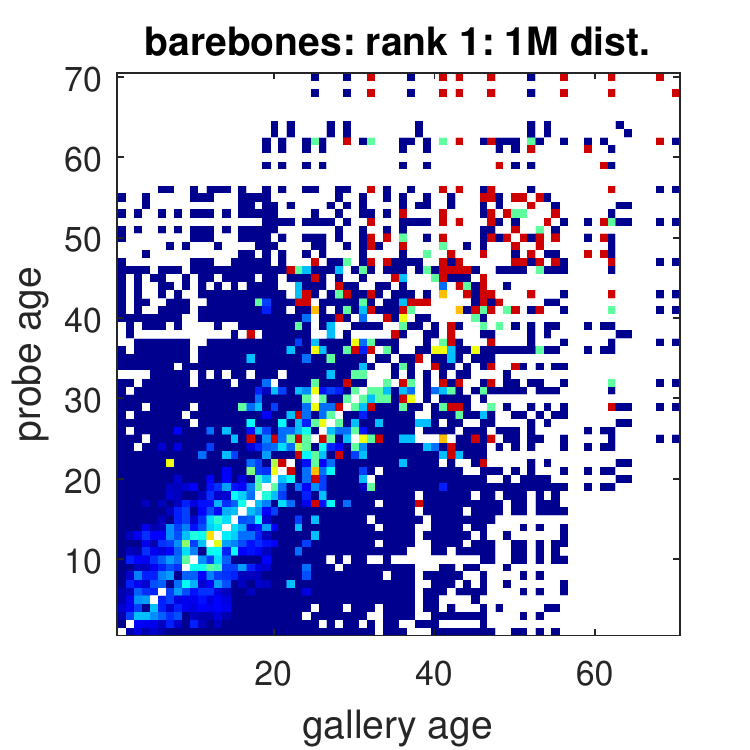} &
						\includegraphics[width=0.18\linewidth]{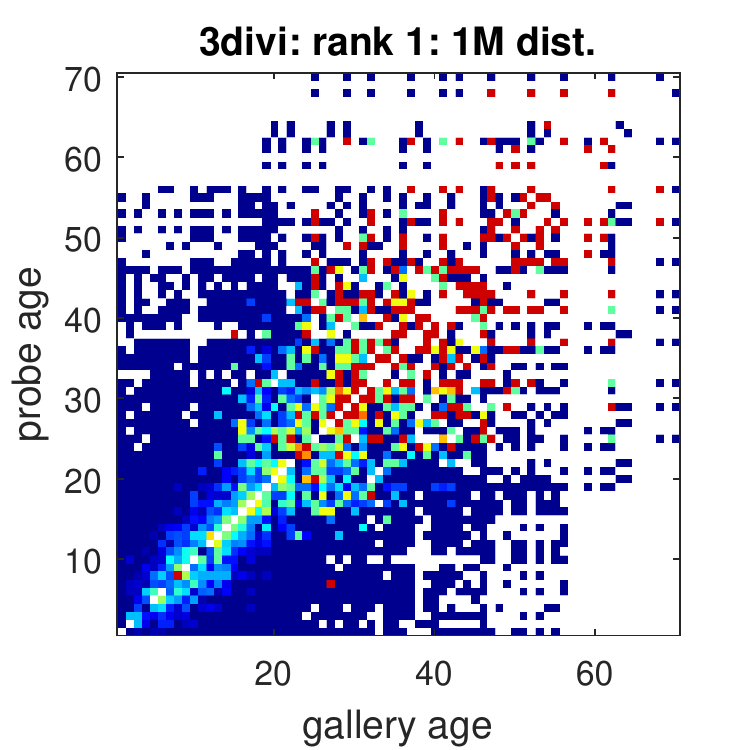} &
						\includegraphics[width=0.18\linewidth]{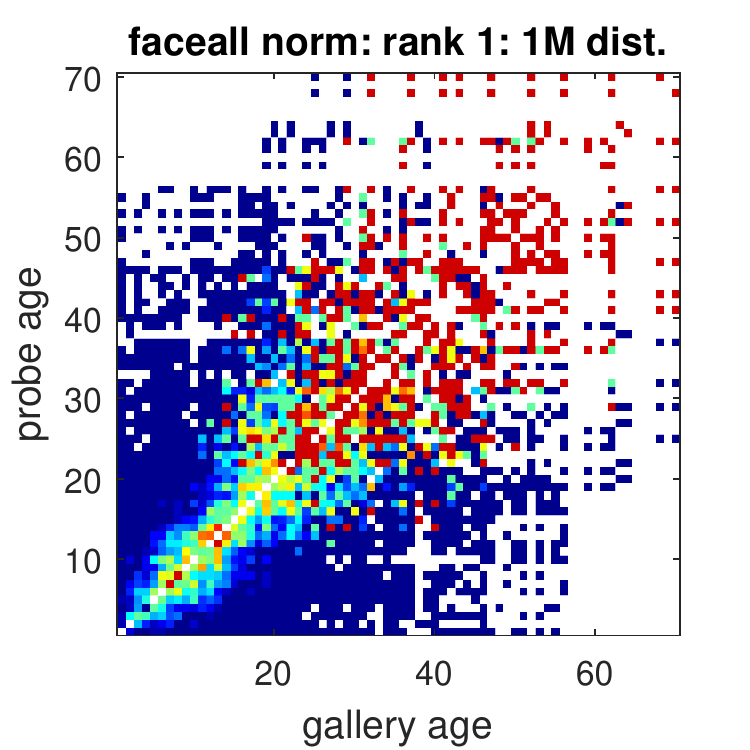} &
						\includegraphics[width=0.18\linewidth]{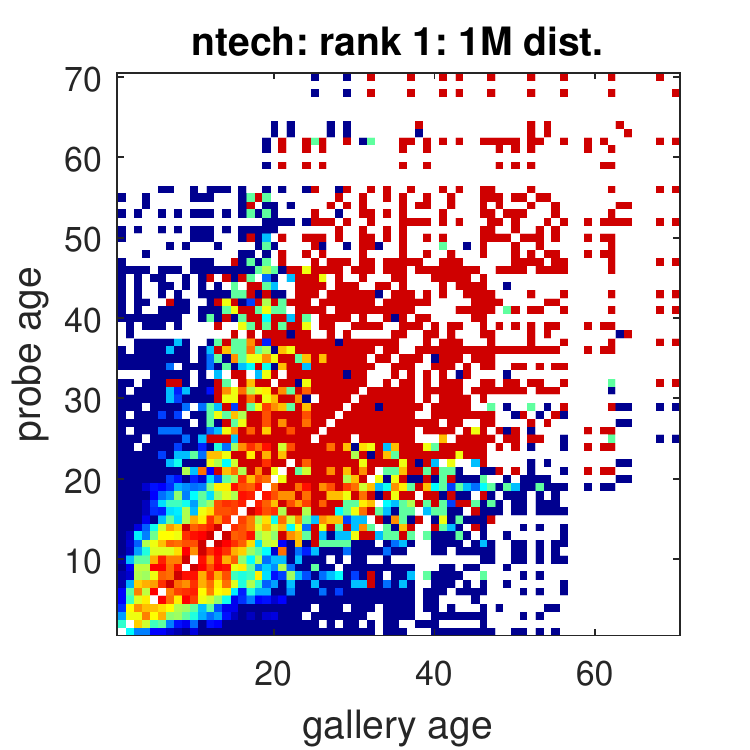} &
                        \includegraphics[width=0.18\linewidth]{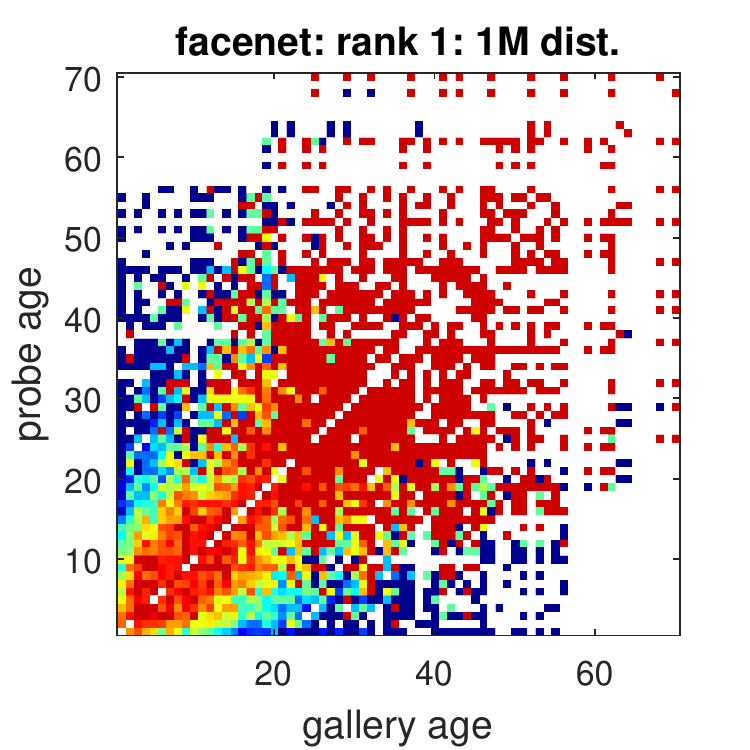}
					\end{tabular}
					
					\caption{Analysis of rank-1  identification with respect to varying \textbf{ages} of gallery and probe.  Columns represent  five algorithms, rows 1K and 1M distractors.   X-axis represents a person's age in the gallery photo and Y-axis age in the probe. The colors represent identification accuracy going from 0(=blue)--none of the true pairs  were matched to 1(=red)--all possible combinations of probe and gallery were matched per probe and gallery ages.  Lower scores on left and bottom indicate worse performance on children, and higher scores along the diagonal indicate that methods are better at matching across small age differences.}
					\label{fig:age}
				\end{figure*}

				\begin{figure}
					\begin{center}
						\includegraphics[width=0.32\linewidth]{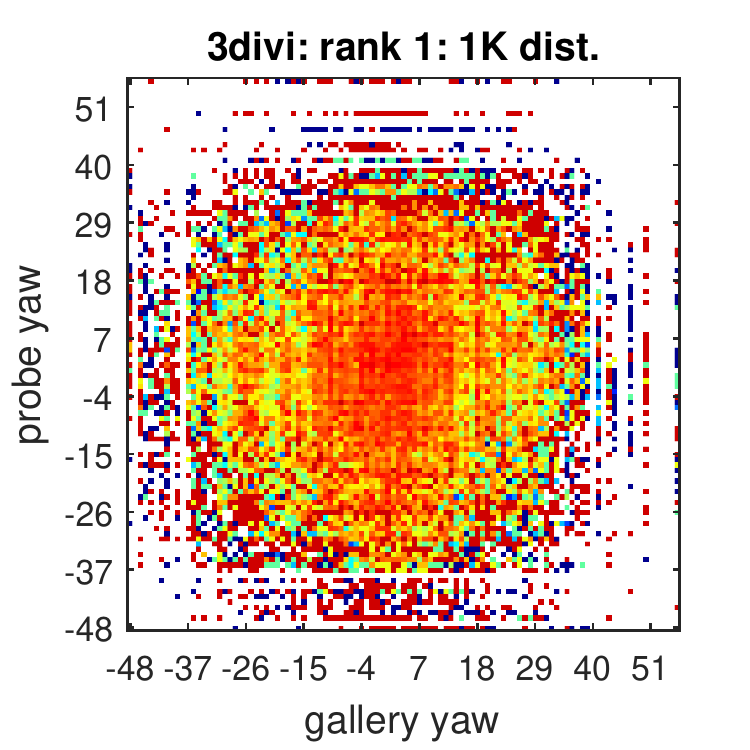}
						\includegraphics[width=0.32\linewidth]{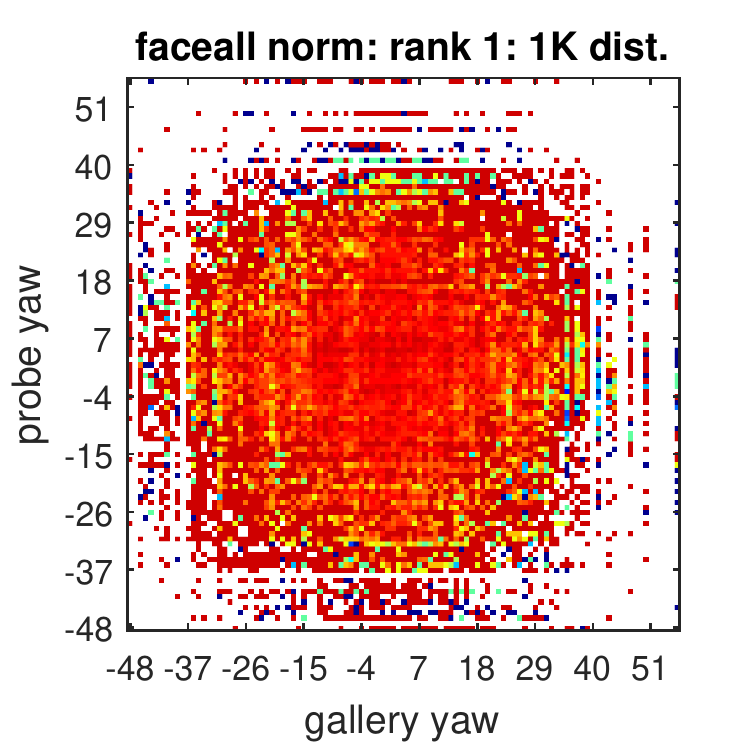}											\includegraphics[width=0.32\linewidth]{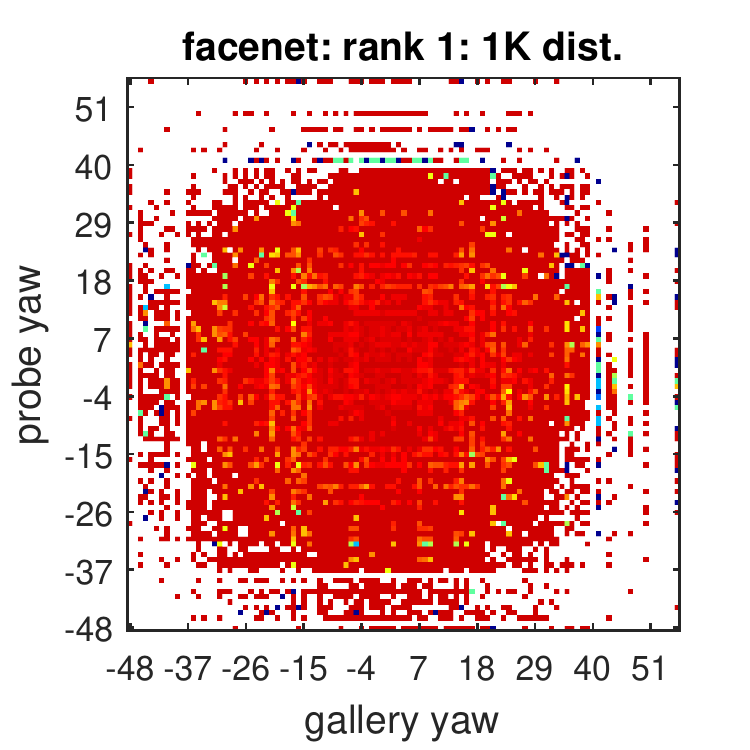}\\
						\includegraphics[width=0.32\linewidth]{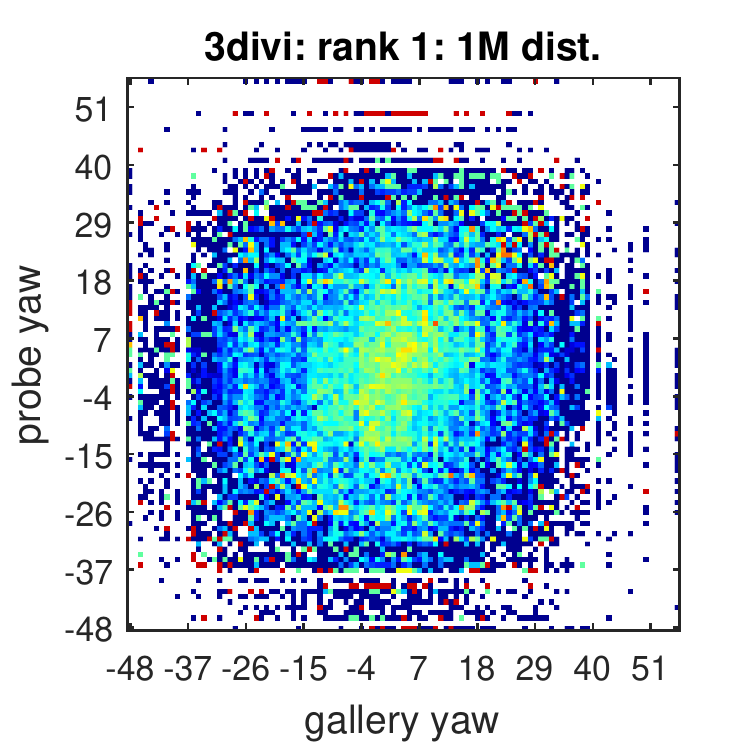}
						\includegraphics[width=0.32\linewidth]{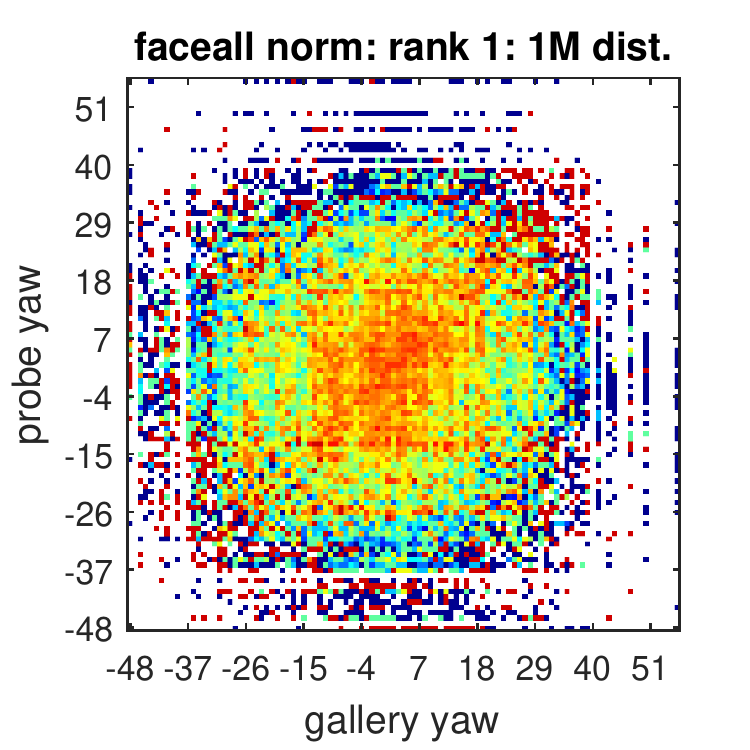}
						\includegraphics[width=0.32\linewidth]{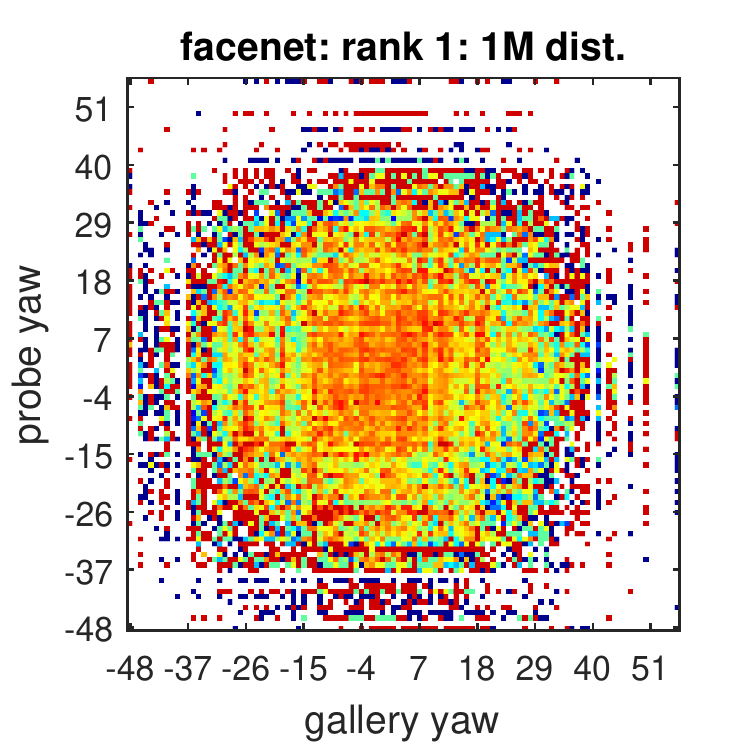}
					\end{center}
					\caption{Analysis of rank-1  identification with varying \textbf{poses} of gallery and probe, for three algorithms.   Top:  1K distractors, Bottom:  1M distractors.
					The colors represent identification accuracy going from 0 (blue) to 1 (red), where 0 means that none of the true pairs  were matched, and 1 means that all possible combinations of probe and gallery were matched per probe and gallery ages. White color indicates combinations of poses that did not exist in our test set.  We can see that evaluation at scale (bottom) reveals large differences in performance, which is not visible at smaller scale (top): frontal poses and smaller difference in poses is easier for identification.  }
					\label{fig:pose}
				\end{figure}

			
				\textbf{Verification results.} Fig.~\ref{fig:verf} shows results of the verification experiment for our two probe sets, (a) and (b) show results on FaceScrub and (c) and (d) on FGNET.  We present results of one random fixed set of distractors per gallery size (see the other two in the supplementary). 
				
				We see that, for FaceScrub, at lower false accept rates the performance of algorithms drops by about 40\% on average. FaceNet and FaceN lead with  only about 15\%. Interestingly, FaceN that was trained on 18M photos is able to achieve comparable results to FaceNet that was trained on 500M. Striving to perform well at low false accept rate  is important with large datasets. Even though the chance of a false accept on the small benchmark is acceptable, it does not scale to even moderately sized galleries. Results at LFW are typically reported at equal error rate which implies false accept rate of 1\%-5\% for top algorithms, while for a large set like MegaFace, only FAR of $10^-5$ or $10^-6$ is meaningful. 
					
					For FGNET the drop in performance is striking--about 60\% for everyone but FaceNet, the latter achieving impressive performance across the board. One factor may be the type of training used by different groups (celebrities vs. photos across ages, etc.).

				Verification rate stays similar when scaling up the gallery, e.g., compare (a) and (b).  The intuition is that verification rate is normalized by the size of the dataset, so that if  a probe face is matched incorrectly to 100 other faces in a 1000 faces dataset,  assuming uniform distribution of the data, the rate will stay the same, and so in a dataset of a million faces one can expect to find 10,000 matches at the same false accept rate (FAR).  
				
		\textbf{Identification results.}   In Fig.~\ref{fig:dataset_size_cmc} we show the performance with respect to different ranks, i.e., rank-1 means that the correct match got the best score from the whole database, rank-10 that the correct match is in the first 10 matches, etc.  (a,b,c) show performance for the FaceScrub dataset and (d,e,f) for FGNET. We observe that rates drop for all algorithms as the  gallery size gets larger. This is visualized in Fig.~\ref{fig:teaser}, the actual accuracies are in Fig.~\ref{fig:rank1numbers}.  The curves also suggest that when evaluated on more than 1M distractors (e.g., 100M), rates will be even lower.    Testing on  FGNET \textbf{at scale} reveals a dramatic performance gap. All algorithms perform much worse, except for FaceNet that has a similar performance to its results on FaceScrub.

		\textbf{Training set size.} Dashed lines in all plots represent algorithms that were trained on data larger than 500K photos and 20K people. We can see that these generally perform better than others.

		\begin{figure}
			\begin{tabular}{c}
				\includegraphics[width=1\linewidth]{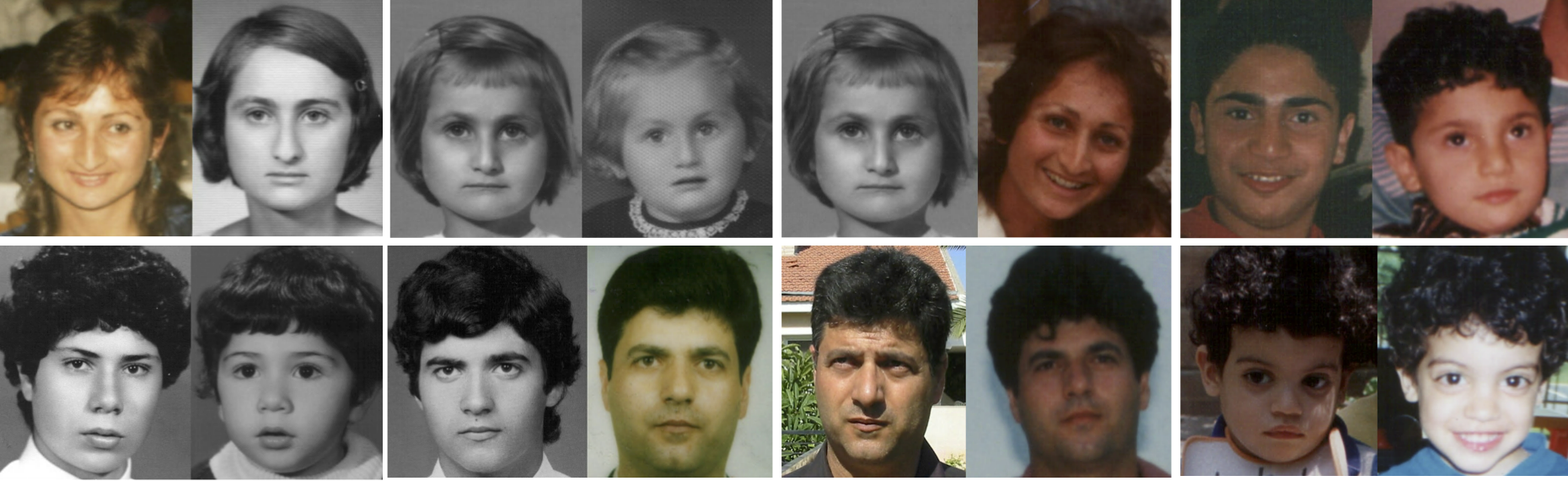} \\
			{\small 	(a) true positives}\\
				\includegraphics[width=1\linewidth]{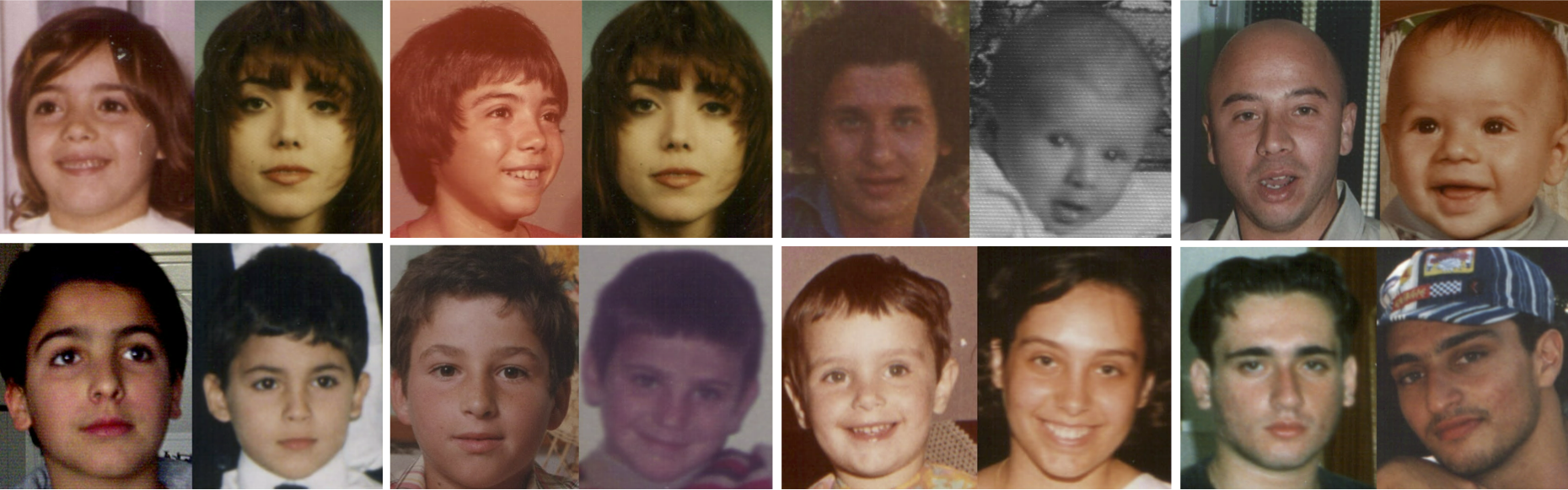}\\
			{\small  (b) false negatives}
			\end{tabular}
			\caption{Examples pairs from FGNET using top performing FaceNet with 10 distractors. 
			Each consecutive left-right pair of images is the same person.
			All algorithms match better with smaller age differences.  }
			\label{fig:agematches}
		\end{figure}

	\textbf{Age.} Evaluating performance using FGNET as a probe set also reveals a major drop in performance for most algorithms when attempting to match across differences in age.  We present a number of results: Fig.~\ref{fig:age} shows differences in performances with varying age across gallery and probe.  Each column represents a different algorithm, rows present results for 1K and 1M distractors.  Red colors indicate higher identification rate, blue lower rate.  We make two key observations: 1)  algorithms perform better when the difference in age between gallery and probe is small (along the diagonal), and 2)  adults are more accurately matched than children, at scale. 
	Fig.~\ref{fig:agematches}  shows examples of  matched pairs (true positives and false negatives) using FaceNet and 10 distractors.  Notice that false negatives have a bigger age gap relative to true positives. It is impressive, however, that the algorithm was able to these and many other true positives, given the variety in lighting, pose, and quality of the photo in addition to age changes.

		\textbf{Pose.}  Fig.~\ref{fig:pose} evaluates error in recognition as a function of  difference in yaw between the probe and gallery. The results are normalized by the total number of pairs for each pose difference.  We can see that recognition accuracy depends strongly on pose and this difference is revealed more prominently when evaluated at scale.  Top row show results of three different algorithms (representative of others) with 1K distractors. Red colors indicate that identification is very high and mostly independent of pose. However, once evaluated at scale (bottom row) with 1M distractors we can see that variation across algorithms as well as poses is more dramatic. Specifically, similar poses identified better, and more frontal (center of the circle) poses are easier to  recognize.

\section{Discussion}

An ultimate face recognition algorithm should perform with billions of people in a dataset. While testing with billions is still challenging, we have done the first step and created a benchmark of a million faces. MegaFace is available to researchers and we presented results from state of the art methods. Our key discoveries are 1) algorithms' performance degrades given a large gallery even though the probe set stays fixed,  2) testing at scale allows to uncover the differences across algorithms (which at smaller scale appear to perform similarly), 3) age differences across probe and gallery are still more challenging for recognition.   We will keep maintaining and updating the MegaFace benchmark online, as well as, create more challenges in the future.  Below are topics we think are exciting to explore.  First, we plan to release all the detected faces from the 100M Flickr dataset. Second,  companies like Google and Facebook have a  head start due to availability of enormous amounts of data. We are interested to level the playing field and provide large \textbf{training} data to the research community that will be assembled from our Flickr data. Our dataset will be separated to testing and training sets for fair evaluation and training.   Finally, the significant number of high resolution faces in our Flickr database 
will also allow  to explore resolution in more depth. Currently, it  is mostly untouched topic in face recognition literature due to lack of data. 

\vspace{.1in}

\textbf{Acknowledgments} We are grateful to the early challenge participants that allowed a very quick turn out and provided great feedback on the baseline code and data.  We thank  Samsung and Google for the generous support of this research. 

\newpage

{\small
\bibliographystyle{ieee}
\bibliography{paper_cvpr16_withnames_withsup}
}

\newpage
\section{Supplementary material}

We have created three different random subsets of MegaFace for each of the distractor sizes (10,100,1K,10K,100K,1M) and reran all the algorithms with each of the set for each of the probe sets (FaceScrub and FGNET). We present the results in the below figures. Set \#1 is the one presented in the main paper. 

\begin{figure*}
	\begin{tabular}{ccc}
		\includegraphics[width=.17\linewidth, valign=t]{plots16/ll.pdf} & 
		\includegraphics[width=.31\linewidth, valign=t]{plots16/graphs/set_1/facescrub_rank-1_cmbnd_set_1.pdf} &
		\includegraphics[width=.31\linewidth, valign=t]{plots16/graphs/set_1/fgnet_rank-1_cmbnd_set_1.pdf} \\
		\includegraphics[width=.17\linewidth, valign=t]{plots16/ll.pdf} & 
		\includegraphics[width=.31\linewidth, valign=t]{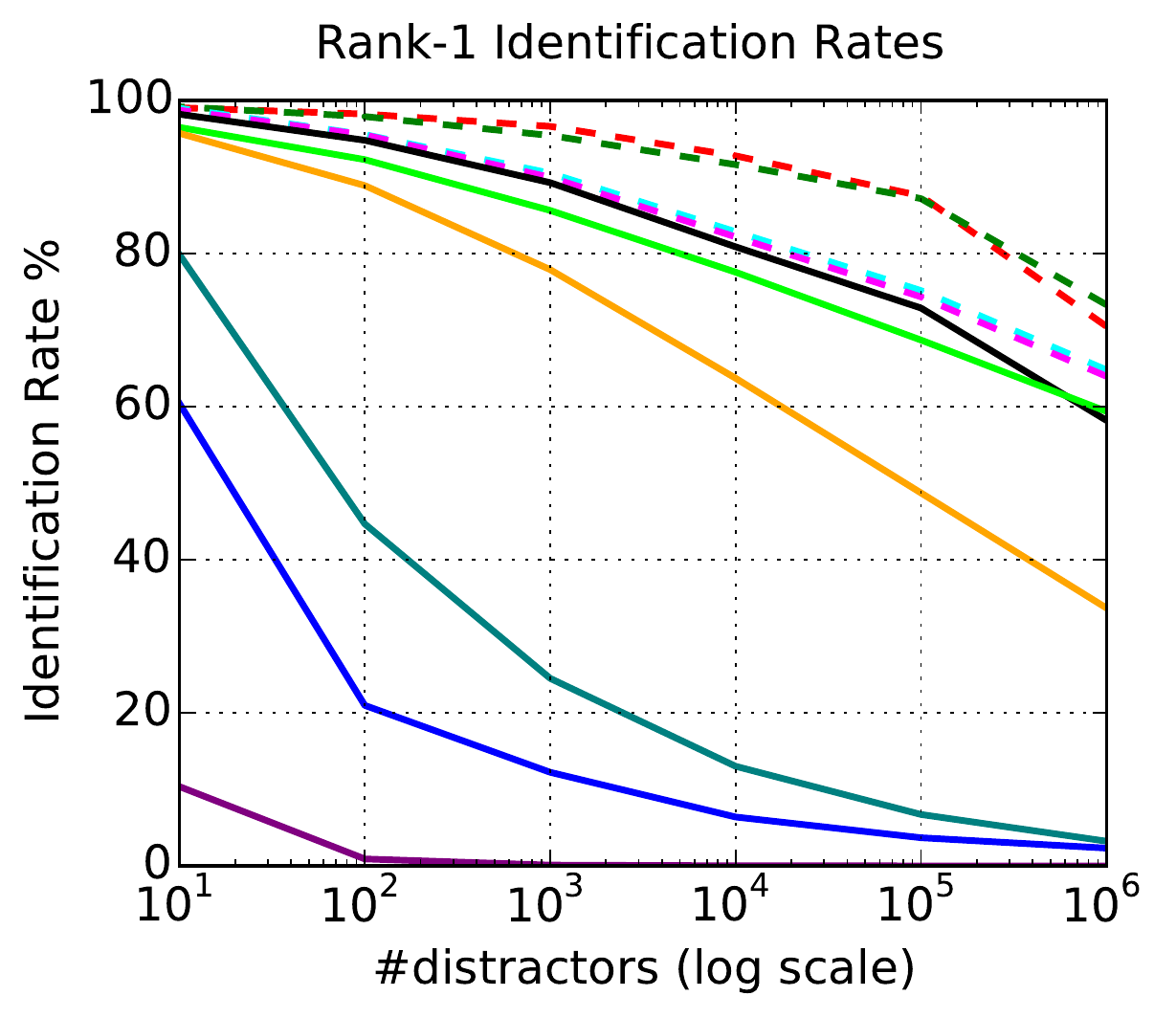} &
		\includegraphics[width=.31\linewidth, valign=t]{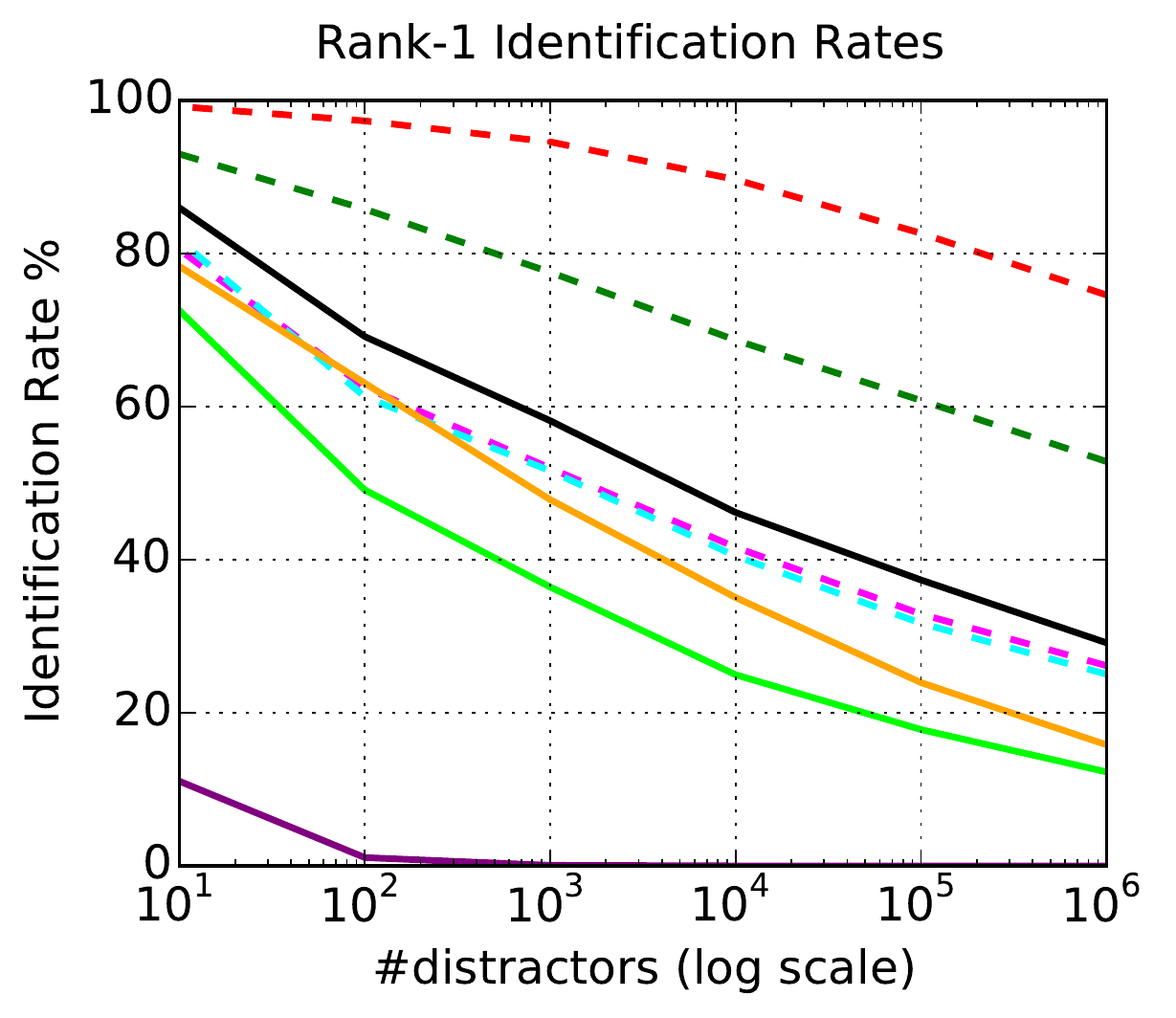} \\
		\includegraphics[width=.17\linewidth, valign=t]{plots16/ll.pdf} & 
		\includegraphics[width=.31\linewidth, valign=t]{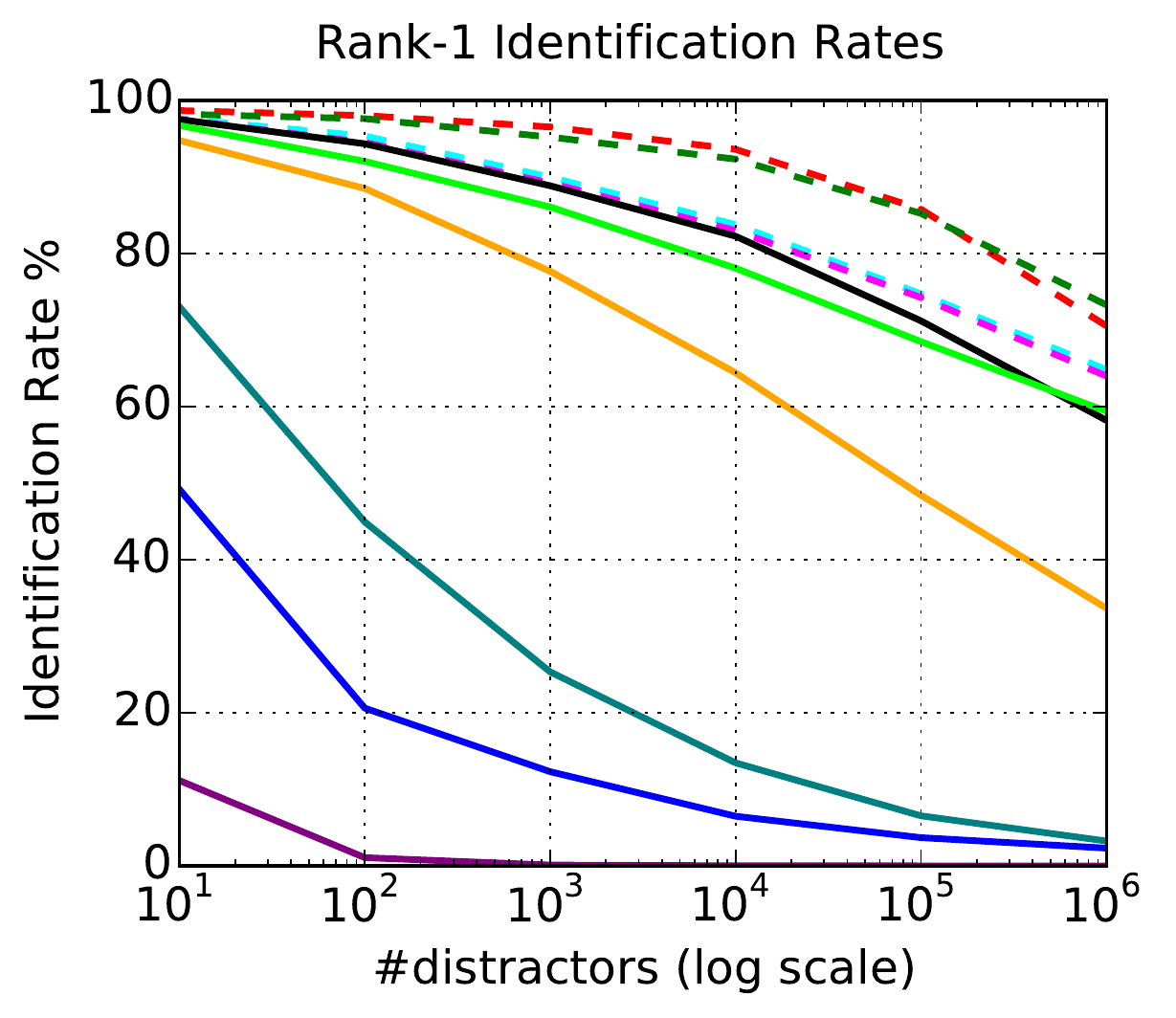} &
		\includegraphics[width=.31\linewidth, valign=t]{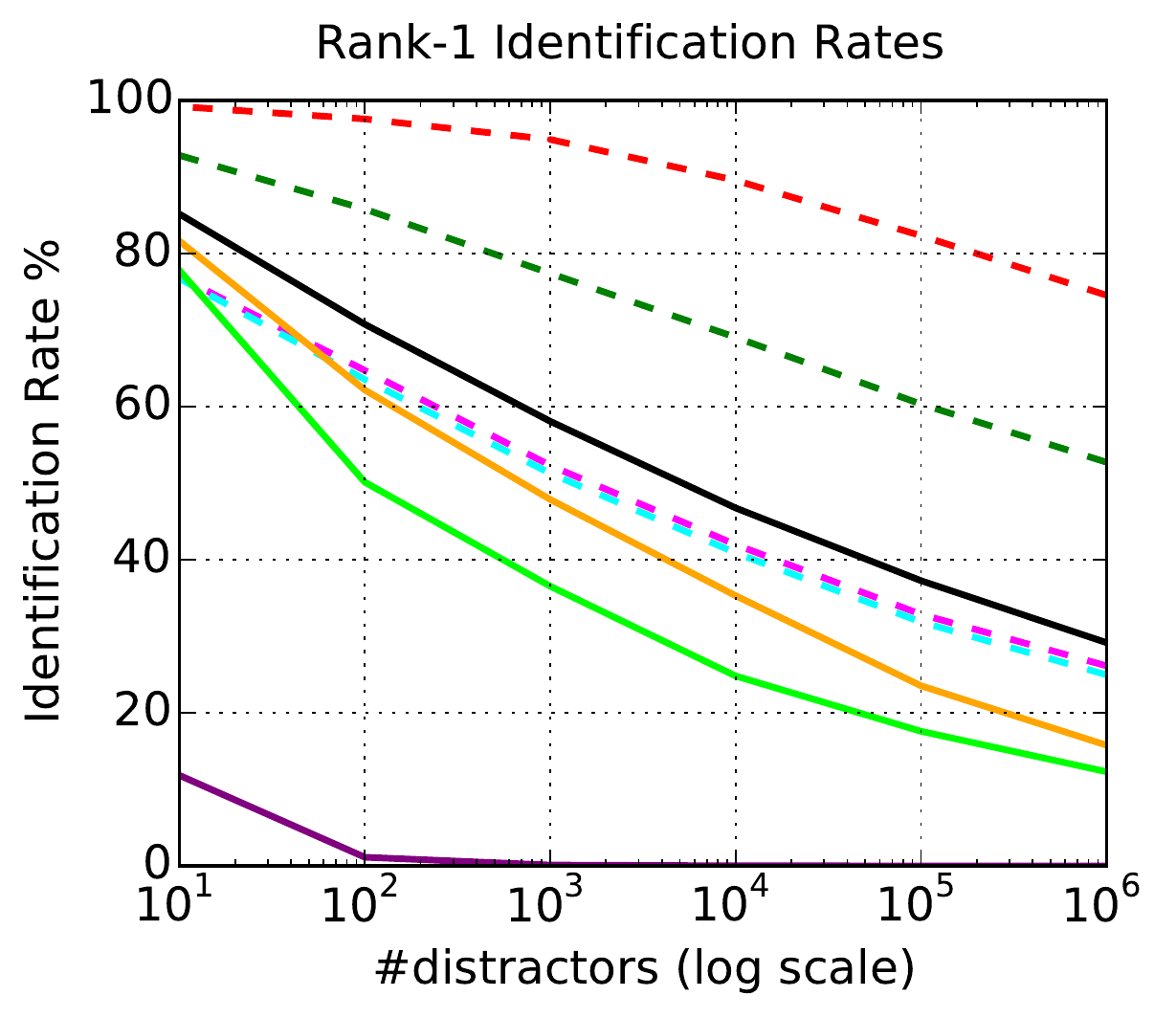} \\
		& {\footnotesize (a) FaceScrub + MegaFace} & {\footnotesize (b)  FGNET + MegaFace}\\
		\vspace{.05in}
	\end{tabular}
	\caption{Sets 1--3 (each row represents a different random gallery set). The MegaFace challenge evaluates  identification and verification as a function of  increasing  number of gallery distractors (going from 10 to 1 Million).  We use two different probe sets (a) FaceScrub--photos of celebrities, (b) FGNET--photos with a large variation in age per person. We present rank-1 identification of  state of the art algorithms that participated in our challenge. On the left side of  each plot is current major benchmark LFW scale (i.e., 10 distractors, see how all the top algorithms are clustered above 95\%).  On the right is mega-scale (with a million distractors).  Observe that  rates drop with increasing numbers of distractors, even though the probe set is fixed, and that algorithms trained on larger sets (dashed lines) generally perform better.  }
	\label{fig:rank1}
\end{figure*}
	
	\begin{figure*}
		\centering
		\begin{tabular}{ccc}
			\includegraphics[width=.15\linewidth, valign=t]{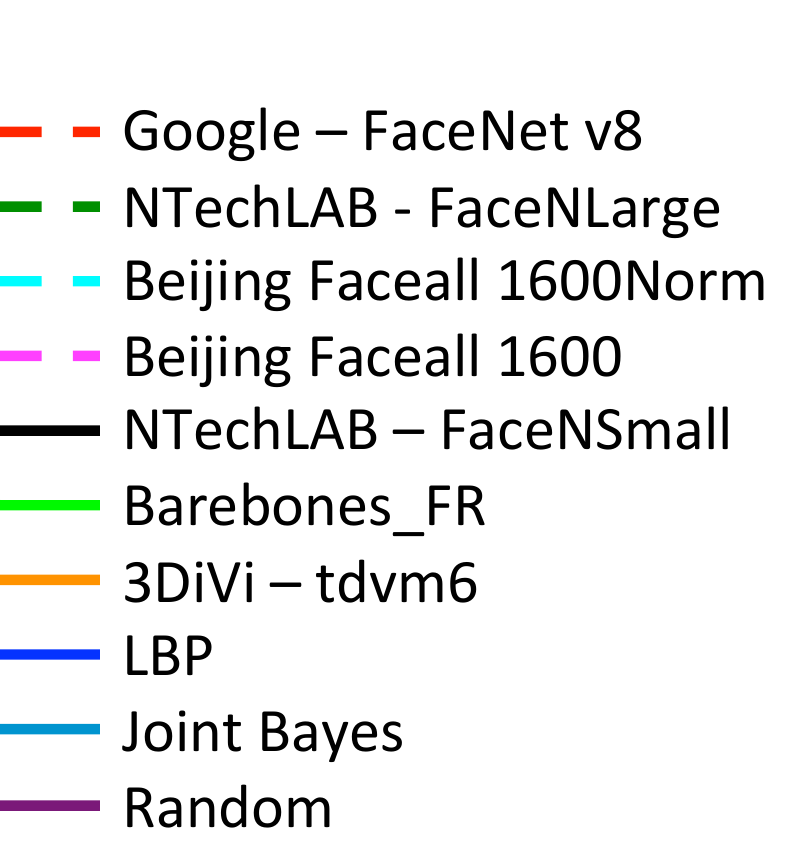} & 
			\includegraphics[width=0.27\linewidth, valign=t]{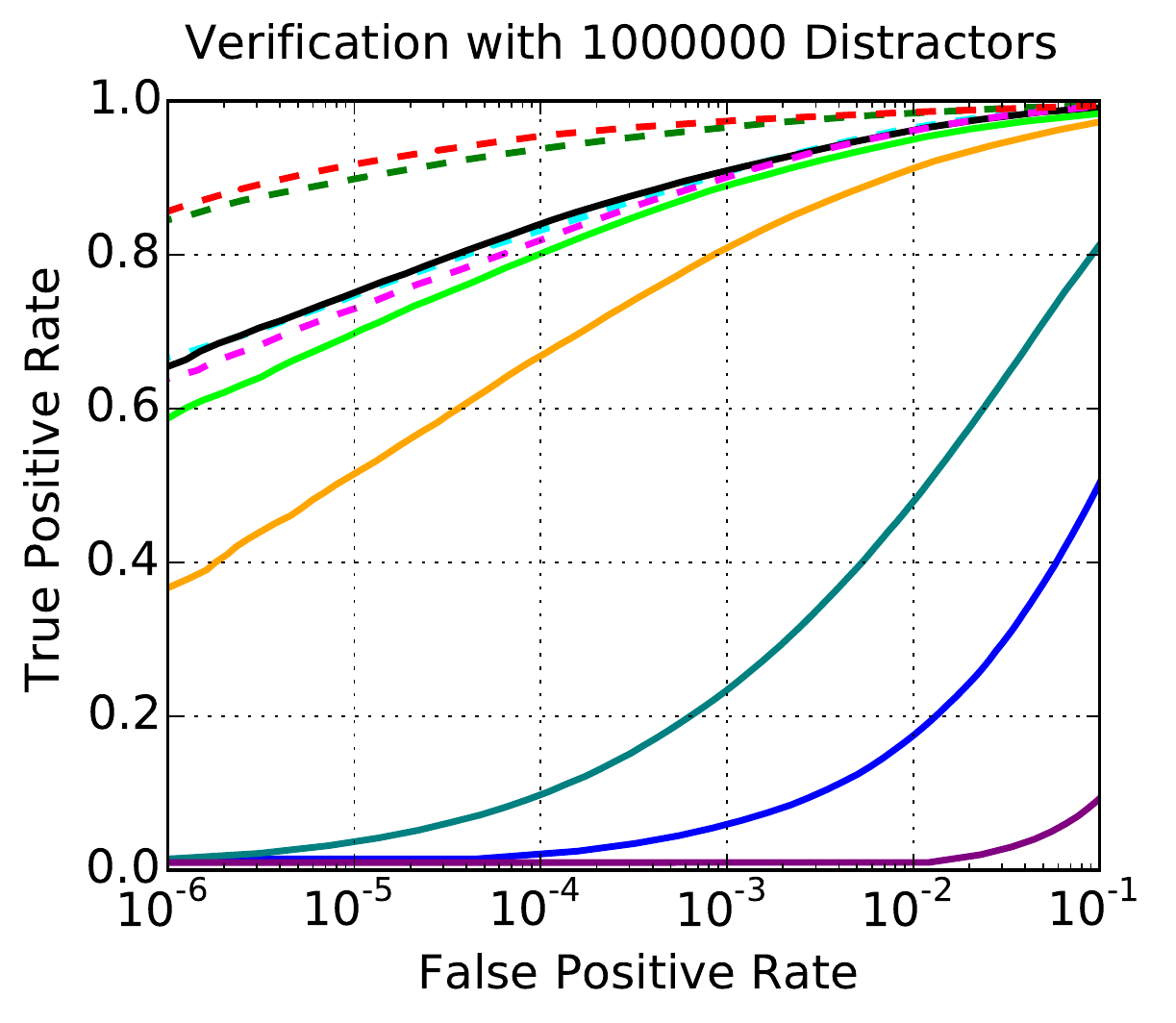} & 
			\includegraphics[width=0.27\linewidth, valign=t]{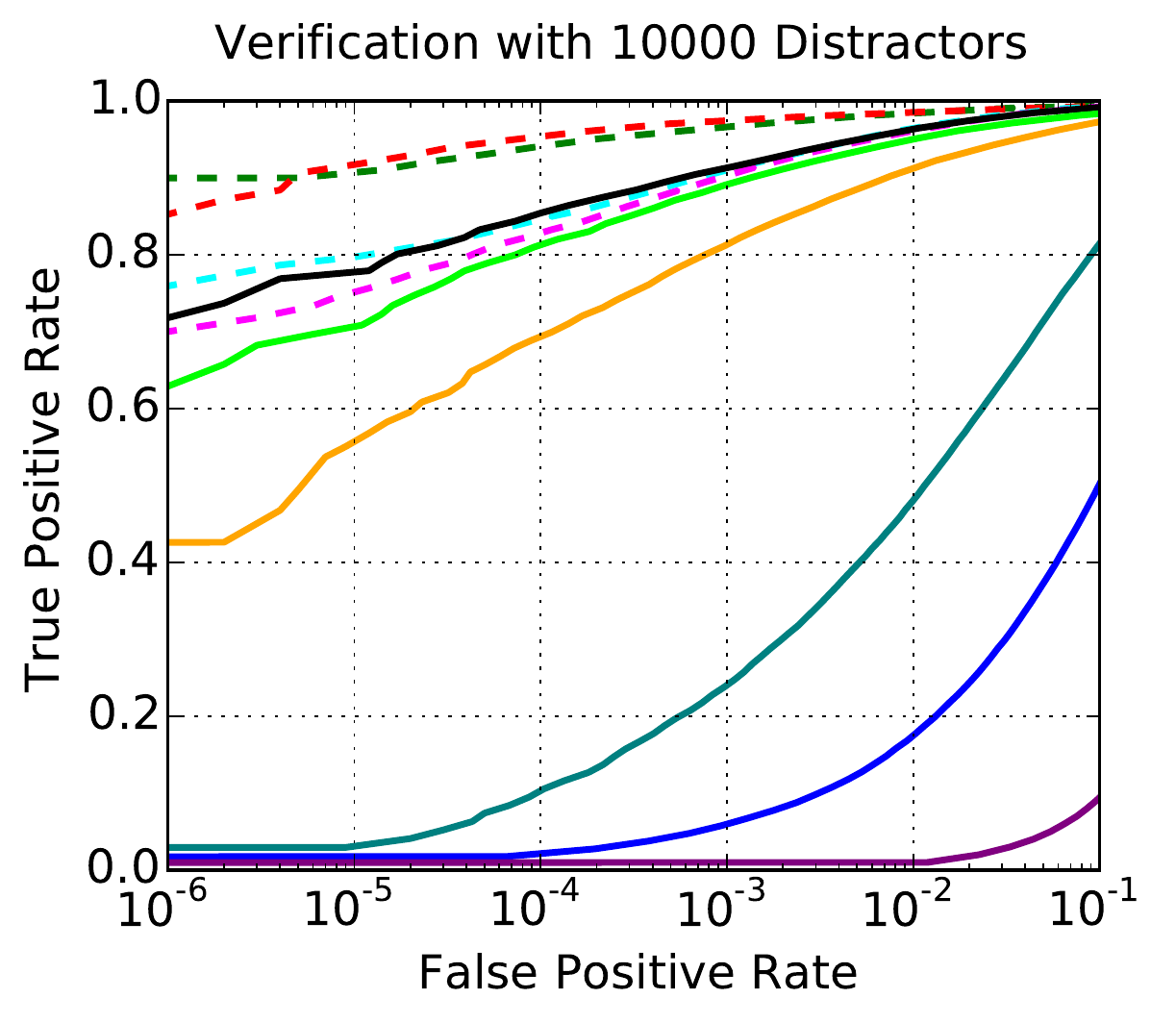} \\
			&{\footnotesize  (a) FaceScrub + 1M} & {\footnotesize (b) FaceScrub + 10K}  \\
			\includegraphics[width=.15\linewidth, valign=t]{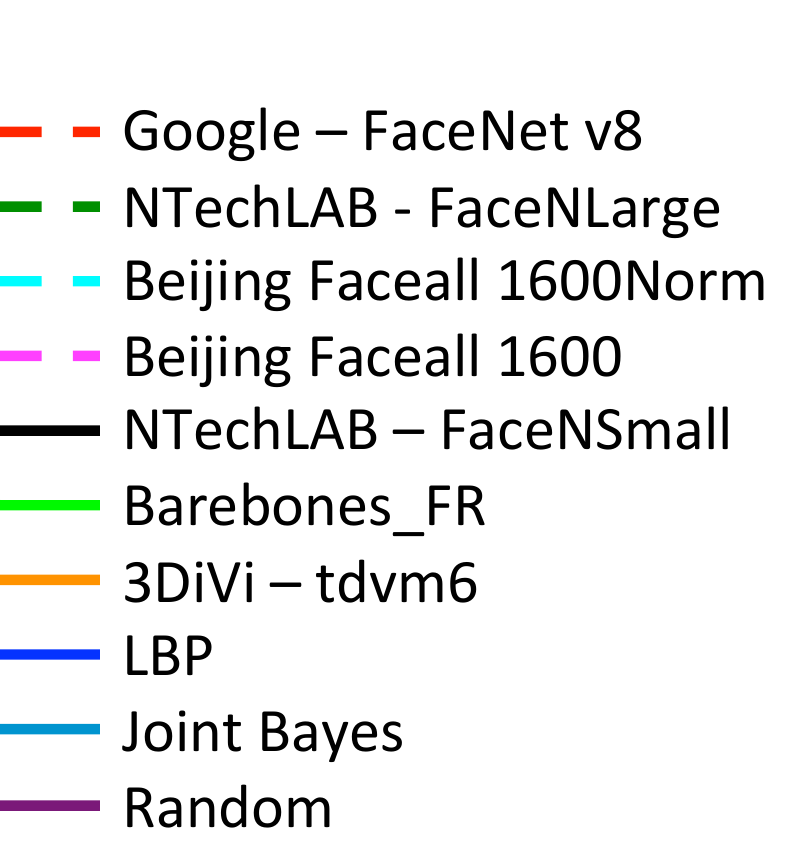} & 
			\includegraphics[width=0.27\linewidth, valign=t]{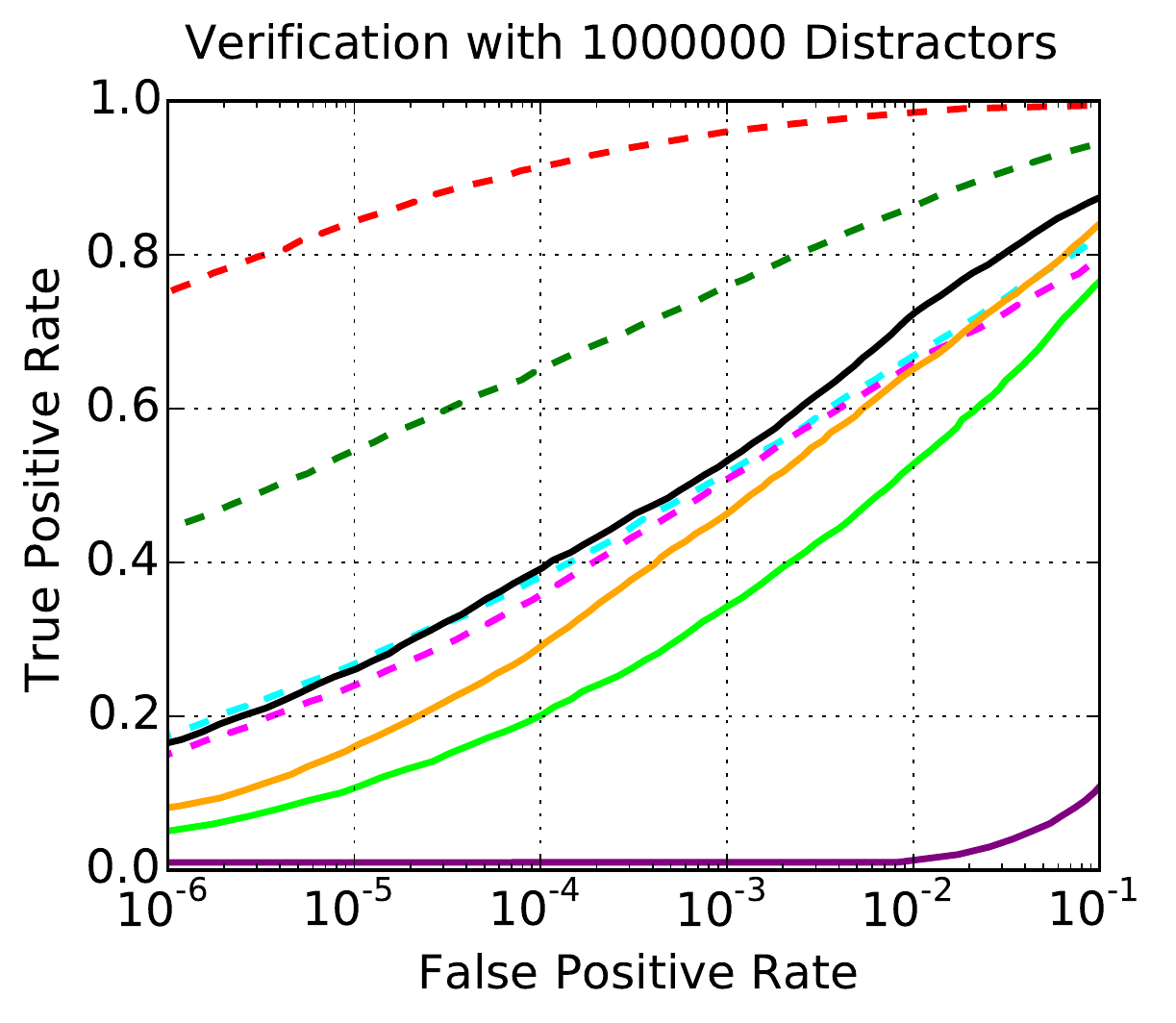} & 
			\includegraphics[width=0.27\linewidth, valign=t]{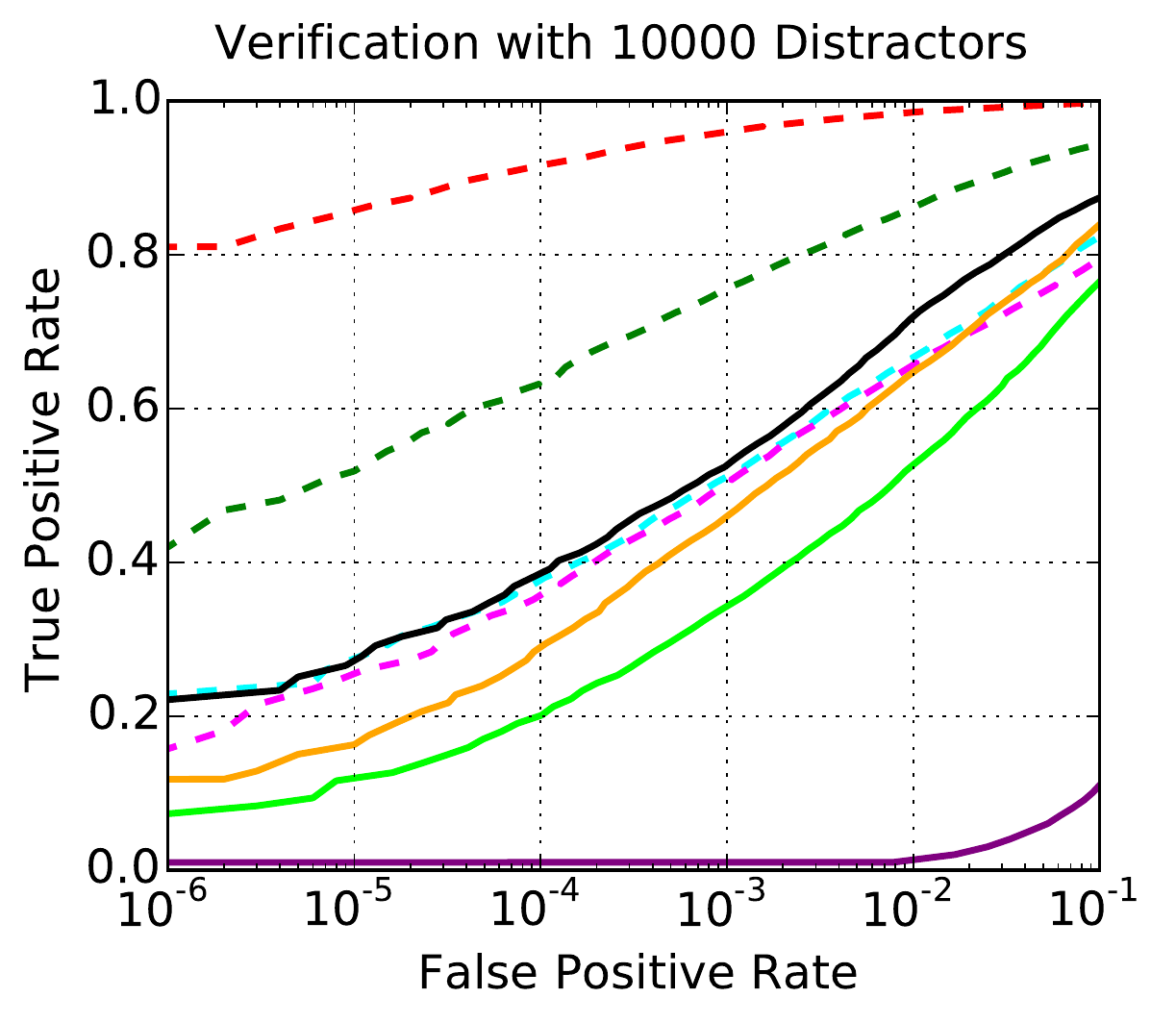} \\
			&{\footnotesize  (c) FGNET + 1M} & {\footnotesize (d) FGNET + 10K}  \\
		\end{tabular}
		\caption{\textbf{Verification (random gallery set 1)} performance with (a,c) 1 Million and (b,d) 10K distractors on both probe sets. Note the performance at low false accept rates (left side of each plot). }\label{fig:verf}
		\label{fig:dataset_size_roc} 
	\end{figure*}
		
			\begin{figure*}
				\centering
				\begin{tabular}{ccc}
					\includegraphics[width=.15\linewidth, valign=t]{plots16/ll.pdf} & 
					\includegraphics[width=0.27\linewidth, valign=t]{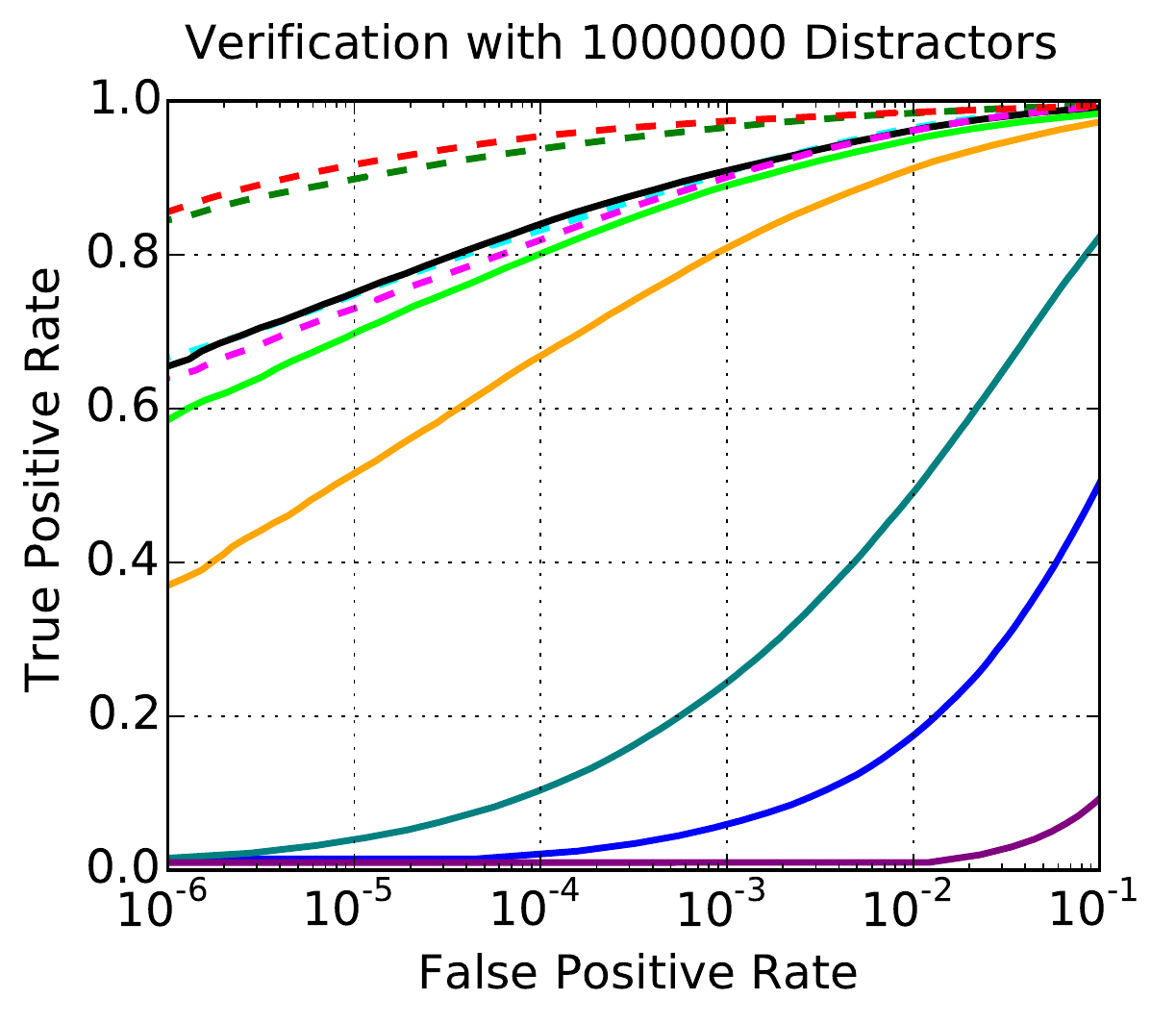} & 
					\includegraphics[width=0.27\linewidth, valign=t]{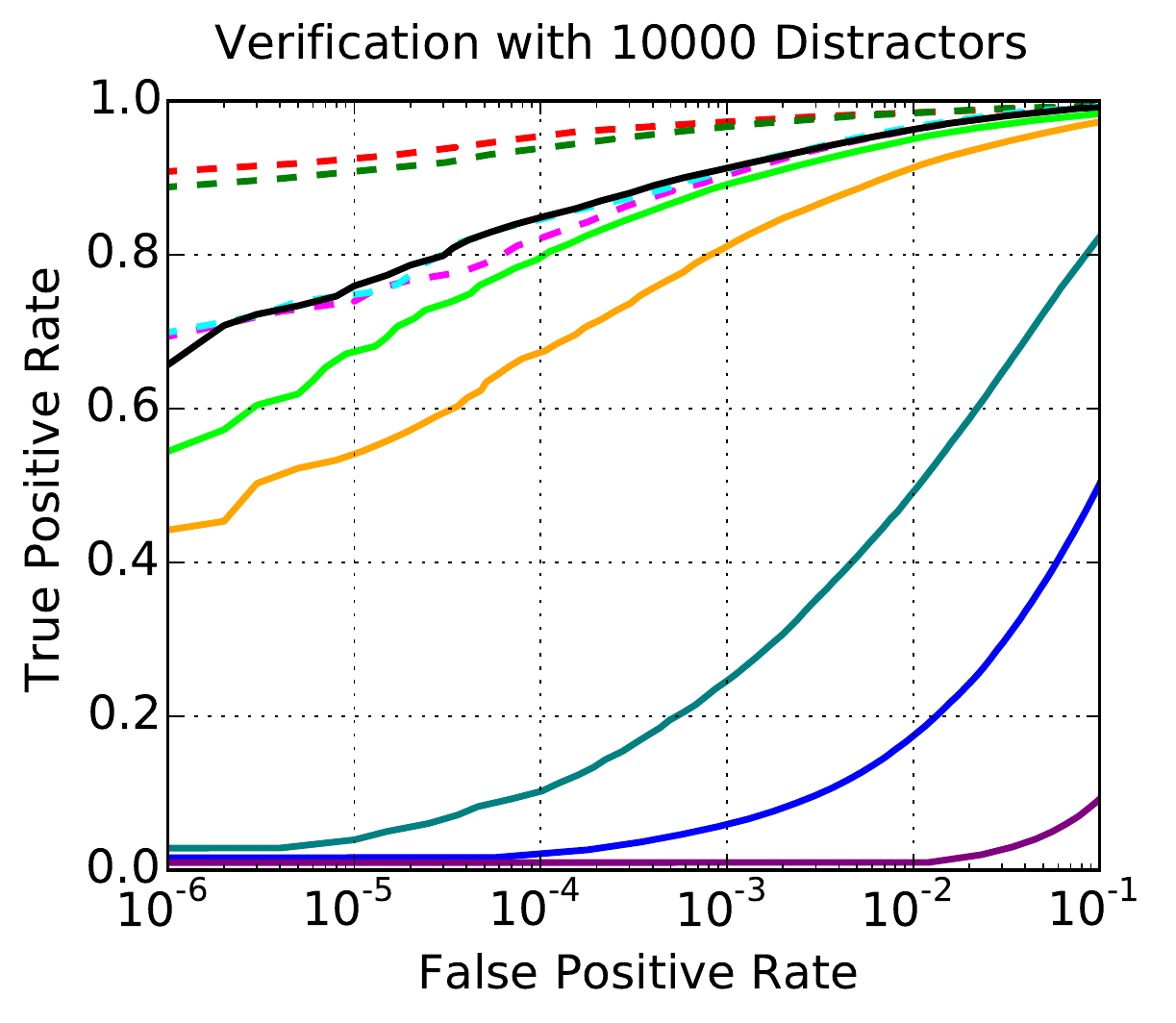} \\
					&{\footnotesize  (a) FaceScrub + 1M} & {\footnotesize (b) FaceScrub + 10K}  \\
					\includegraphics[width=.15\linewidth, valign=t]{plots16/ll.pdf} & 
					\includegraphics[width=0.27\linewidth, valign=t]{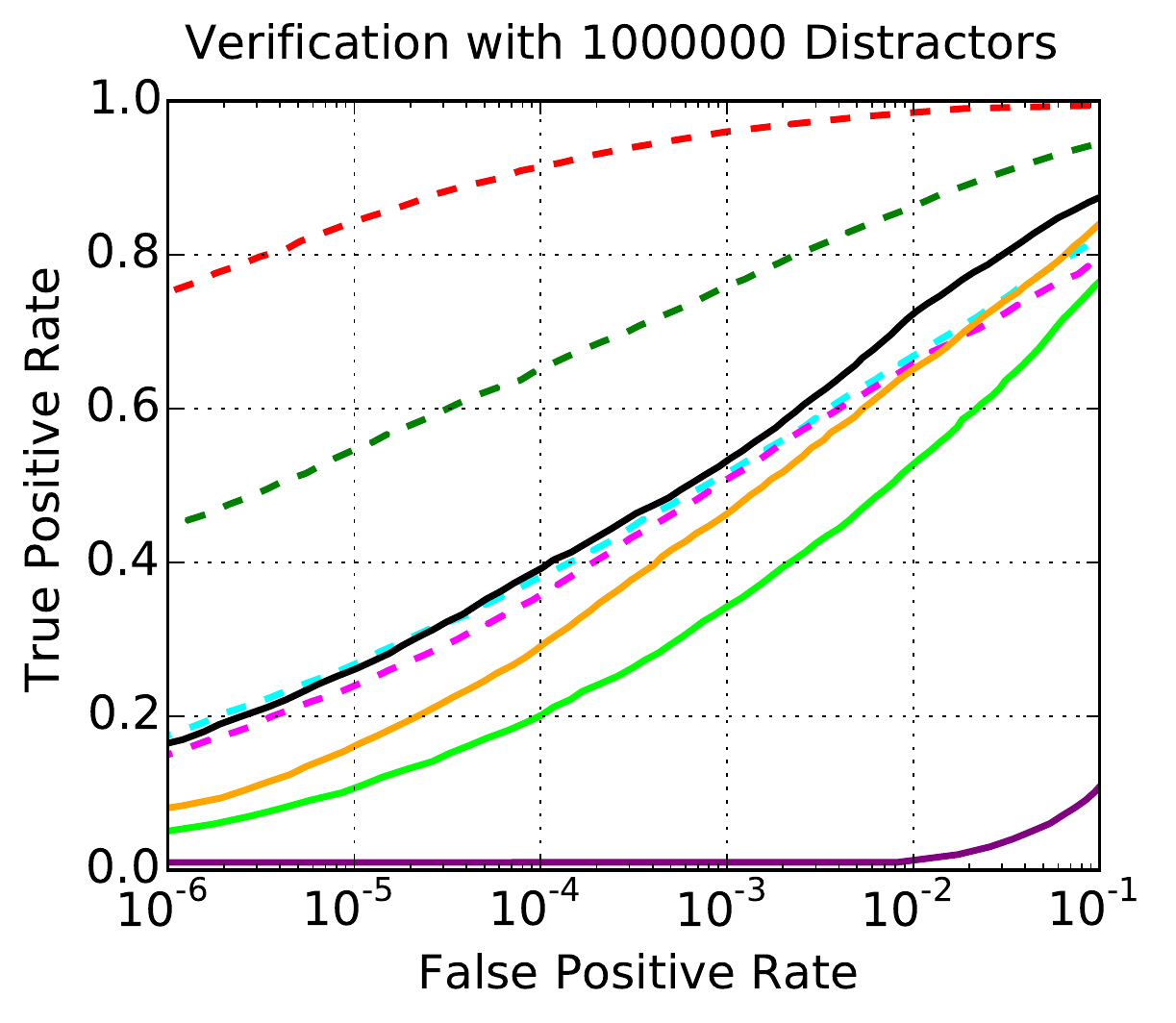} & 
					\includegraphics[width=0.27\linewidth, valign=t]{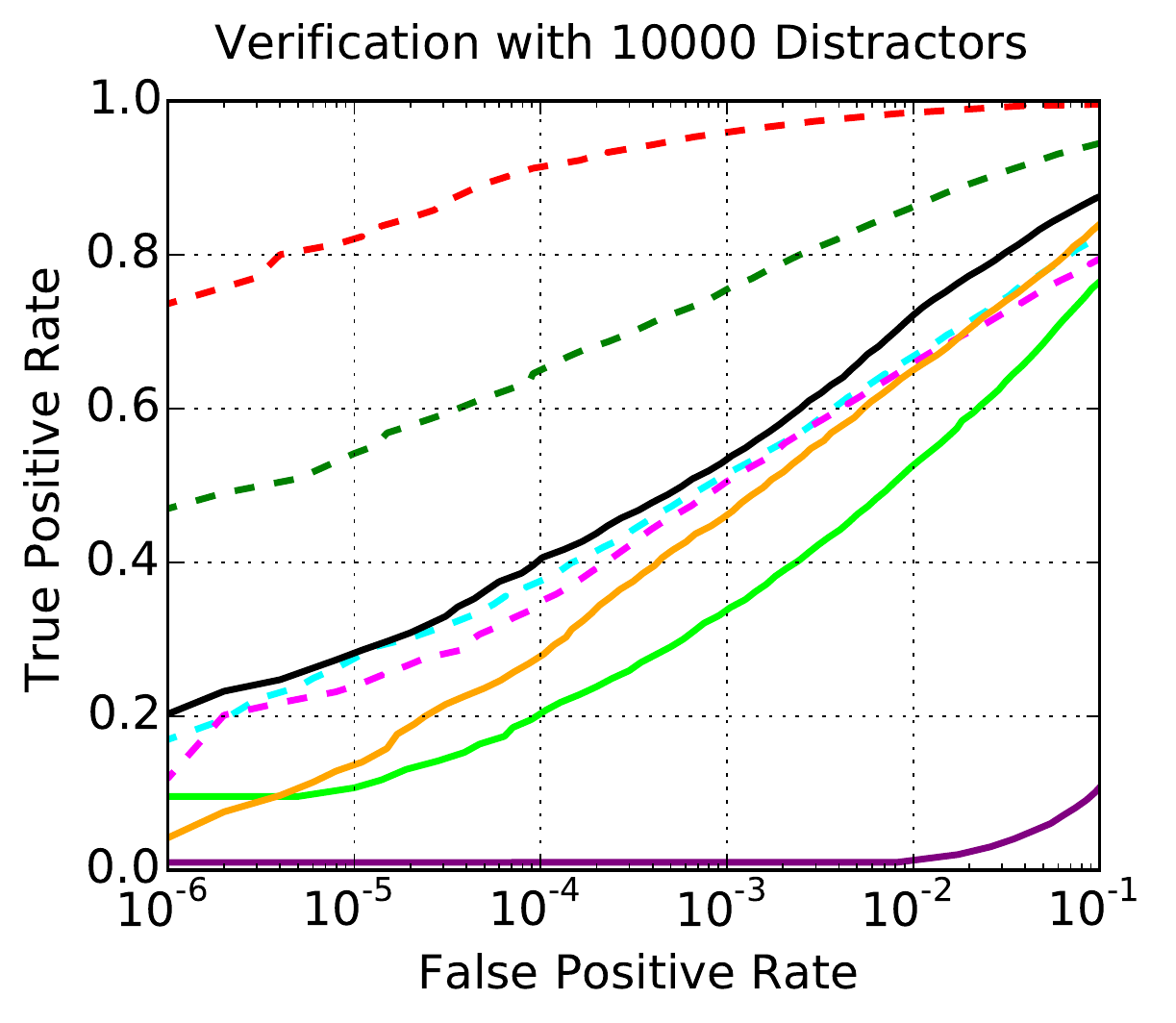} \\
					&{\footnotesize  (c) FGNET + 1M} & {\footnotesize (d) FGNET + 10K}  \\
				\end{tabular}
				\caption{\textbf{Verification  (random gallery set 2)} performance with (a,c) 1 Million and (b,d) 10K distractors on both probe sets. Note the performance at low false accept rates (left side of each plot). }\label{fig:verf}
				\label{fig:dataset_size_roc} 
			\end{figure*}

				\begin{figure*}
					\centering
					\begin{tabular}{ccc}
						\includegraphics[width=.15\linewidth, valign=t]{plots16/ll.pdf} & 
						\includegraphics[width=0.27\linewidth, valign=t]{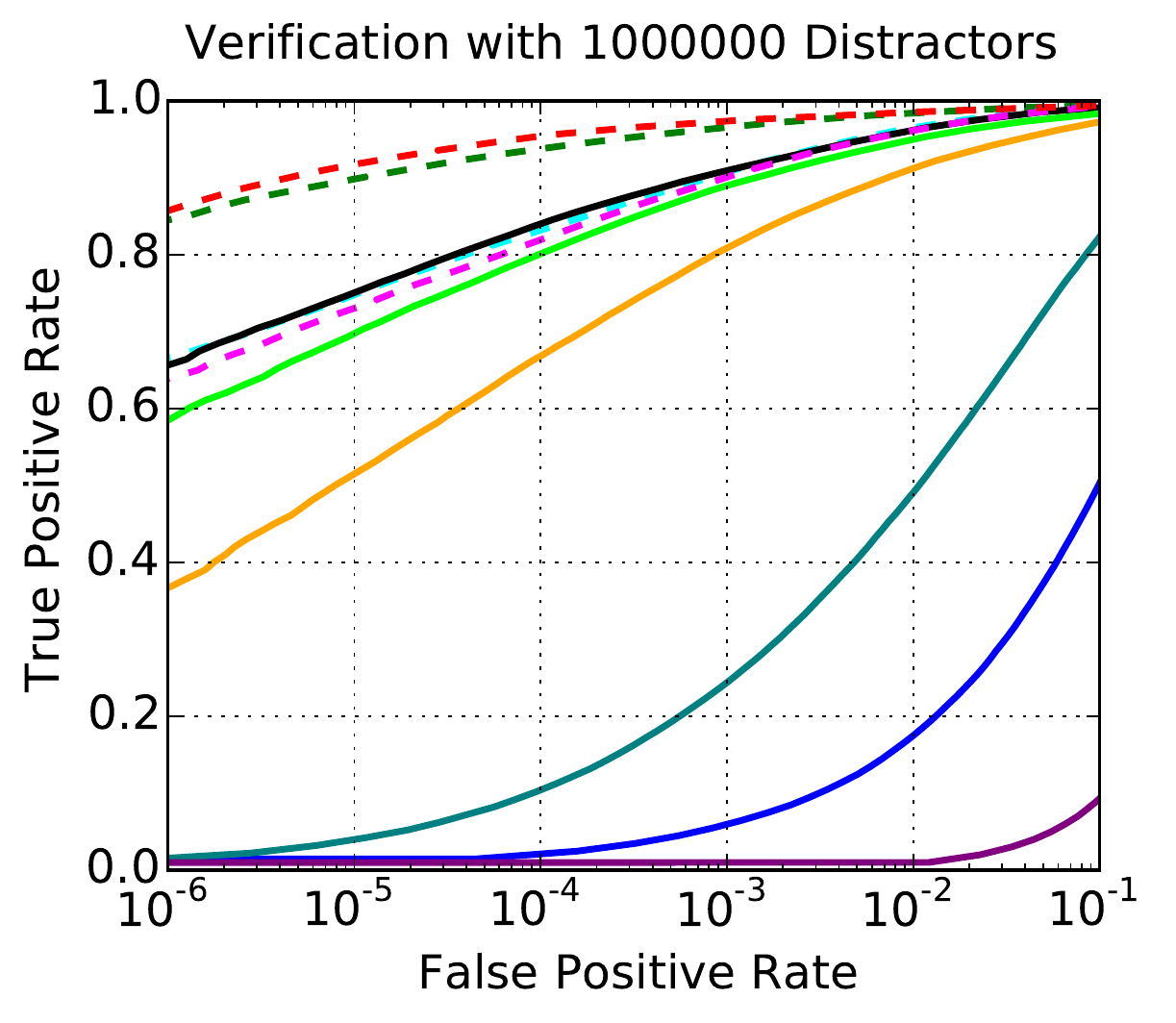} & 
						\includegraphics[width=0.27\linewidth, valign=t]{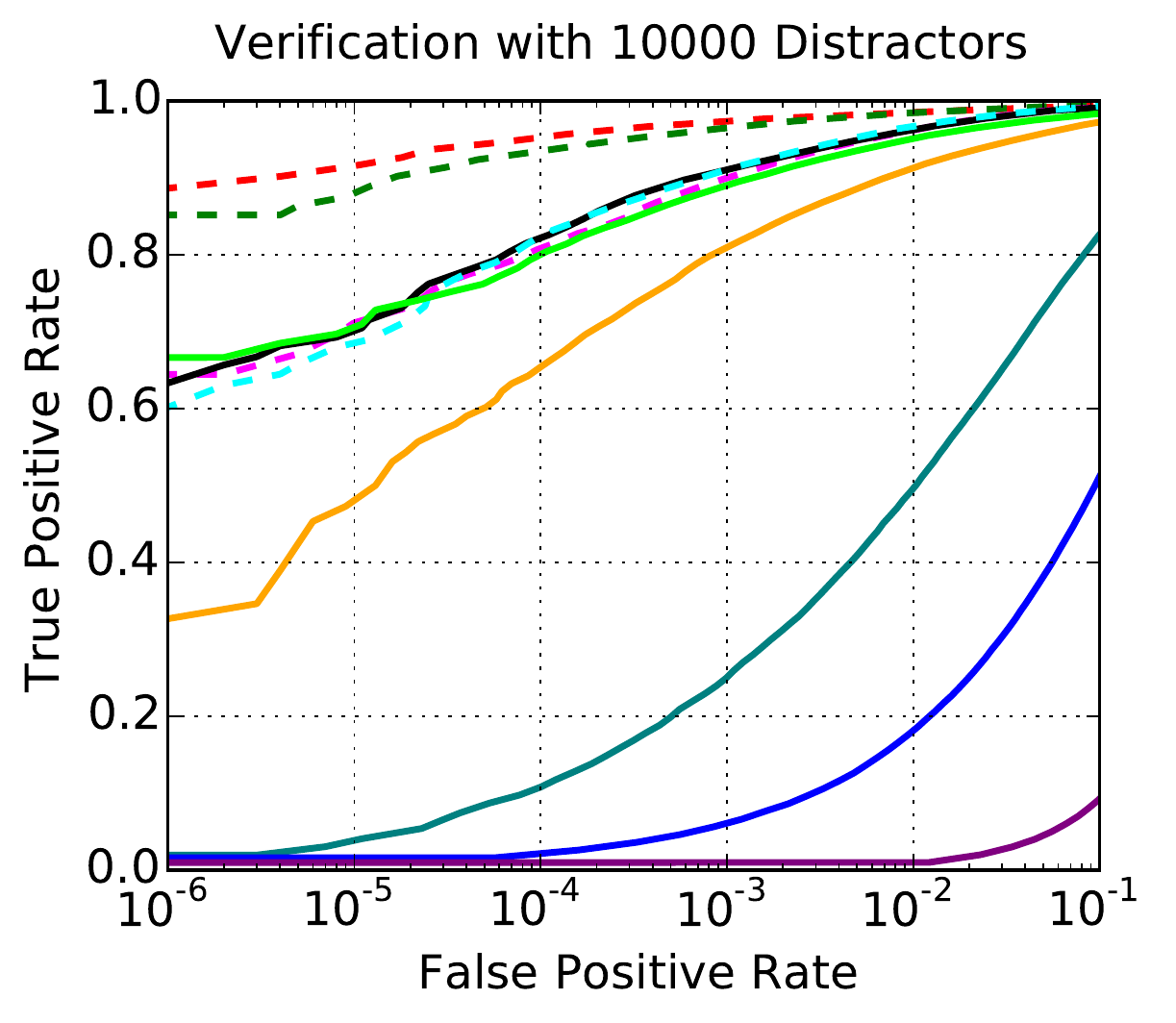} \\
						&{\footnotesize  (a) FaceScrub + 1M} & {\footnotesize (b) FaceScrub + 10K}  \\
						\includegraphics[width=.15\linewidth, valign=t]{plots16/ll.pdf} & 
						\includegraphics[width=0.27\linewidth, valign=t]{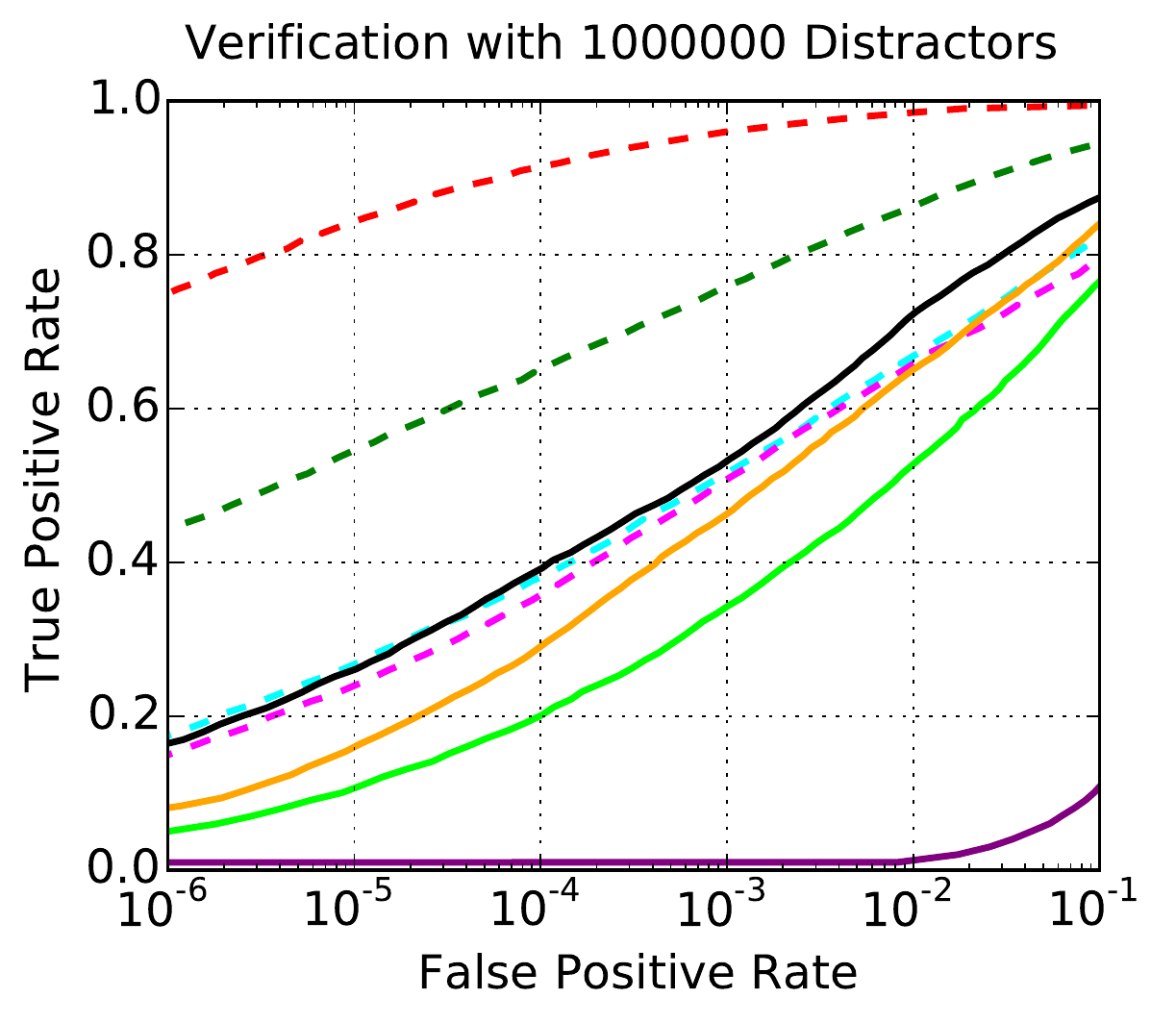} & 
						\includegraphics[width=0.27\linewidth, valign=t]{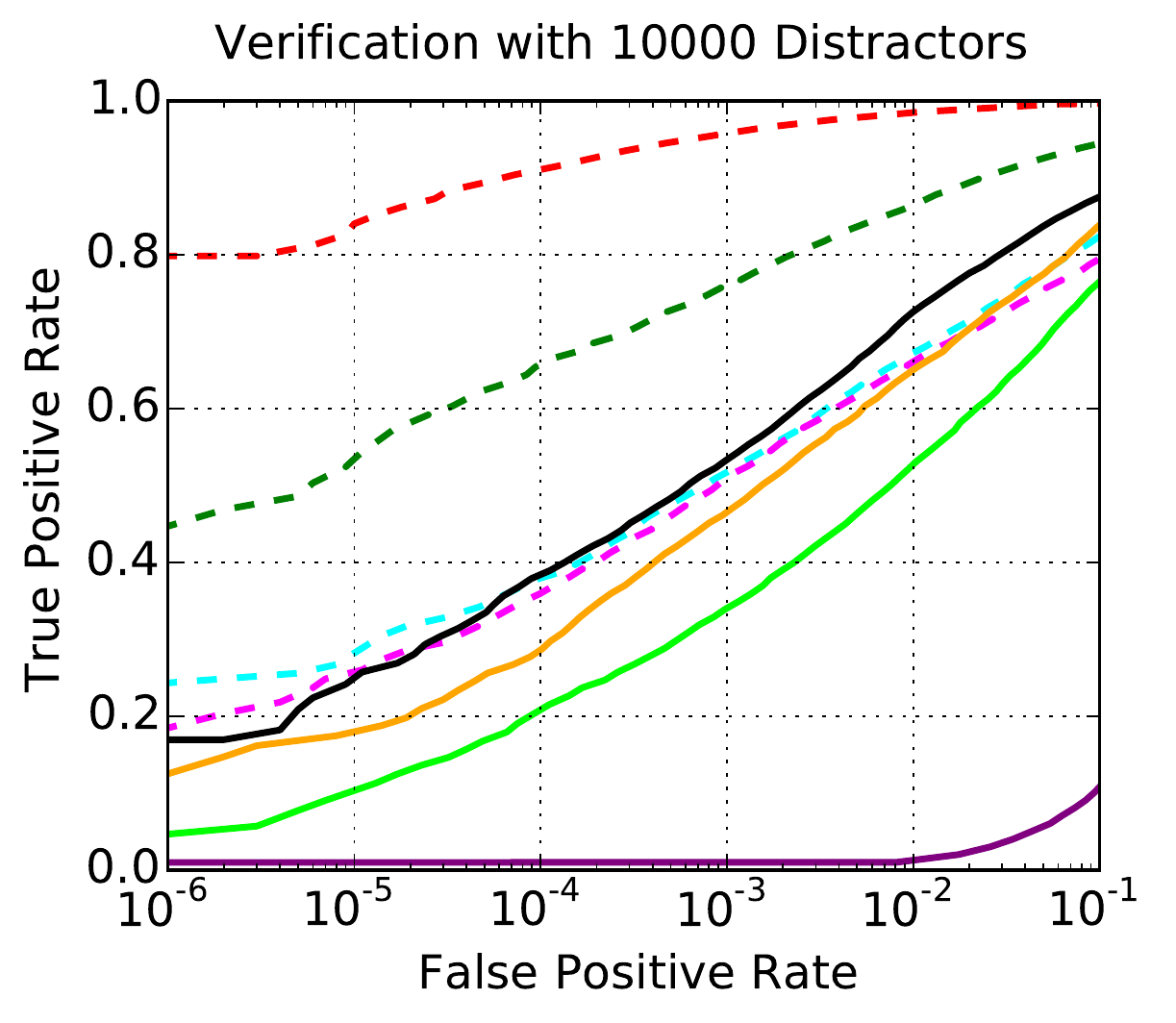} \\
						&{\footnotesize  (c) FGNET + 1M} & {\footnotesize (d) FGNET + 10K}  \\
					\end{tabular}
					\caption{\textbf{Verification  (random gallery set 3)} performance with (a,c) 1 Million and (b,d) 10K distractors on both probe sets. Note the performance at low false accept rates (left side of each plot). }\label{fig:verf}
					\label{fig:dataset_size_roc} 
				\end{figure*}

	\begin{figure*}
		\centering
		\begin{tabular}{cccc}
			\includegraphics[width=.15\linewidth, valign=t]{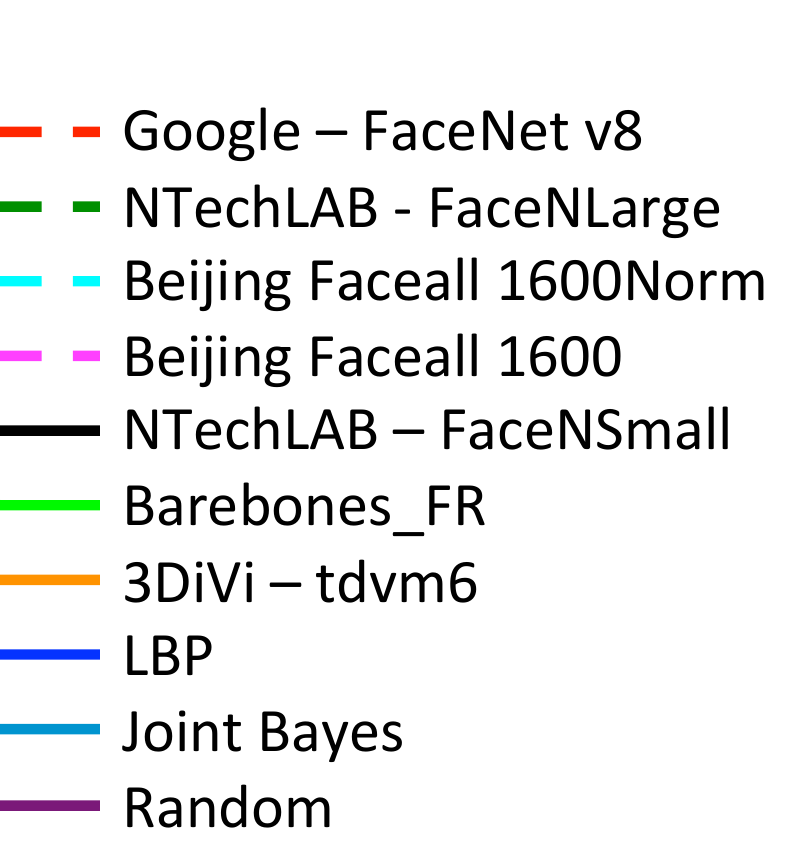} & 
			\includegraphics[width=0.25\linewidth, valign=t]{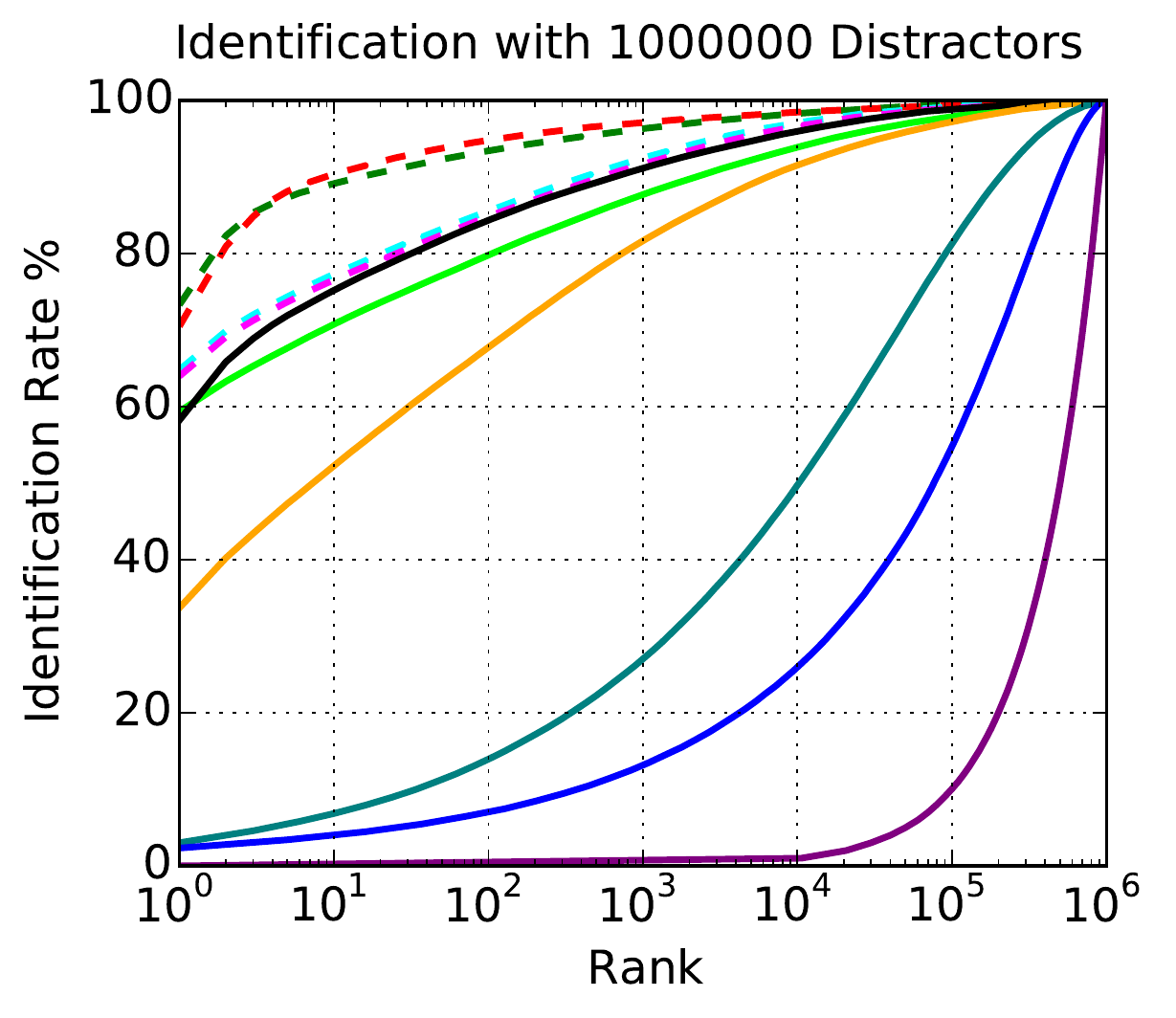} & 
			\includegraphics[width=0.25\linewidth, valign=t]{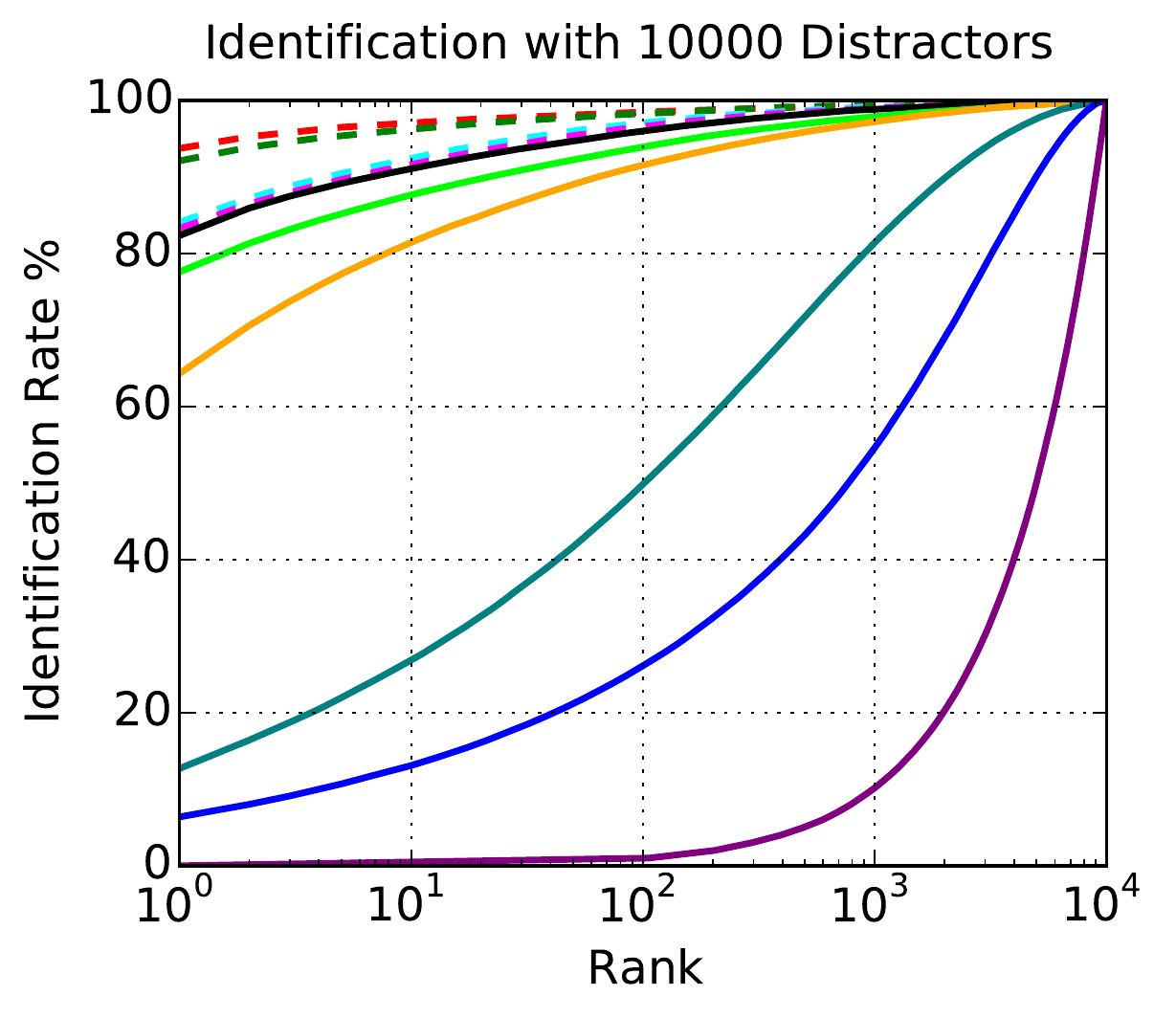} & 
			\includegraphics[width=0.25\linewidth, valign=t]{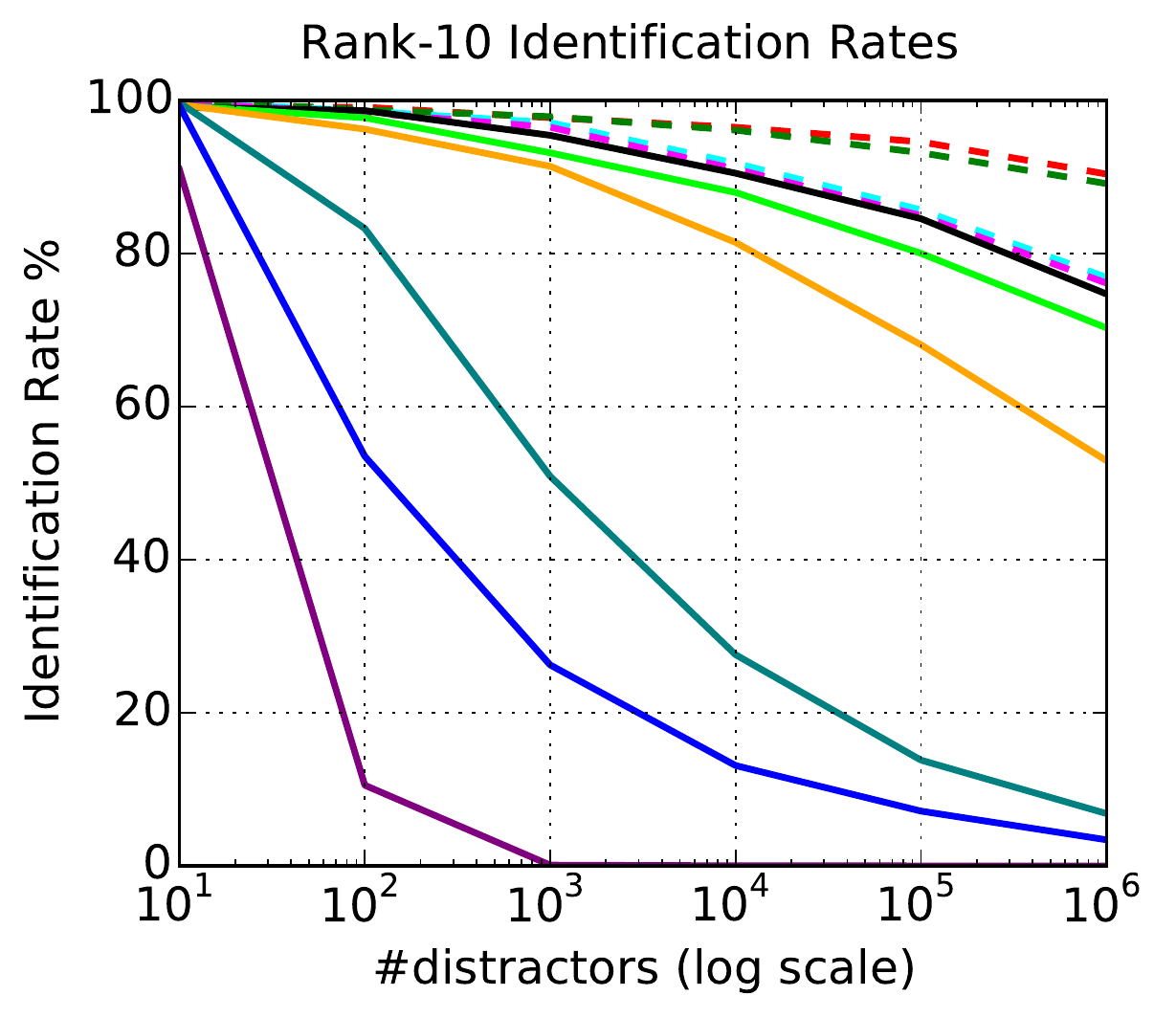}\\  
			&{\footnotesize  (a) FaceScrub + 1M} & {\footnotesize (b) FaceScrub + 10K} &    {\footnotesize (c) FaceScrub + rank-10} \\
			\includegraphics[width=.15\linewidth, valign=t]{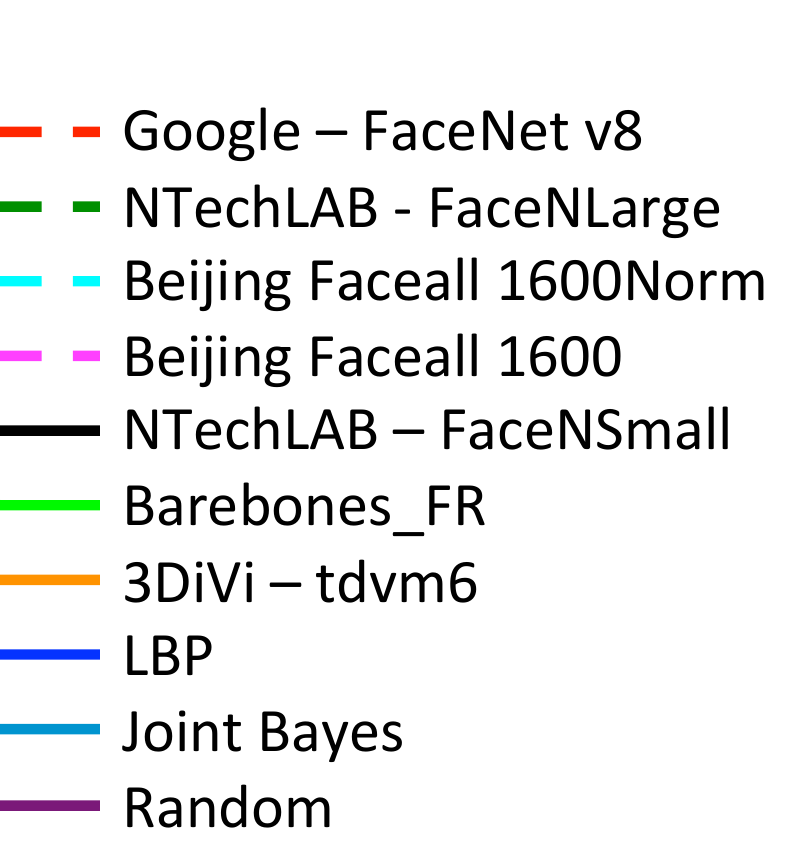} & 
			\includegraphics[width=0.25\linewidth, valign=t]{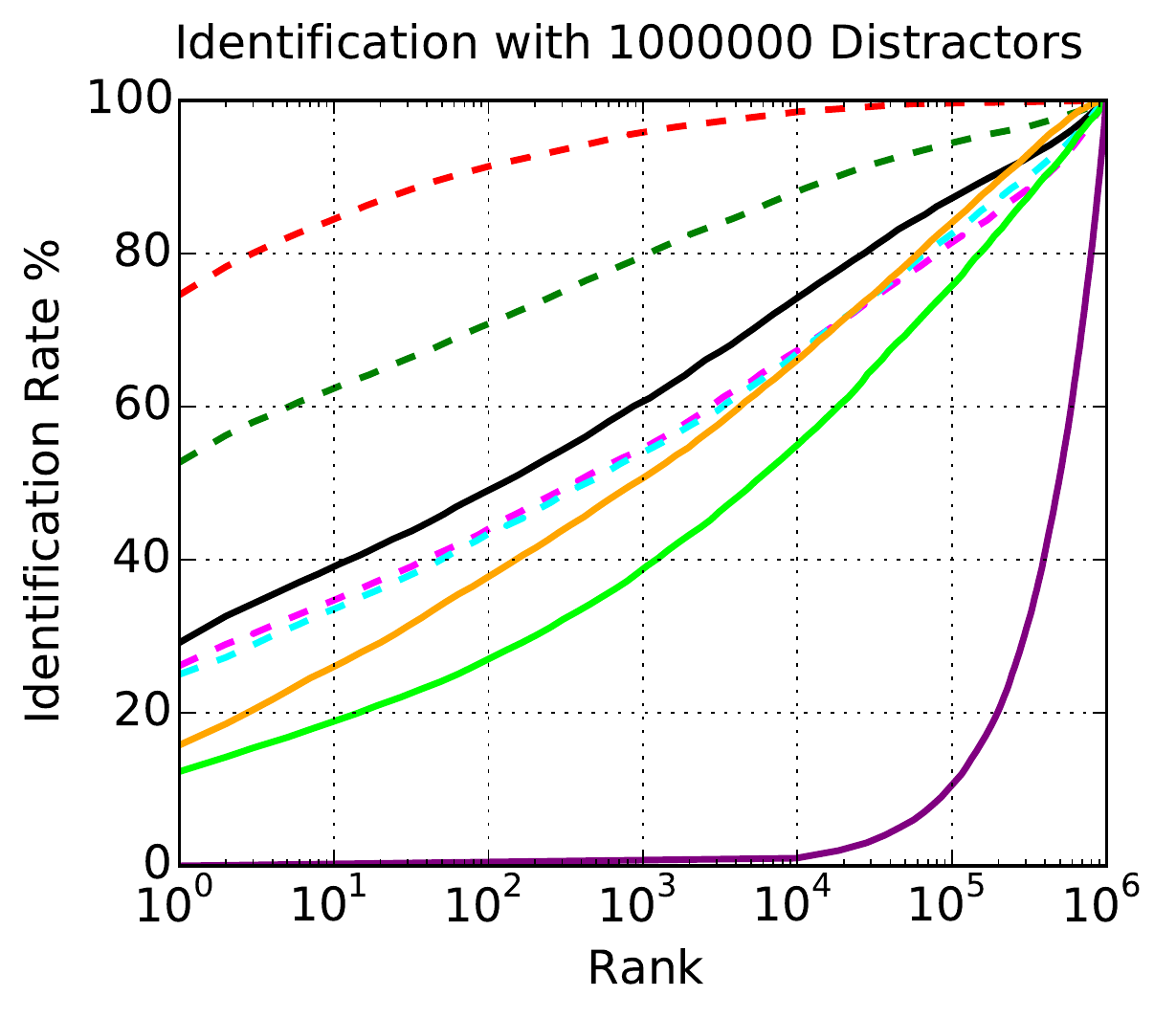} & 
			\includegraphics[width=0.25\linewidth, valign=t]{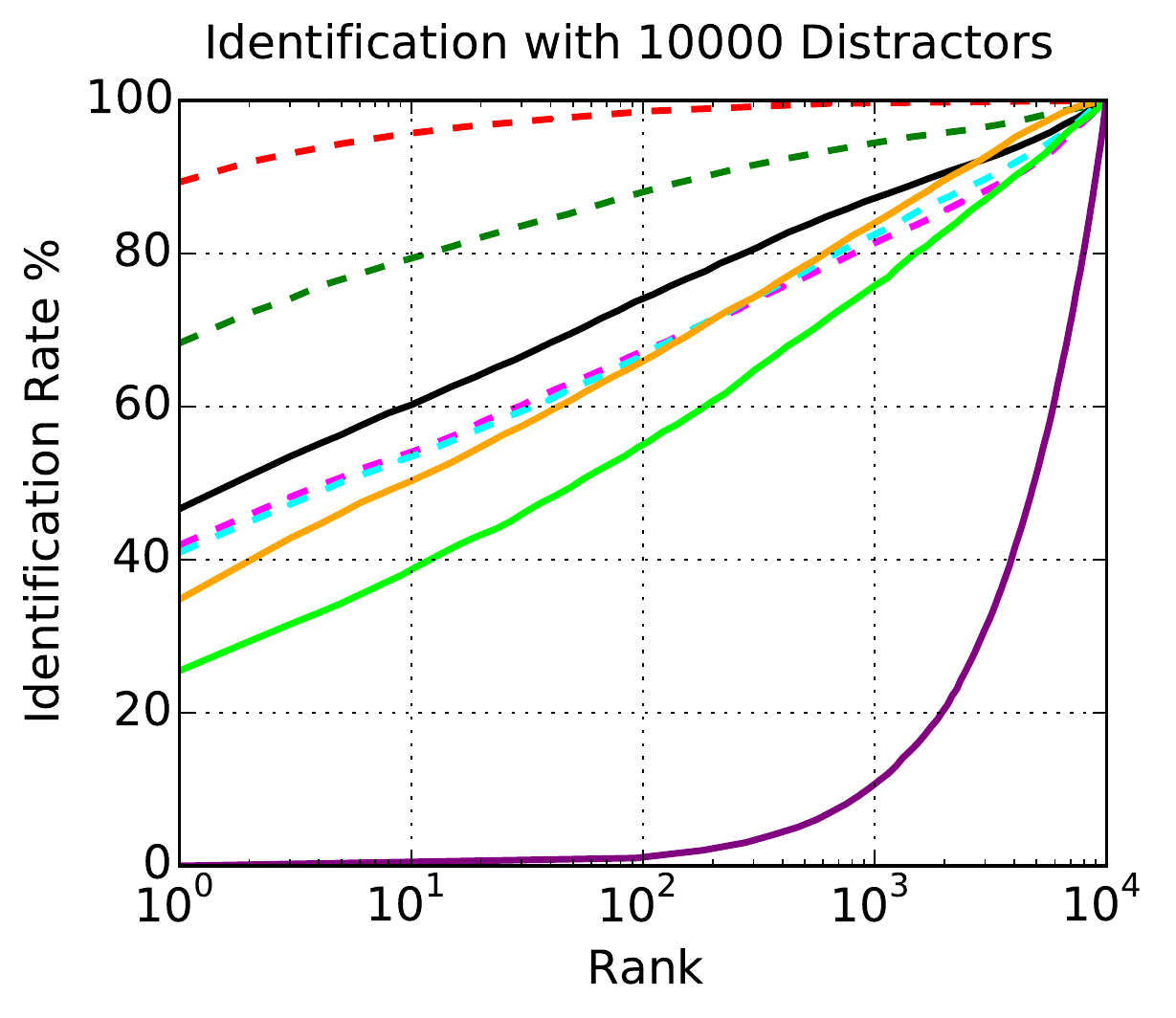} &
			\includegraphics[width=0.25\linewidth, valign=t]{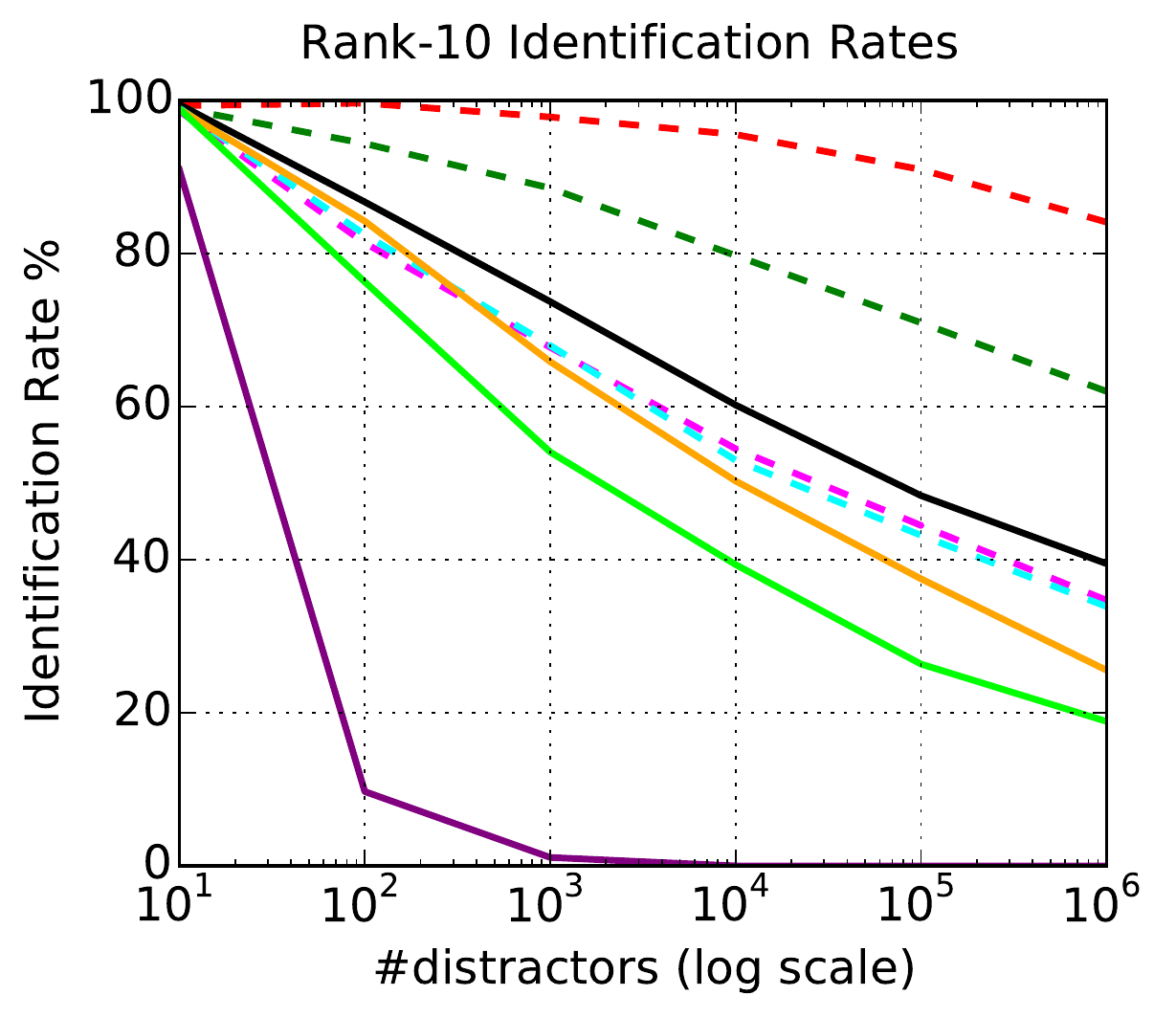}\\
			&		{\footnotesize (d) FGNET + 1M} & {\footnotesize (e) FGNET + 10K}   & {\footnotesize (f) FGNET + rank-10} 
		\end{tabular}
		
		\caption{\textbf{Identification  (random gallery set 1)} performance for all methods with (a,d) 1M distractors and (b,e) 10K distractors, and (c,f) rank-10 for both probe sets.  Fig.~\ref{fig:teaser} also shows  rank-1 performance as a function of number of distractors on both probe sets. }
		\label{fig:dataset_size_cmc}
	\end{figure*}

	\begin{figure*}
		\centering
		\begin{tabular}{cccc}
			\includegraphics[width=.15\linewidth, valign=t]{plots16/ll.pdf} & 
			\includegraphics[width=0.25\linewidth, valign=t]{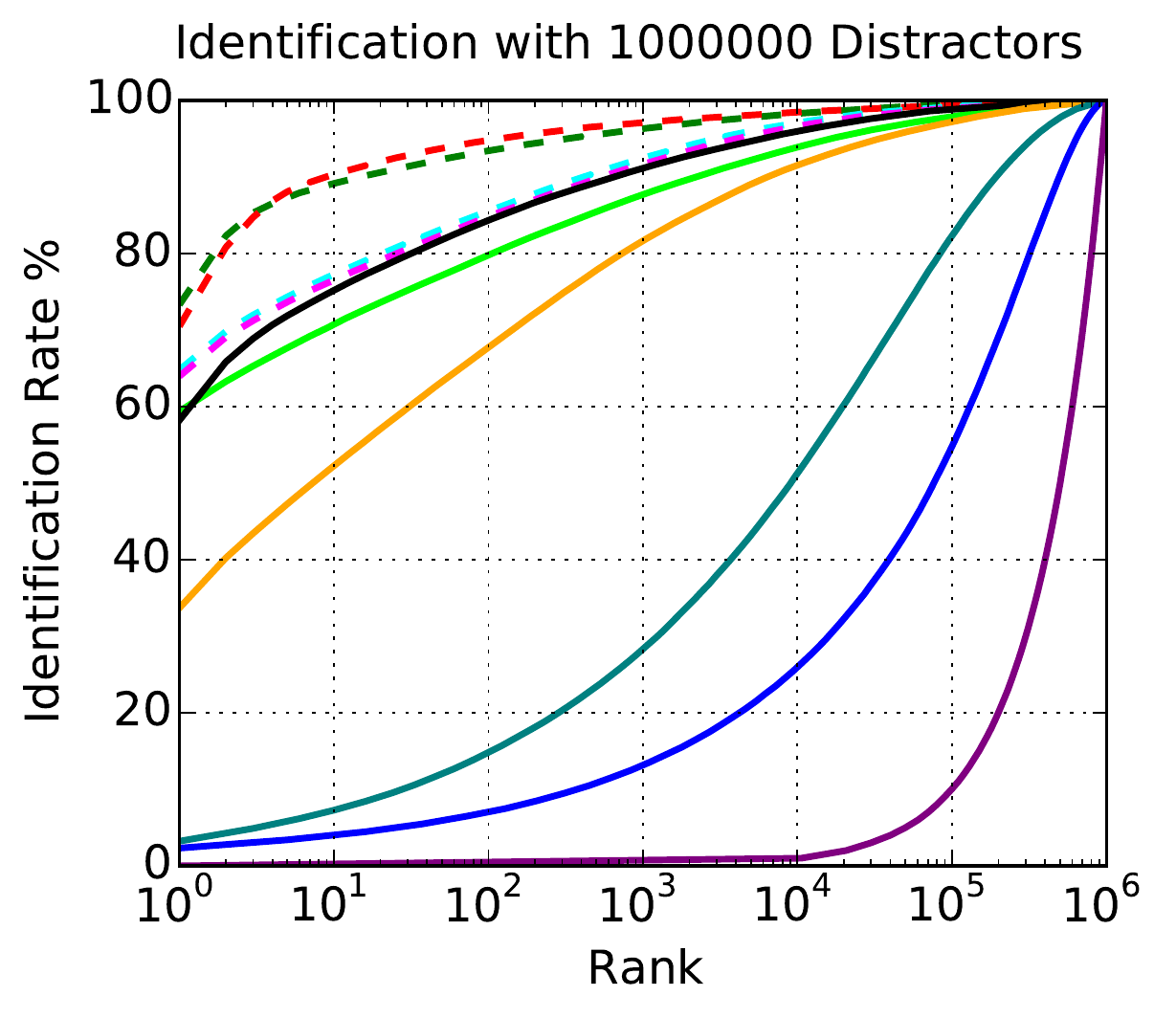} & 
			\includegraphics[width=0.25\linewidth, valign=t]{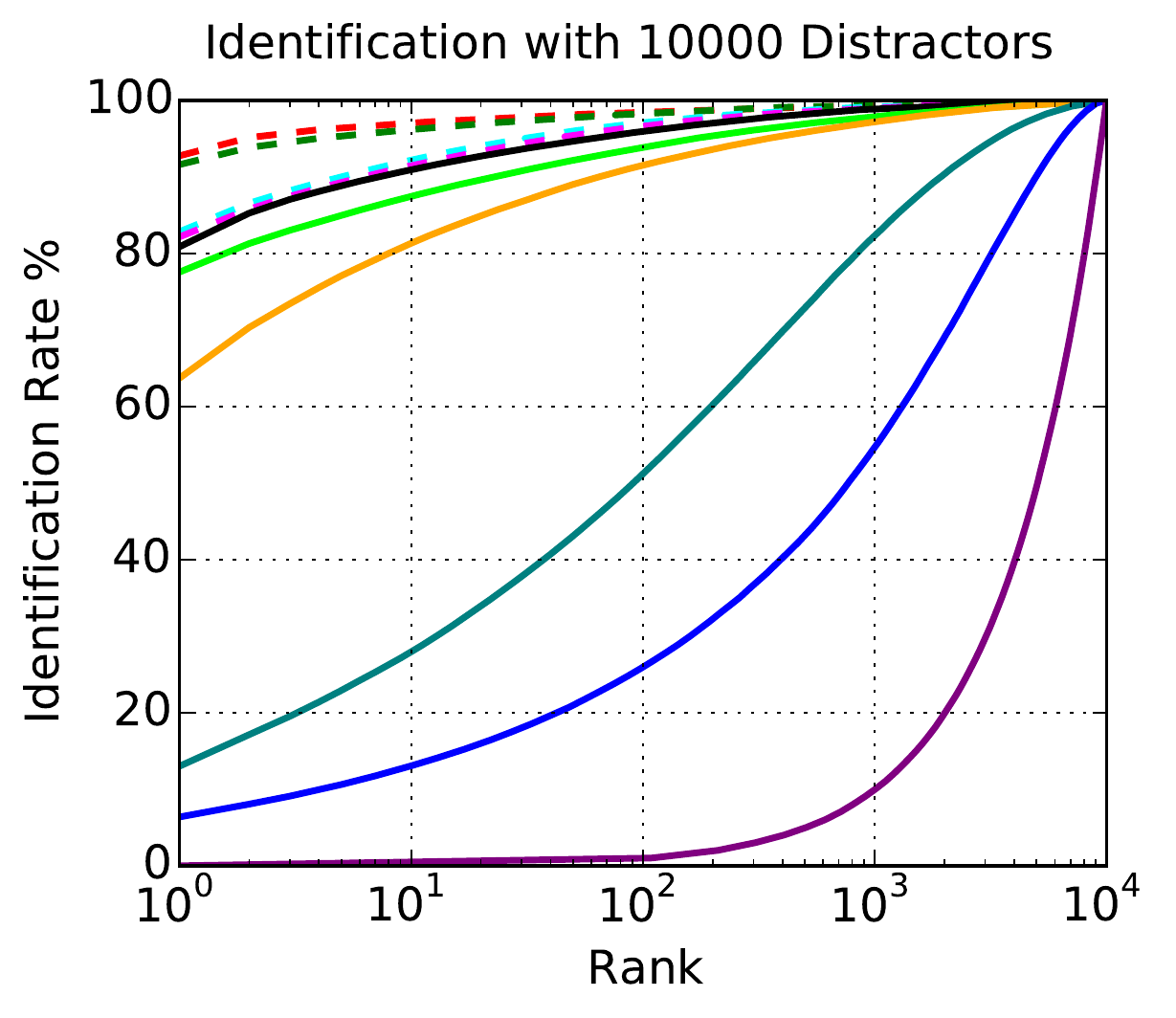} & 
			\includegraphics[width=0.25\linewidth, valign=t]{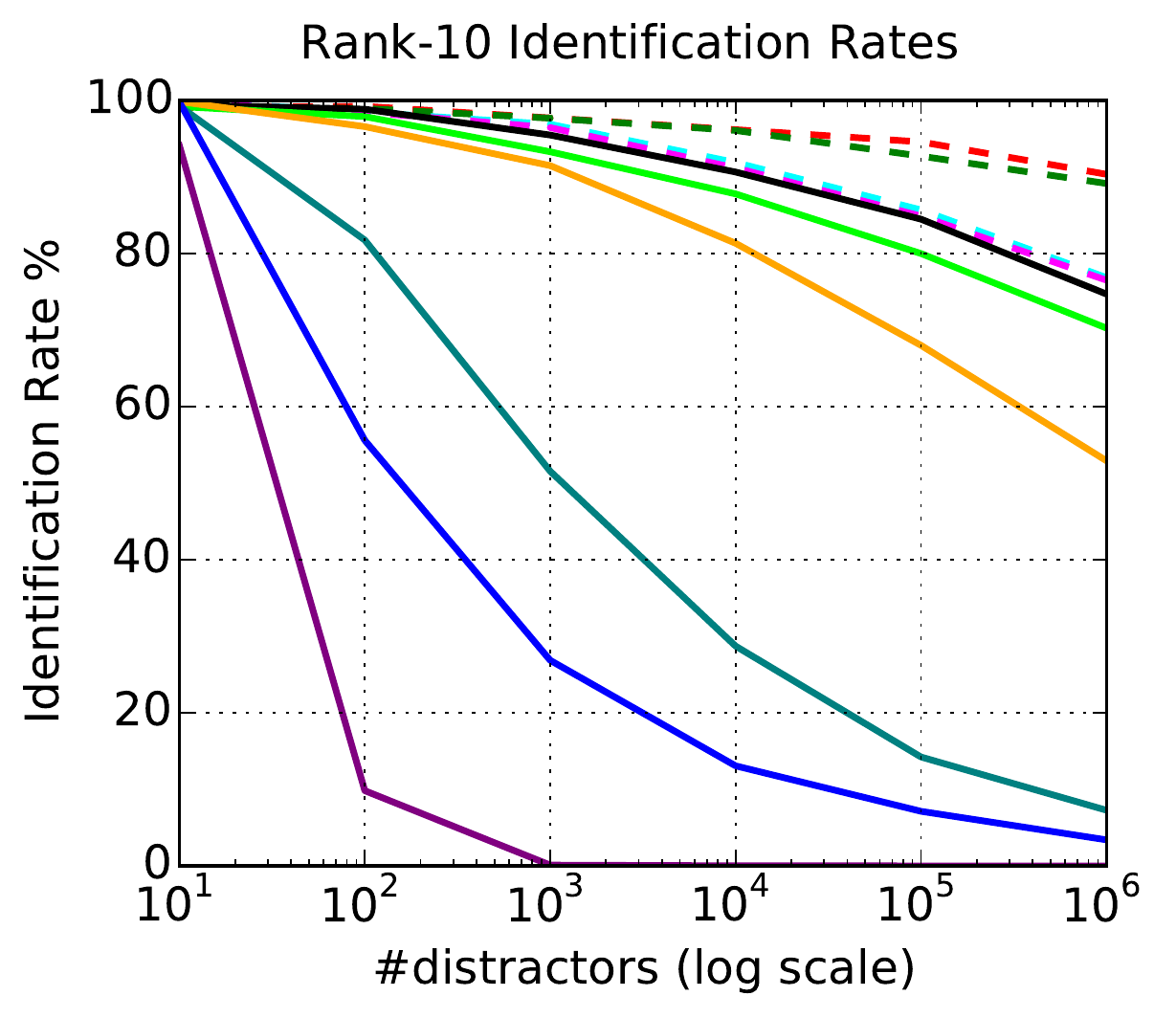}\\  
			&{\footnotesize  (a) FaceScrub + 1M} & {\footnotesize (b) FaceScrub + 10K} &    {\footnotesize (c) FaceScrub + rank-10} \\
			\includegraphics[width=.15\linewidth, valign=t]{plots16/ll.pdf} & 
			\includegraphics[width=0.25\linewidth, valign=t]{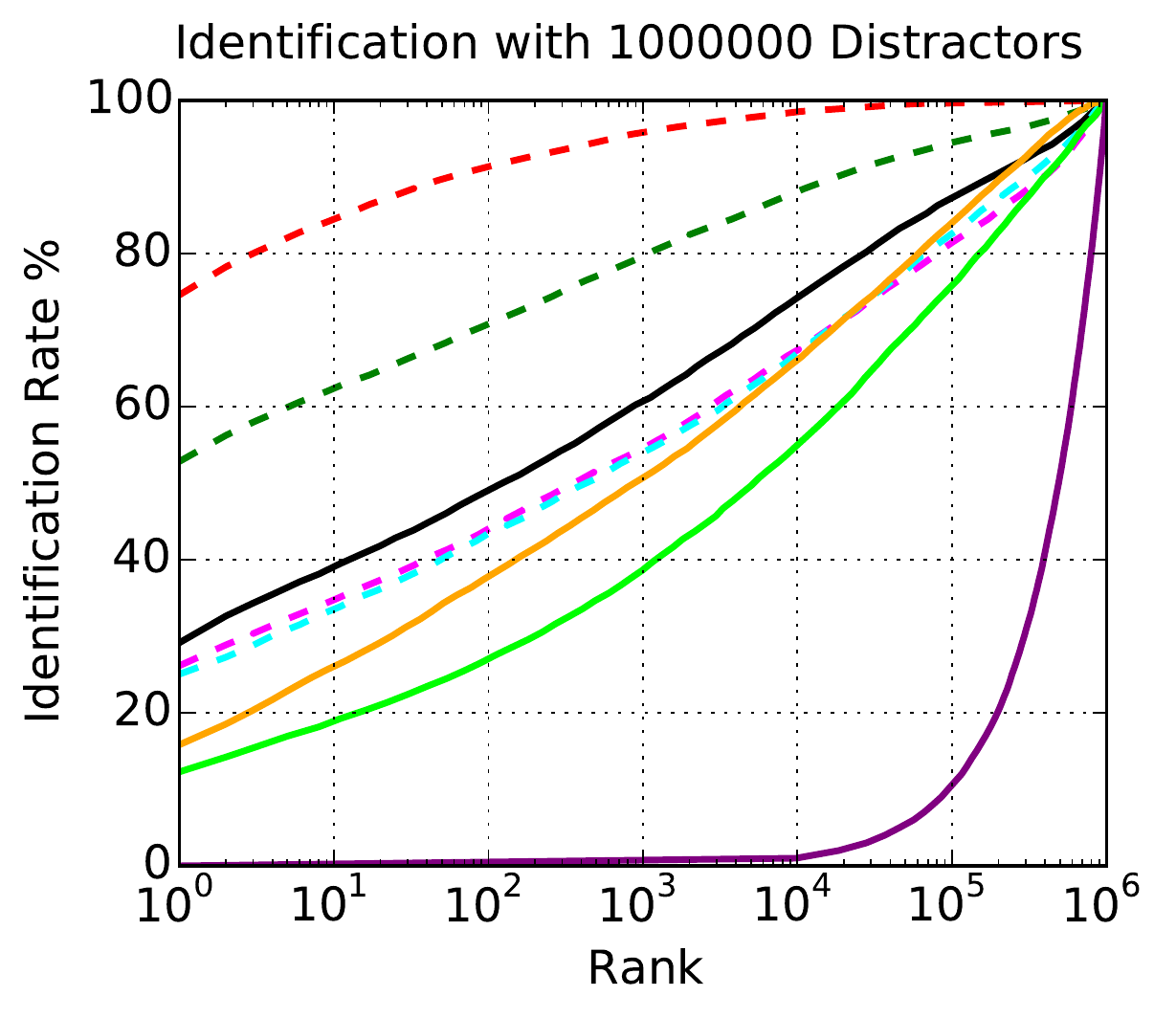} & 
			\includegraphics[width=0.25\linewidth, valign=t]{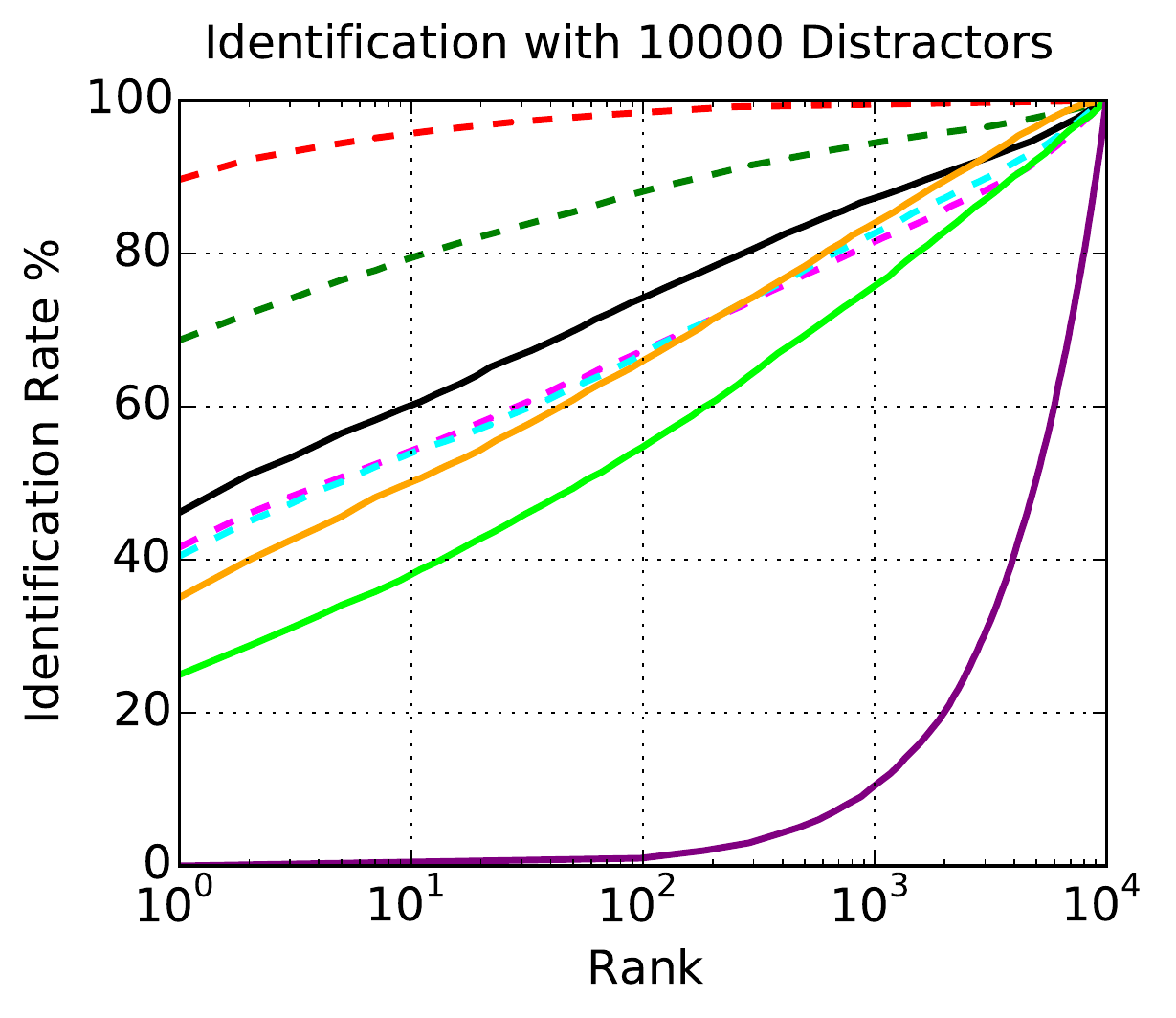} &
			\includegraphics[width=0.25\linewidth, valign=t]{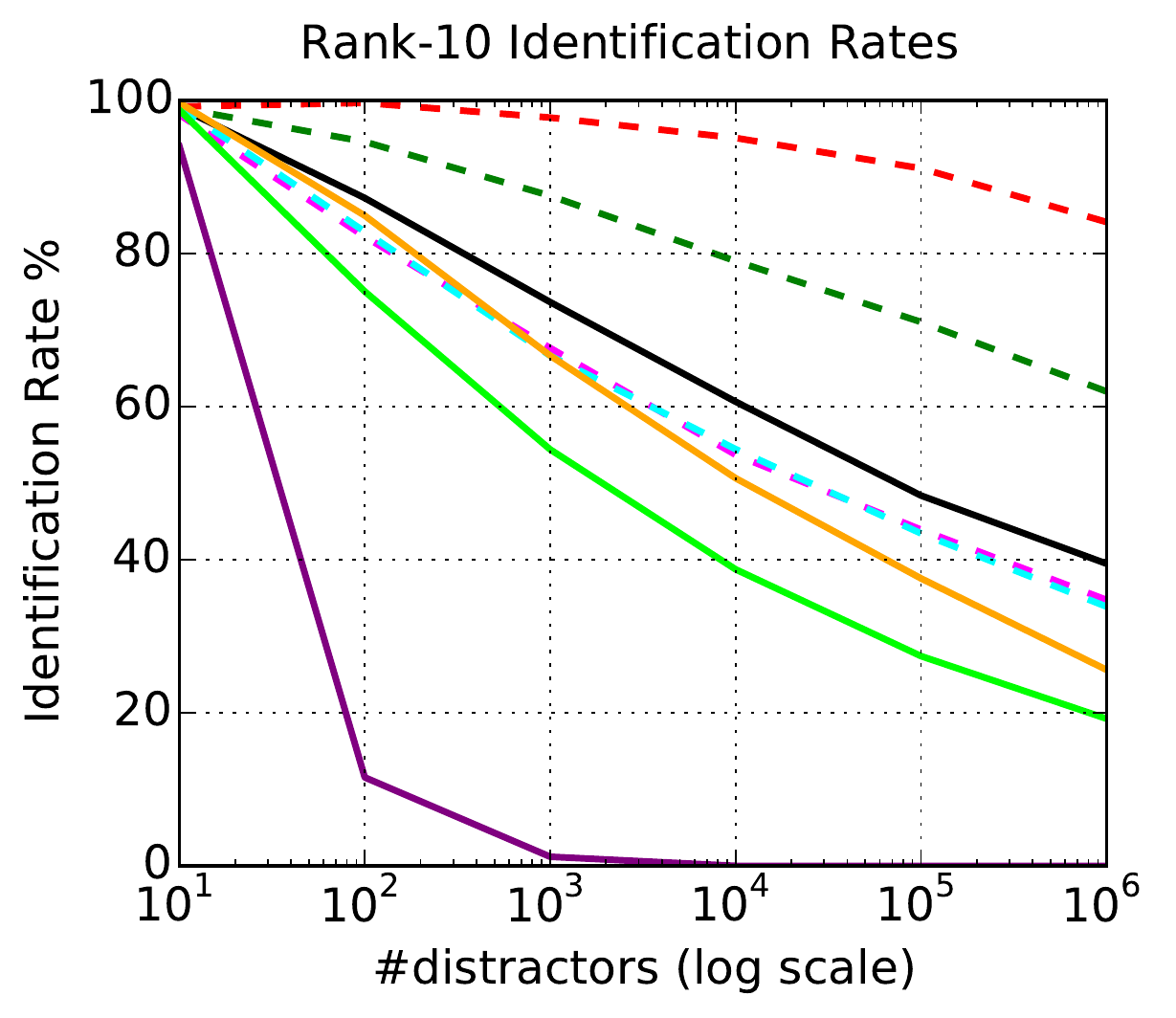}\\
			&		{\footnotesize (d) FGNET + 1M} & {\footnotesize (e) FGNET + 10K}   & {\footnotesize (f) FGNET + rank-10} 
		\end{tabular}
		
		\caption{\textbf{Identification  (random gallery set 2)} performance for all methods with (a,d) 1M distractors and (b,e) 10K distractors, and (c,f) rank-10 for both probe sets.  Fig.~\ref{fig:teaser} also shows  rank-1 performance as a function of number of distractors on both probe sets. }
		\label{fig:dataset_size_cmc}
	\end{figure*}
	
		\begin{figure*}
			\centering
			\begin{tabular}{cccc}
				\includegraphics[width=.15\linewidth, valign=t]{plots16/ll.pdf} & 
				\includegraphics[width=0.25\linewidth, valign=t]{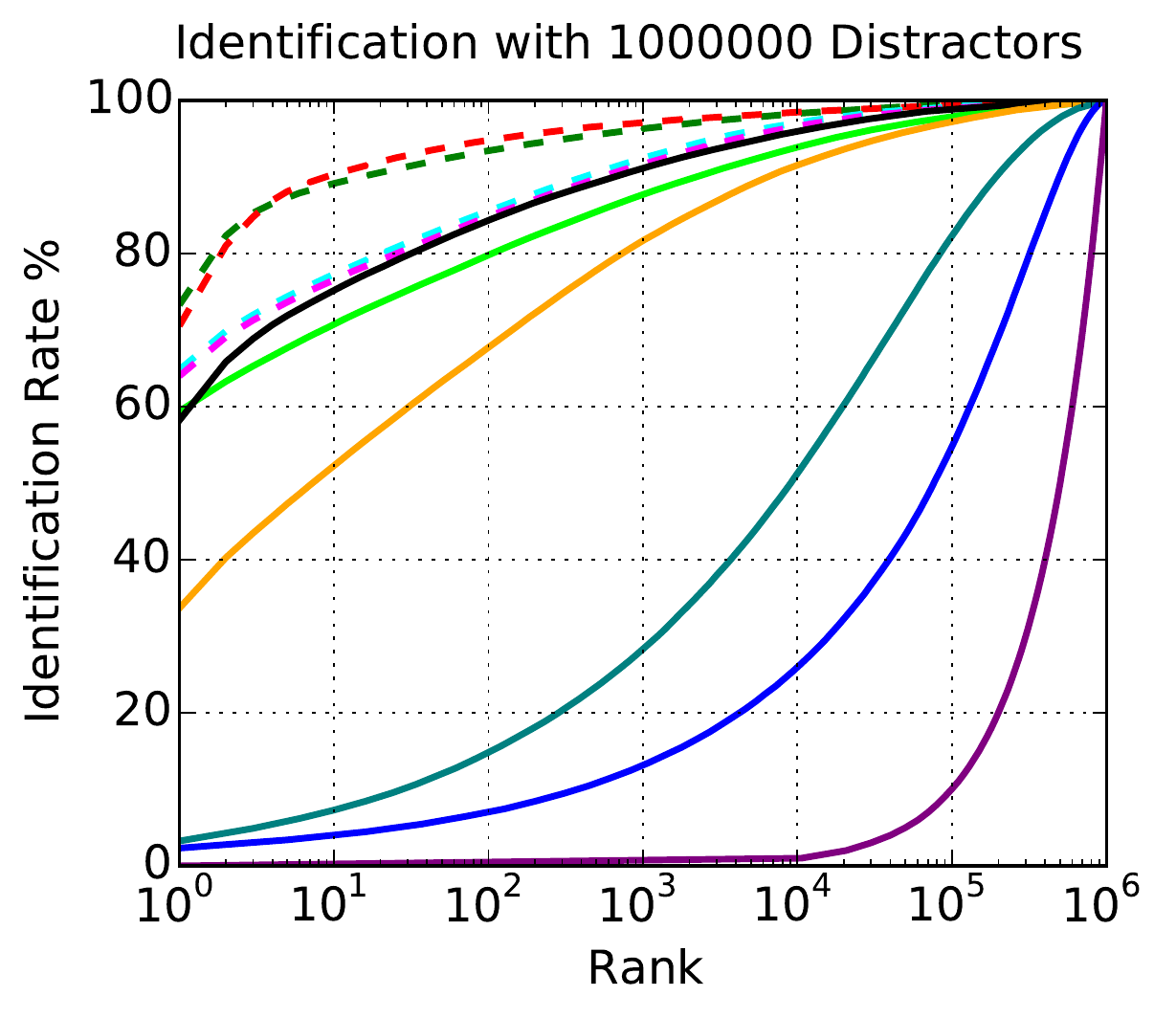} & 
				\includegraphics[width=0.25\linewidth, valign=t]{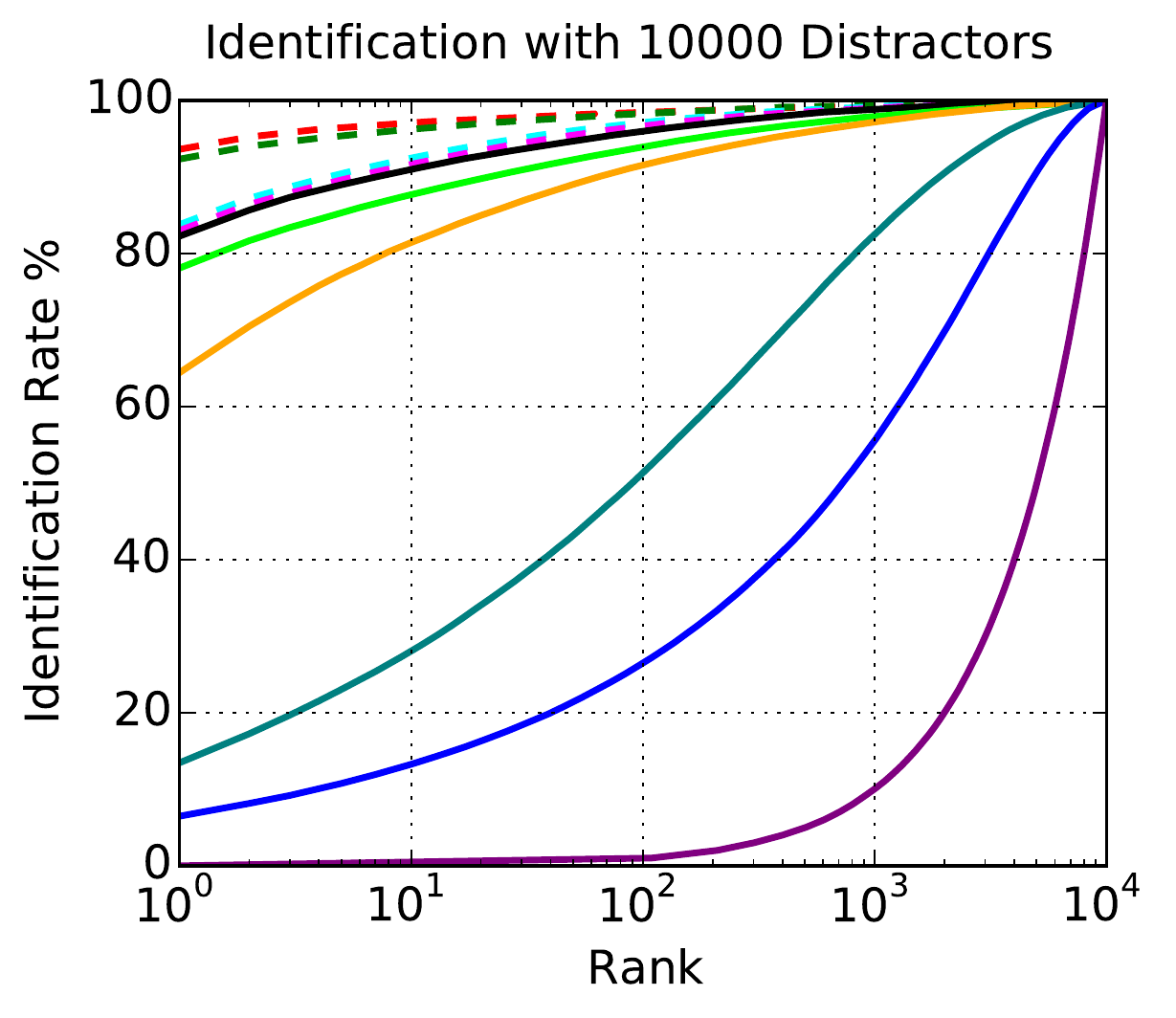} & 
				\includegraphics[width=0.25\linewidth, valign=t]{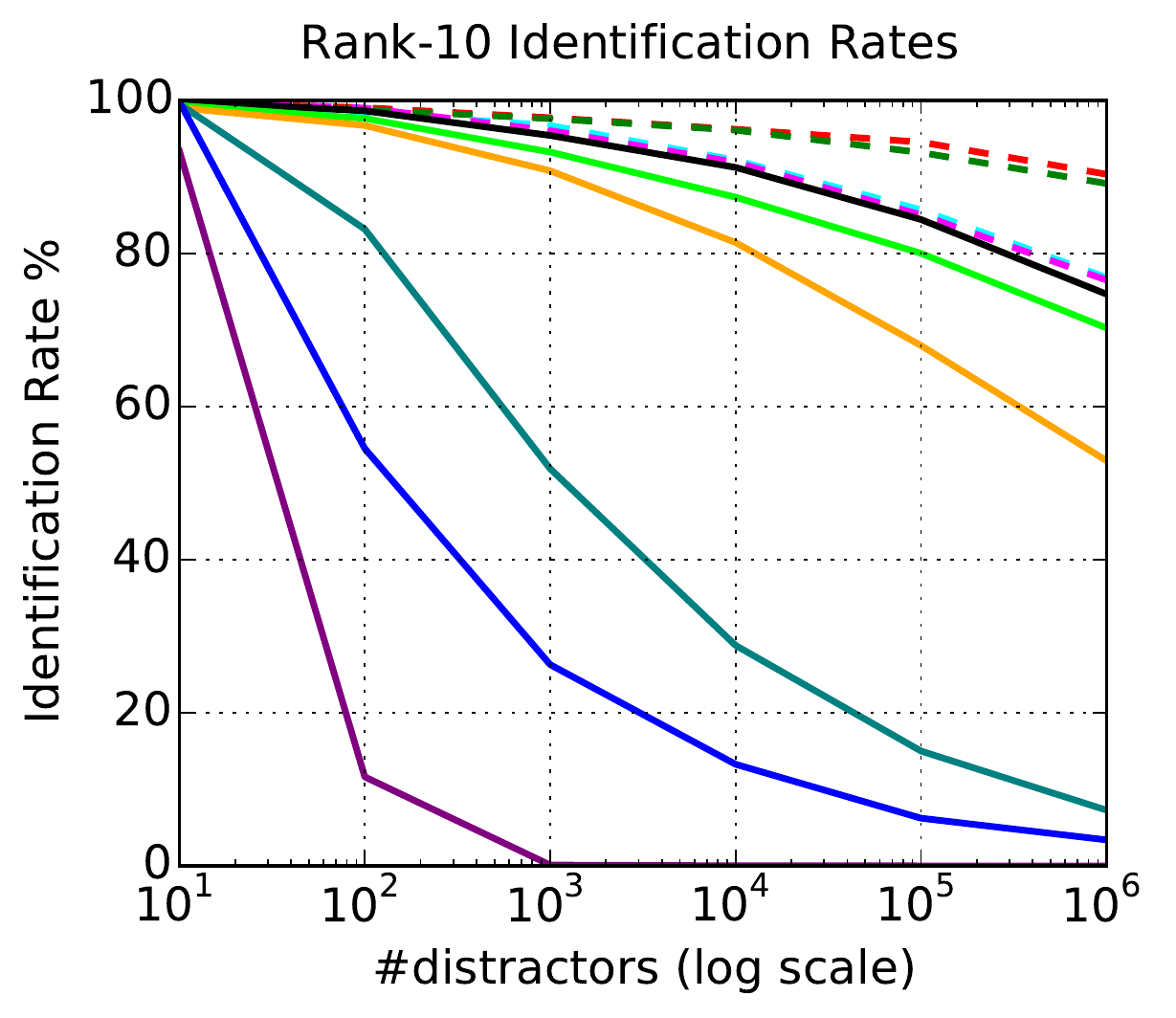}\\  
				&{\footnotesize  (a) FaceScrub + 1M} & {\footnotesize (b) FaceScrub + 10K} &    {\footnotesize (c) FaceScrub + rank-10} \\
				\includegraphics[width=.15\linewidth, valign=t]{plots16/ll.pdf} & 
				\includegraphics[width=0.25\linewidth, valign=t]{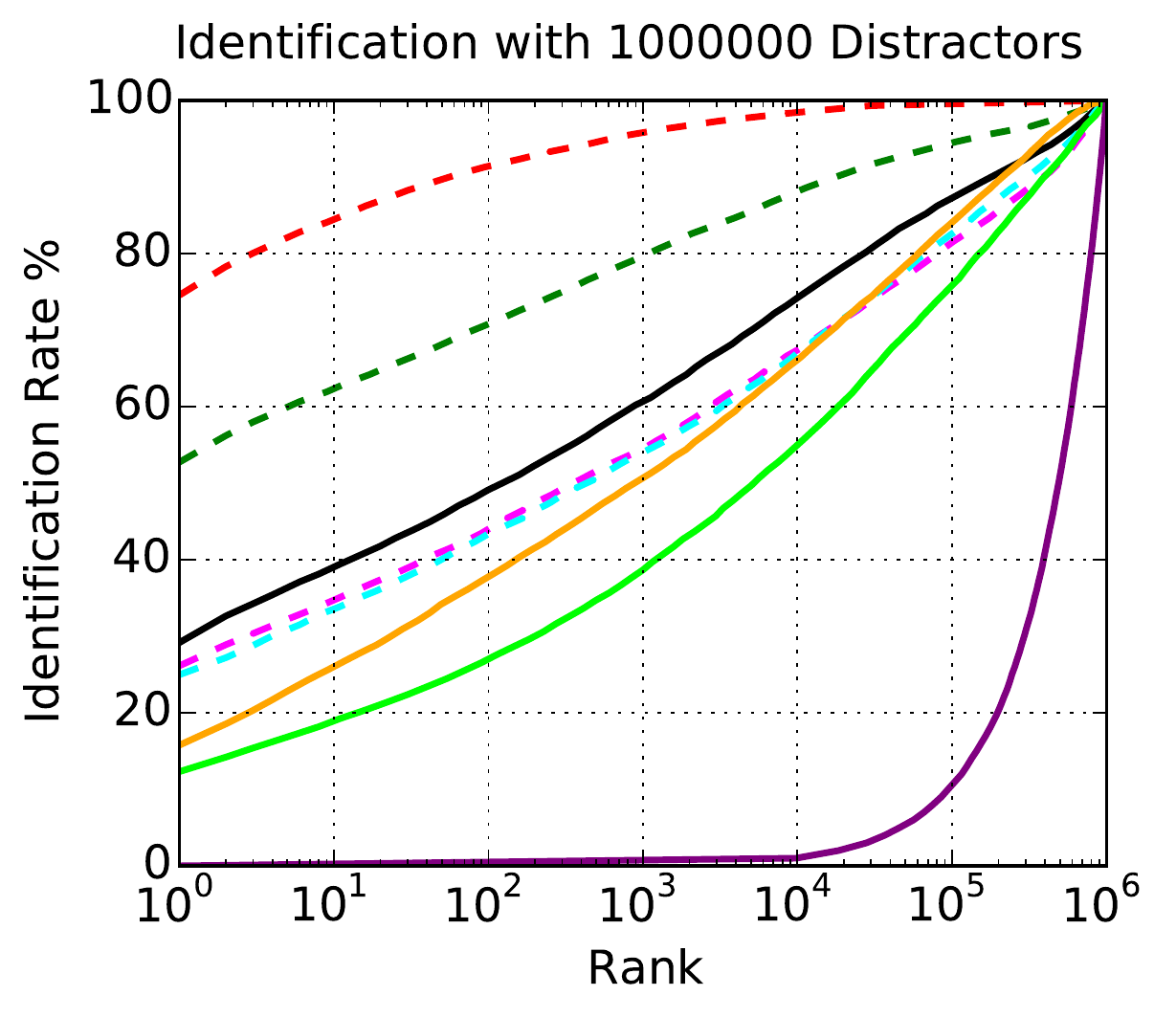} & 
				\includegraphics[width=0.25\linewidth, valign=t]{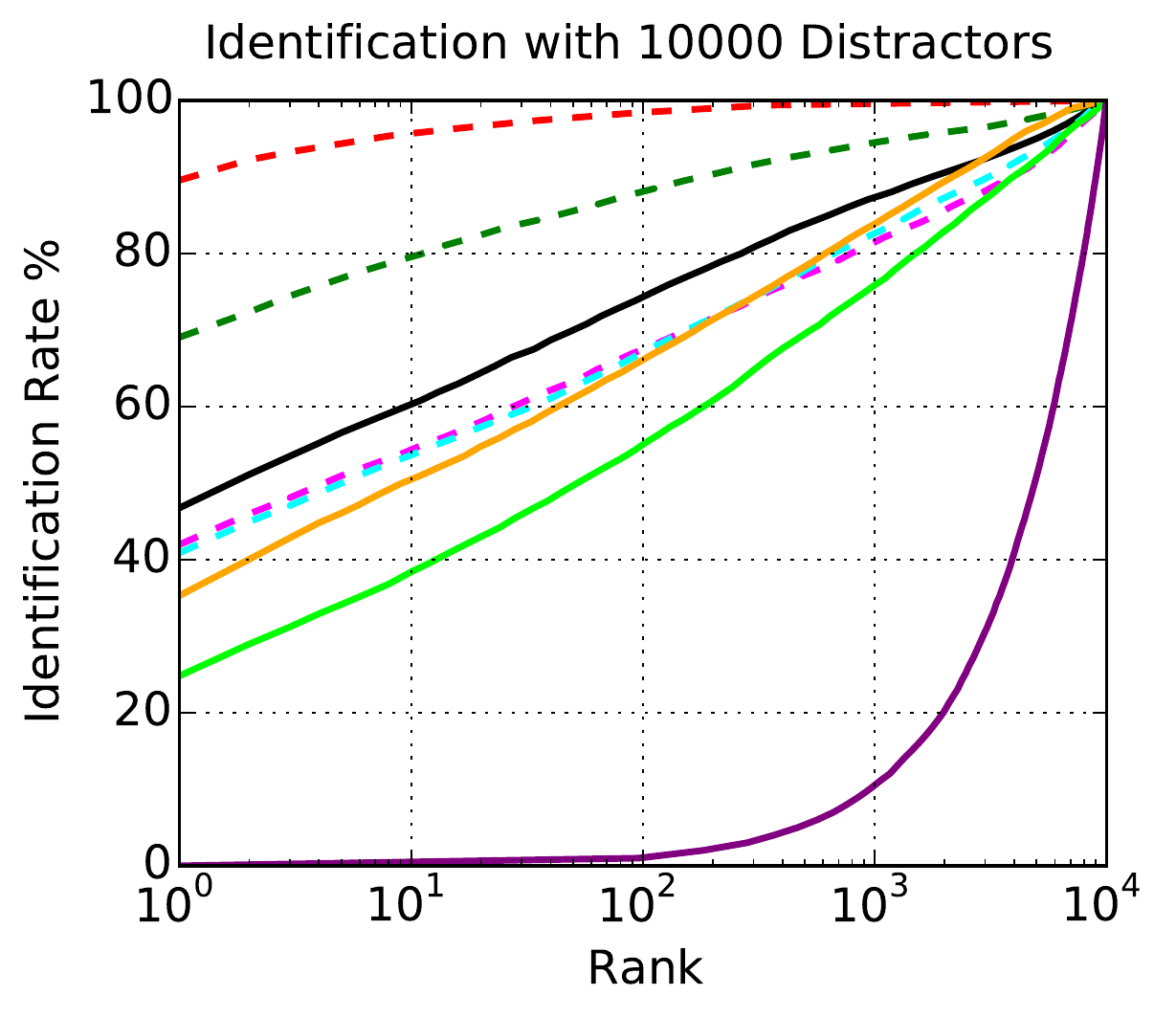} &
				\includegraphics[width=0.25\linewidth, valign=t]{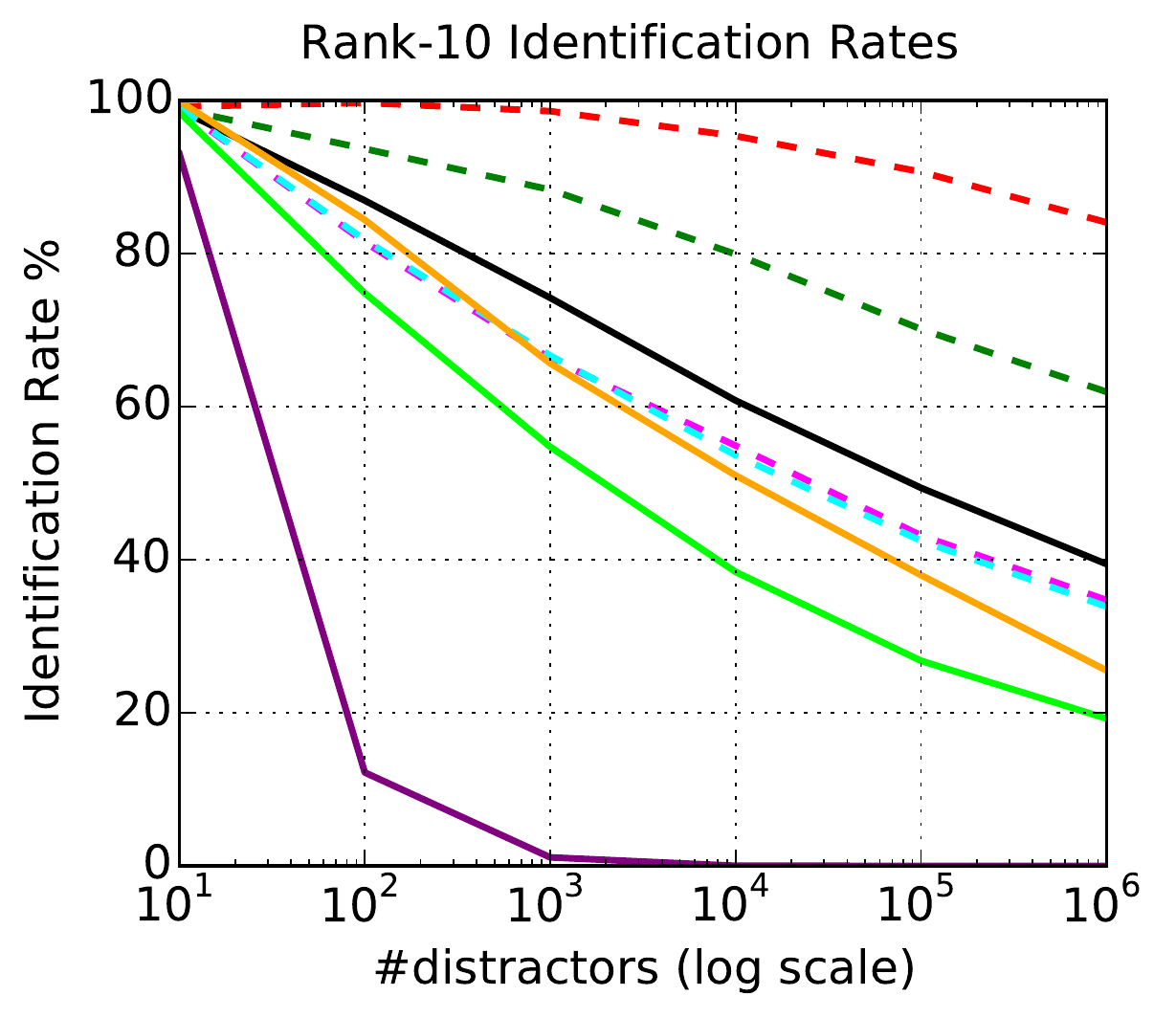}\\
				&		{\footnotesize (d) FGNET + 1M} & {\footnotesize (e) FGNET + 10K}   & {\footnotesize (f) FGNET + rank-10} 
			\end{tabular}
			
			\caption{\textbf{Identification  (random gallery set 3)} performance for all methods with (a,d) 1M distractors and (b,e) 10K distractors, and (c,f) rank-10 for both probe sets.  Fig.~\ref{fig:teaser} also shows  rank-1 performance as a function of number of distractors on both probe sets. }
			\label{fig:dataset_size_cmc}
		\end{figure*}

%
%
%
%
%


\end{document}